\documentclass{article} 

\usepackage[margin=1in]{geometry}

\usepackage[table,xcdraw]{xcolor}
\usepackage[hidelinks,colorlinks=true,linkcolor=blue,citecolor=blue]{hyperref}

\usepackage[numbers,sort&compress]{natbib}

\usepackage{amsmath,amsfonts,bm}

\def\eqref#1{equation~\ref{#1}}

\def\1{\bm{1}}

\DeclareMathAlphabet{\mathsfit}{\encodingdefault}{\sfdefault}{m}{sl}
\SetMathAlphabet{\mathsfit}{bold}{\encodingdefault}{\sfdefault}{bx}{n}

\usepackage{url}

\usepackage[noend]{algpseudocode}
\usepackage{multirow}

\usepackage{cleveref, algorithm, listings}
\algrenewcommand\algorithmicrequire{\textbf{Input:}}
\algrenewcommand\algorithmicensure{\textbf{Output:}}

\usepackage{tcolorbox}
\tcbset{
    boxsep=0mm,       
    left=2mm,         
    right=2mm,        
    top=2mm,          
    bottom=2mm,       
    colback=gray!5,   
    colframe=gray!80, 
    arc=1.5mm,          
    outer arc=1.5mm,    
}

\definecolor{codegray}{rgb}{0.5,0.5,0.5}
\definecolor{codegreen}{rgb}{0,0.6,0}
\definecolor{codepurple}{rgb}{0.58,0,0.82}
\definecolor{backcolour}{rgb}{0.95,0.95,0.92}

\lstdefinestyle{mystyle}{
    backgroundcolor=\color{backcolour},   
    commentstyle=\color{codegreen},       
    keywordstyle=\color{codepurple},      
    numberstyle=\tiny\color{codegray},    
    stringstyle=\color{red},              
    basicstyle=\ttfamily\footnotesize,    
    breakatwhitespace=false,              
    breaklines=true,                      
    captionpos=b,                         
    keepspaces=true,                      
    numbers=left,                         
    numbersep=5pt,                        
    showspaces=false,                     
    showstringspaces=false,               
    showtabs=false,                       
    tabsize=2,                            
}

\usepackage{url}

\UseRawInputEncoding

\usepackage{booktabs}  

\usepackage{subcaption}
\usepackage{caption} 
\usepackage{amssymb}
\usepackage{pifont}

\usepackage[normalem]{ulem}
\useunder{\uline}{\ul}{}

\usepackage[utf8]{inputenc}  

\usepackage{stackengine,scalerel}

\UseRawInputEncoding  

\title{\name{}: On Low-Rank Linearizing of Large Language Models}

\author{Michael Zhang$$\footnote{Corresponding author; correspondence available at \texttt{mzhang@cs.stanford.edu}} $^{\dagger\ddagger}$, Simran Arora$^{\dagger\ddagger}$, Rahul Chalamala$^{\ddagger\S}$, Alan Wu$^\S$, \\ Benjamin Spector$^\dagger$, Aaryan Singhal$^\dagger$, Krithik Ramesh$^{\ddagger\S\S}$, and Christopher R\'{e}$^\dagger$ \\
\vspace{-0.35cm}
\\
 $^\dagger$Department of Computer Science, Stanford University \\
$^\ddagger$Together AI \\
$^\S$California Institute of Technology \\
$^{\S\S}$Massachusetts Institute of Technology \\
\vspace{-0.35cm}
\\
}

\date{}

\newcommand{\name}{\textsc{LoLCATs}}

\newcommand{\cmark}{\ding{51}}%
\newcommand{\xmark}{\ding{55}}%

\newcommand{\eg}{\textit{e.g.,}}
\newcommand{\ie}{\textit{i.e.,}}

\usepackage{enumitem}  
\usepackage{wrapfig,lipsum}

\usepackage{listings}  
\definecolor{codegreen}{rgb}{0,0.6,0}
\definecolor{codeblue}{rgb}{0,0.6,1}
\definecolor{codegray}{rgb}{0.5,0.5,0.5}
\definecolor{codepurple}{rgb}{0.58,0,0.82}
\definecolor{backcolour}{rgb}{0.95,0.95,0.95}

\lstdefinestyle{mystyle}{
    backgroundcolor=\color{backcolour},   
    commentstyle=\color{codegreen},  
    keywordstyle=\color{blue},
    numberstyle=\tiny\color{codegray},
    stringstyle=\color{codepurple},
    basicstyle=\ttfamily\footnotesize,
    breakatwhitespace=false,         
    breaklines=true,                 
    captionpos=b,                    
    keepspaces=true,                 
    numbers=left,                    
    numbersep=5pt,                  
    showspaces=false,                
    showstringspaces=false,
    showtabs=false,                  
    tabsize=2,
    escapeinside={<@}{@>}
}

\begin{document}

\maketitle

\begin{abstract}

\noindent Recent works show we can linearize large language models (LLMs)---swapping the quadratic attentions of popular Transformer-based LLMs with subquadratic analogs, such as linear attention---avoiding the expensive pretraining costs.  
However, linearizing LLMs
often significantly degrades model quality, still requires training over billions of tokens, and remains limited to smaller 1.3B to 7B LLMs.
We thus propose Low-rank Linear Conversion via Attention Transfer (\name{}), a simple two-step method that improves LLM linearizing quality with orders of magnitudes less memory and compute. 
We base these steps on two findings.
First, we can replace an LLM's softmax attentions with closely-approximating linear attentions, simply by 
\textit{training} the linear attentions to match their softmax counterparts with an output MSE loss (``\textit{attention transfer}'').
Then, this enables adjusting for approximation errors and recovering LLM quality simply with \textit{low-rank} adaptation (LoRA).
\name{} significantly improves linearizing quality, training efficiency, and scalability. We significantly reduce the linearizing quality gap and produce state-of-the-art subquadratic LLMs from Llama 3 8B and Mistral 7B v0.1, leading to 20+ points of improvement on 5-shot MMLU. Furthermore, \name{} does so with only 0.2\% of past methods' model parameters and 0.4\% of their training tokens.  
Finally, we apply \name{} to create the first linearized 70B and 405B LLMs (50$\times$ larger than prior work).
When compared with prior approaches under the same compute budgets, \name{} significantly improves linearizing quality, closing the gap between linearized and original Llama 3.1 70B and 405B LLMs by 77.8\% and 78.1\% on 5-shot MMLU.
\end{abstract}

\section{Introduction}
\label{sec:intro}

``Linearizing'' large language models (LLMs)---or converting existing Transformer-based LLMs into attention-free or subquadratic alternatives---has shown promise for scaling up efficient architectures. While many such architectures offer complexity-level efficiency gains, like \textit{linear-time} and \textit{constant-memory} generation, they are often limited to smaller models pretrained on academic budgets~\citep{gu2023mamba,peng2023rwkv,yang2023gated,arora2024simple,beck2024xlstm}.  
In a complementary direction,
linearizing aims to start with openly available LLMs---\eg{} those with 7B+ parameters trained on trillions of tokens~\citep{AI_2024,jiang2023mistral}---and (i) swap their softmax attentions with subquadratic analogs, before (ii) further finetuning to recover quality. This holds exciting promise for quickly scaling up subquadratic capabilities.

However, to better realize this promise and allow anyone to convert LLMs into subquadratic models, we desire methods that are
(1) \textbf{quality-preserving}, \eg{} to recover the zero-shot abilities of modern LLMs; (2) \textbf{parameter and token efficient}, to linearize LLMs on widely accessible compute; and (3) \textbf{highly scalable}, to support linearizing the various 70B+ LLMs available today~\citep{touvron2023llama,touvron2023llama2}. 

Existing methods present opportunities to improve all three criteria.
On quality, despite using motivated subquadratic analogs such
as RetNet-inspired 
linear attentions~\citep{sun2023retentive,mercat2024linearizing} 
or state-space model (SSM)-based 
Mamba layers~\citep{gu2023mamba,yang2024parallelizing,wang2024mamba},
prior works significantly reduce performance on popular LM Evaluation Harness tasks (LM Eval)~\citep{eval-harness} (up to 23.4-28.2 pts on 5-shot MMLU~\citep{hendrycks2020measuring}).
On parameter and token efficiency, to adjust for architectural differences, prior methods update \textit{all} model parameters in at least one stage of training~\citep{mercat2024linearizing,wang2024mamba,yang2024parallelizing}, and use 20~-~100B tokens to linearize 7B LLMs.
On scalability, these training costs make linearizing larger models on academic compute more difficult; 
existing works only linearize up to 8B LLMs. This makes it unclear how to support linearizing 70B to 405B LLMs~\citep{dubey2024llama}.


\begin{figure*}[t]
  \centering
  \includegraphics[width=1\textwidth]
  {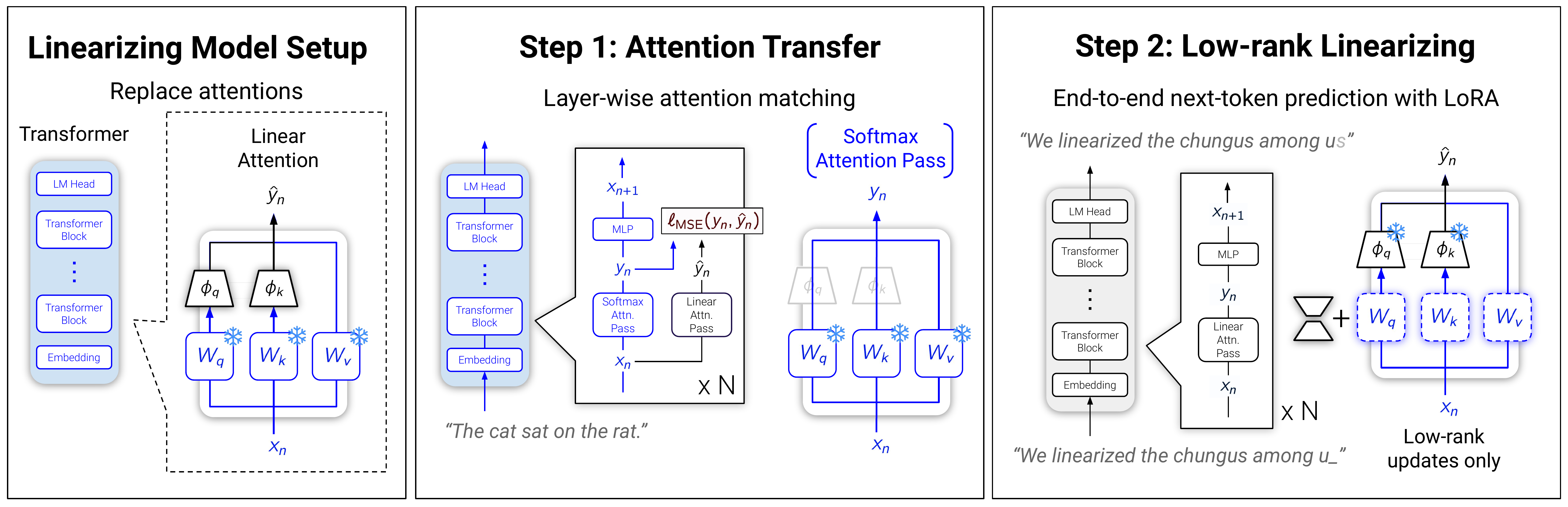}
  \vspace{-0.5cm}
  \caption{\small \textbf{\name{} framework}. We linearize LLMs by (1) training attention analogs to approximate softmax attentions (attention transfer), before swapping attentions and (2) minimally adjusting (with LoRA).}
\vspace{-0.5cm}
\label{fig:pull_method}
\end{figure*}

In this work, we thus propose \textbf{\name{}} 
(\textbf{\textsc{Lo}}w-rank \textbf{\textsc{L}}inear \textbf{\textsc{C}}onversion with \textbf{\textsc{A}}ttention \textbf{\textsc{T}}ran\textbf{\textsc{s}}fer), a simple approach to improve the quality, efficiency, and scalability of linearizing LLMs. 
As guiding motivation, we ask if we can linearize LLMs by simply reducing architectural differences,~\ie{}
\begin{enumerate}[leftmargin=*]
    \item Starting with simple softmax attention analogs such as linear attention (Eq.~\ref{eq:linear_attention}), and \textit{training} their parameterizations explicitly to approximate softmax attention (``\textbf{attention transfer}'').
    \item Subsequently only training with low-cost finetuning to adjust for any approximation errors,~\eg{} with low-rank adaptation (LoRA)~\citep{hu2021lora} (``\textbf{low-rank linearizing}'').
\end{enumerate}

In evaluating this hypothesis, we make several contributions. 
First, to better understand linearizing feasibility, we empirically study attention transfer and low-rank linearizing with existing linear attentions. While intuitive---by swapping in perfect subquadratic softmax attention approximators, we could get subquadratic LLMs with no additional training---prior works suggest linear attentions struggle to match softmax expressivity~\citep{keles2023computational,qin-etal-2022-devil} 
or need full-model updates to recover linearizing quality~\citep{kasai-etal-2021-finetuning,mercat2024linearizing}. 
%
In contrast, we find that while \textit{either} attention transfer or LoRA alone is insufficient, we can rapidly recover quality by simply doing \textit{both} (Figure~\ref{fig:train_efficiency}, Table~\ref{table:train_efficiency}). 
At the same time, we do uncover quality issues related to attention-matching architecture and training. With prior linear attentions, the best low-rank linearized LLMs still significantly degrade in quality  
vs.~original Transformers (up to 42.4 pts on 5-shot MMLU). 
With prior approaches that train all attentions jointly~\citep{zhang2024the}, we also find that later layers can result in 200$\times$ the MSE of earlier ones (Figure~\ref{fig:mse_analogs}). We later find this issue aggravated by larger LLMs; jointly training all of Llama 3.1 405B's 126 attention layers fails to viably linearize the LLM. 

Next, to resolve these issues and improve upon our original criteria, we detail \name{}' method components. 
For \textbf{quality}, we generalize prior notions of learnable linear attentions to sliding window + linear attention variants. These remain subquadratic to compute yet consistently yield better attention transfer via lower mean-squared error (MSE) on attention outputs. For \textbf{parameter and token efficiency}, we maintain our simple 2-step framework of (1) training subquadratic attentions to match softmax attentions, 
before (2) adjusting for any errors via only LoRA. For \textbf{scalability}, we use finer-grained ``block-by-block''
training. We split LLMs into blocks of $k$ layers before jointly training attentions only within each block to improve layer-wise attention matching. We pick $k$ to balance the speed of training blocks in parallel with the memory of saving hidden state outputs of prior blocks (as inputs for later ones). We provide a simple cost model to navigate these tradeoffs.

Finally, in experiments, we validate that \name{} improves on
each of our desired criteria.
\begin{itemize}[leftmargin=*]
    \item On \textbf{quality}, when linearizing popular LLMs such as Mistral-7B and Llama 3 8B, \name{} significantly improves past linearizing methods (by 
    1.1$-$8.6 
    points (pts) on zero-shot LM Eval tasks; 
    +17.2 pts 
    on 5-shot MMLU)). With Llama 3 8B, \name{} for the first time closes the zero-shot LM Eval gap between linearized and Transformer models 
    (73.1 vs 74.2 pts),
    while supporting 3$\times$ higher throughput and 64$\times$ larger batch sizes vs.~popular FlashAttention-2~\citep{dao2023flashattention} implementations (generating 4096 token samples on an 80GB H100). We further validate \name{} as a high-quality training method, outperforming strong 7B subquadratic LLMs (RWKV-v6~\citep{peng2024eagle}, Mamba~\citep{gu2023mamba}, Griffin~\citep{de2024griffin}) and hybrids (StripedHyena~\citep{poli2023hyena}, Zamba~\citep{glorioso2024zamba}) trained from scratch by 
    1.2 to 9.9 pts
    on average over popular LM Eval tasks.
    
    \item On \textbf{parameter and token-efficiency}, by only training linear attention feature maps in Stage 1, while only using LoRA on linear attention projections in Stage 2, \name{} enables these gains while updating only $<$0.2\% of past linearizing methods' model parameters (doable on a single 40GB GPU). This also only takes 40M tokens,~\ie{} 0.003\% and 0.04\% of prior pretraining and linearizing methods' token counts.

    \item On \textbf{scalability}, with \name{} we scale up linearizing to support Llama 3.1 70B and 405B LLMs~\citep{dubey2024llama}. \name{} presents the first viable approach to linearizing larger LLMs. We create the first linearized 70B LLM, taking only 18 hours on one 8$\times$80GB H100 node, and the first linearized 405B LLM with  
    a combination of 5 hours on 14 80GB H100 GPUs (attention transfer) +
    16 hours on three 8$\times$80GB H100 nodes (LoRA finetuning) for Llama 3.1 405B. 
    For both models, this amount to under half the total GPU hours than prior methods reported to linearize 8B models (5 days on 8$\times$80GB A100s)~\citep{wang2024mamba}. Furthermore, under these computational constraints, \name{} significantly improves quality versus prior linearizing approaches without attention transfer. With Llama 3.1 70B and 405B, we close 
    77.8\% and 78.1\% of the 5-shot MMLU gap between Transformers and linearized variants respectively.
\end{itemize}

\noindent Our code is available at: \url{https://github.com/HazyResearch/lolcats}.
\begin{figure}[t!]
    \centering
    \begin{minipage}{0.56\textwidth}  
        \centering
\resizebox{1\linewidth}{!}{
\setlength{\tabcolsep}{2pt}
\begin{tabular}{@{}lccccc@{}}
\toprule
Name           & Architecture                                                                      & \begin{tabular}[c]{@{}c@{}}Quality\\ Preserving\end{tabular} & \begin{tabular}[c]{@{}c@{}}Parameter \\ Efficient\end{tabular} & \begin{tabular}[c]{@{}c@{}}Token \\ Efficient\end{tabular} & \begin{tabular}[c]{@{}c@{}}Validated \\ at Scale\end{tabular} \\ \midrule
\rowcolor[HTML]{EFEFEF} Pretrained     & Attention                                                                         & \cmark \cmark                                              & \xmark \xmark                                                & \xmark \xmark                                            & \cmark \cmark                                               \\
SUPRA          & Linear Attention                                                                  & \xmark                                                     & \xmark                                                       & \cmark                                                   & \cmark                                                      \\
Mohawk         & Mamba (2)                                                                            & \xmark                                                     & \xmark                                                       & \cmark                                                   & \xmark                                                      \\
Mamba in Llama & Mamba (2)                                                                         & \xmark                                                     & \xmark                                                       & \cmark                                                   & \cmark                                                      \\
LoLCATs        & \begin{tabular}[c]{@{}c@{}}Softmax-Approx.\\  Linear Attention\end{tabular} & \cmark                                                     & \cmark                                                       & \cmark \cmark                                            & \cmark \cmark                                               \\ \bottomrule
\end{tabular}
}
\caption{\small \textbf{Linearizing comparison}. \name{} significantly improves LLM linearizing quality and training efficiency. 
No. of \cmark~or \xmark~indicate relatively better or worse support.
}
\label{table:prelim_mmlu}
    \end{minipage}
    \hfill
    \begin{minipage}{0.42\textwidth}  
    \vspace{-0.25cm}
        \centering
        \begin{subfigure}{0.9\textwidth}
            \centering
            \includegraphics[width=\textwidth]
            {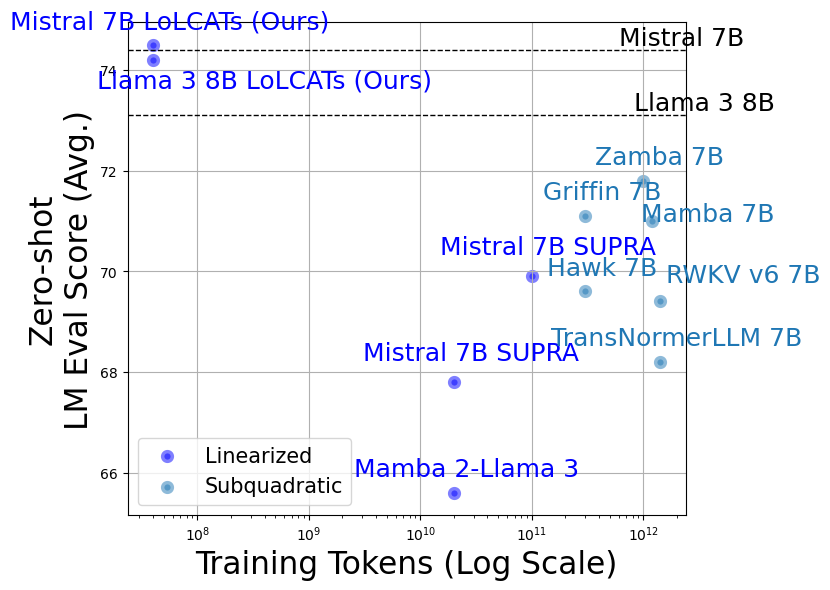} 
        \end{subfigure}
\label{fig:overall_results}
    \end{minipage}
\vspace{-0.5cm}
\end{figure}

\vspace{-0.25cm}
\section{Preliminaries}
\label{sec:preliminaries}
\vspace{-0.25cm}
To motivate \name{}, we first go over Transformers, attention, and linear attention. We then briefly discuss related works on linearizing Transformers and Transformer-based LLMs.

\paragraph{Transformers and Attention.}
Popular LLMs such as Llama 3 8B~\citep{llama3modelcard} and Mistral 7B~\citep{jiang2023mistral} are decoder-only Transformers, with repeated blocks of multi-head \textit{softmax attention} followed by MLPs~\citep{vaswani2017attention}.
For one head, attention computes outputs $\bm{y} \in \mathbb{R}^{l \times d}$ from inputs $\bm{x} \in \mathbb{R}^{l \times d}$ (where 
$l$ is sequence length, 
$d$ is head dimension) with query, key, and value weights $\bm{W}_q, \bm{W}_k, \bm{W}_v \in \mathbb{R}^{d \times d}$. In causal language modeling, we compute $\bm{q} = \bm{x}\bm{W}_q$, $ \bm{k} = \bm{x}\bm{W}_k$, $ \bm{v} = \bm{x}\bm{W}_v$, before getting attention weights $\bm{a}$ and outputs $\bm{y}$ via
\begin{equation}
    a_{n, i} = \frac{\exp(\bm{q}_n^\top \bm{k}_i / \sqrt{d})}{\sum_{i=1}^n \exp(\bm{q}_n^\top \bm{k}_i / \sqrt{d})},
    \;\;\;
    \bm{y}_n = 
    \sum_{i=1}^n a_{n, i} \bm{v}_i,
    \;\;\;
    \text{for $n$ in $[1, \ldots, l]$}
\label{eq:causal_attention}
\end{equation}
Multi-head attention maintains inputs, outputs, and weights for each head, \textit{e.g.,} $\bm{x} \in \mathbb{R}^{h \times l \times d}$ or $ \bm{W}_q \in \mathbb{R}^{h \times d \times d}$ ($h$ being number of heads), and computes Eq.~\ref{eq:causal_attention} for each head. In both cases, we compute final outputs by 
concatenating $\bm{y}_n$ across heads, before using output weights $\bm{W}_o \in \mathbb{R}^{hd \times hd}$ to compute $\bm{y}_n\bm{W}_o \in \mathbb{R}^{l \times hd}$.
While expressive, causal softmax attention requires all $\{ \bm{k}_i, \bm{v}_i\}_{i \leq n}$ to compute $\bm{y}_n$. 
For long context or large batch settings,
this growing \textit{KV cache} can incur prohibitive memory costs even with state-of-the-art implementations such as FlashAttention~\citep{dao2023flashattention}.

\paragraph{Linear Attention.} To get around this,~\citet{katharopoulos2020transformers} show a similar attention operation, but with \textit{linear} time and \textit{constant} memory over generation length (linear time and space when processing inputs).
To see how, note that 
softmax attention's exponential is a kernel function $\mathcal{K}(\bm{q}_n, \bm{k}_i)$, which in general can be expressed as the dot product of feature maps $\phi: \mathbb{R}^{d} \mapsto \mathbb{R}^{d'}$. Swapping $\exp(\bm{q}_n^\top \bm{k}_i/\sqrt{d})$ with $\phi(\bm{q}_n)^\top \phi(\bm{k}_i)$ in Eq.~\ref{eq:causal_attention} gives us \emph{linear attention} weights 
and outputs:
\begin{equation}
    \hat{\bm{y}}_n 
    =
    \sum_{i=1}^n \frac{ \phi(\bm{q}_n)^\top \phi(\bm{k}_i)\bm{v}_i}{\sum_{i=1}^n \phi(\bm{q}_n)^\top \phi(\bm{k}_i) } 
    =    
    \frac{ \phi(\bm{q}_n)^\top \Big(\sum_{i=1}^n  \phi(\bm{k}_i)\bm{v}_i^\top \Big)}{\phi(\bm{q}_n)^\top \sum_{i=1}^n
 \phi(\bm{k}_i)}
\label{eq:linear_attention_2}
\end{equation}
This lets us compute both the numerator $\bm{s}_n = \sum_{i=1}^n \phi(\bm{k}_i)\bm{v}_i^\top$ and denominator $\bm{z}_n = \sum_{i=1}^n \phi(\bm{k}_i)$ as recurrent ``KV states''. With $\bm{s}_0 = \bm{0},\bm{z}_0 = \bm{0}$, we recurrently compute linear attention outputs as
\begin{equation}
\hat{\bm{y}}_n 
= \frac{\phi(\bm{q}_n)^\top \bm{s}_n}{\phi(\bm{q}_n)^\top \bm{z}_n}
\;\;
\text{for}
\;\;
\bm{s}_n = \bm{s}_{n-1} + \phi(\bm{k}_n)\bm{v}_n^\top
\;\;
\text{and}
\;\;
\bm{z}_n = \bm{z}_{n-1} + \phi(\bm{k}_n)
\;\;\;
\label{eq:linear_attention_recurrent}
\end{equation}
Eq.~\ref{eq:linear_attention_2} lets us compute attention over an input sequence of length $n$ in $\mathcal{O}(ndd')$ time and space, while Eq.~\ref{eq:linear_attention_recurrent} lets us compute $n$ new tokens in $\mathcal{O}(ndd')$ time and $\mathcal{O}(dd')$ memory.
Especially during generation, when softmax attention has to compute new tokens sequentially anyway, Eq.~\ref{eq:linear_attention_recurrent} enables time and memory savings if $d' <$ (prompt length + prior generated tokens).


\paragraph{Linearizing Transformers.} 
To combine efficiency with quality, various works propose different $\phi$, (\eg{} $\phi(x) = 1 + \text{ELU}(x)$ as in ~\cite{katharopoulos2020transformers}). However, they typically train linear attention Transformers from scratch. 
%
%
We build upon recent works that \textit{swap} the softmax attentions of \textit{existing} Transformers with linear attention before finetuning the modified models with next-token prediction to recover language modeling quality. 
These include methods proposed for LLMs~\citep{mercat2024linearizing}, and those for smaller Transformers---\textit{e.g.,} 110M BERTs~\citep{devlin2018bert})---reasonably adaptable to modern LLMs~\citep{kasai-etal-2021-finetuning,mao-2022-fine,zhang2024the}.




\section{Method: Linearizing LLMs with \name{}}
\label{sec:method}
We now study how to build a high-quality and highly efficient linearizing method.
In Section~\ref{sec:method_framework}, we present our motivating framework,
which aims to (1) learn good softmax attention approximators with linear attentions and (2) enable low-rank adaptation for recovering linearized quality.
%
In Section~\ref{sec:method_study}, we find that while this attention transfer works surprisingly well for low-rank linearizing with existing linear attentions, on certain tasks, it still results in sizable quality gaps compared to prior methods. 
We also find that attention-transfer quality strongly corresponds with the final linearized model's performance. In Section~\ref{sec:method_new_stuff}, we use our learned findings to overcome prior issues, improving attention transfer to subsequently improve low-rank linearizing quality.

\subsection{\name{} Part 1: A Framework for Low-cost Linearizing}
\label{sec:method_framework}

In this section, we present our initial \name{} framework for linearizing LLMs in an effective yet efficient manner. Our main hypothesis is that by first learning linear attentions that approximate softmax, we can then swap these attentions in as drop-in subquadratic replacements. We would then only need a minimal amount of subsequent training---\eg{} that is supported by low-rank updates---to recover LLM quality in a cost-effective manner effectively. We thus proceed in two steps.

\begin{enumerate}[leftmargin=*]
    \item \textbf{Parameter-Efficient Attention Transfer.} For each softmax attention in an LLM, we aim to learn a closely-approximating linear attention, \ie{} one that computes attention outputs $\hat{\bm{y}} \approx \bm{y}$ for all natural inputs $\bm{x}$. 
    We call this ``attention transfer'', as we aim to \textit{transfer} the attention modeling of existing softmax attentions into target linear attentions. Due to architectural similarity,
    we can treat this as a feature map learning problem, learning $\phi$ to approximate softmax. For each head and layer, let $\phi_q$ and $ \phi_k$ be query and key feature maps. Per head, we compute:
    \begin{equation}
        \bm{y}_n = \underbrace{
    \sum_{i=1}^n \frac{ \exp(\bm{q}_n^\top \bm{k}_i / \sqrt{d})}{\sum_{i=1}^n \exp(\bm{q}_n^\top \bm{k}_i  / \sqrt{d} )} \bm{v}_i
    }_{\text{Softmax Attention}},\;\;\;\;
    \hat{\bm{y}}_n = 
    \underbrace{
    \sum_{i=1}^n \frac{ \phi_q(\bm{q}_n)^\top \phi_k(\bm{k}_i)}{\sum_{i=1}^n \phi_q(\bm{q}_n)^\top \phi_k(\bm{k}_i) } \bm{v}_i
    }_{\text{Linear Attention}}
    \end{equation}
    for all $n \in [l]$ with input $\in \mathbb{R}^{l \times d}$,
    and train $\phi_q, \phi_k$ to minimize sample mean squared error (MSE)
    \begin{equation}
        \ell_\text{MSE} = \frac{1}{MH}\sum_{m = 1}^M \sum_{h=1}^H
        \ell_\text{MSE}^{h, m}\;\;,\;\;\;\; 
        \ell_\text{MSE}^{h, m} = \frac{1}{d} \sum_{n=1}^d (\bm{y}_n - \hat{\bm{y}}_n)^2
        \label{eq:mse_loss}
    \end{equation}
    \ie{} jointly for each head $h$ in layer $m$.
    Similar to past work~\citep{kasai-etal-2021-finetuning,zhang2024the}, rather than manually design $\phi$, we parameterize each $\phi : \mathbb{R}^{d} \mapsto \mathbb{R}^{d'}$ as a \textit{learnable} layer:
    \[
    \phi_q(\bm{q}_n) := f(\bm{q}_n\tilde{\bm{W}}_{(q)} + \tilde{\bm{b}}_{(q)}),\;\;\;
    \phi_k(\bm{k}_i) := f(\bm{k}_i\tilde{\bm{W}}_{(k)} + \tilde{\bm{b}}_{(k)})
    \]
    Here $\tilde{\bm{W}} \in \mathbb{R}^{d \times d'}$ and $\tilde{\bm{b}} \in \mathbb{R}^{d'}$ are trainable weights and optional biases, 
    $f(\cdot)$ is a nonlinear activation, and $d'$ is an arbitrary feature dimension (set to equal head dimension $d$ in practice).


\item \textbf{Low-rank Adjusting.} After training the linearizing layers, we replace the full-parameter training of prior work with low-rank adaptation (LoRA)~\citep{hu2021lora}. 
Like prior work, to adjust for the modifying layers and recover language modeling quality, we now train the modified LLM end-to-end over tokens to minimize a sample next-token prediction loss
\(
\ell_\text{xent} = -\sum \log P_\Theta(\bm{u}_{t+1} \mid \bm{u}_{1:t})
\).
%
Here $P_\Theta$ is the modified LLM, $\Theta$ is the set of LLM parameters, and we aim to maximize the probability of true $\bm{u}_{t+1}$ given past tokens $\bm{u}_{1:t}$ (Fig.~\ref{fig:pull_method}~right). 
However, rather than train all LLM parameters, we only train the swapped linear attention $\bm{W}_q, \bm{W}_k, \bm{W}_v, \bm{W}_o$ with LoRA, 
updating $\bm{W}' \leftarrow \bm{W} + \Delta \bm{W}$ with $\Delta \bm{W}$
as the product of two low-rank matrices $\bm{B}\bm{A}$, $\bm{B} \in \mathbb{R}^{d \times r}$, $\bm{A} \in \mathbb{R}^{r \times d}$. 
This enables parameter efficiency for 
$r \ll d$. 
\end{enumerate}

\textbf{Training footprint and efficiency}. Both steps remain parameter-efficient.
    For Step 1, optimizing Eq.~\ref{eq:mse_loss} is similar to a layer-by-layer cross-architecture distillation. We compute layer-wise ($\bm{x}, \bm{y}$) as pretrained attention inputs and outputs, using an LLM forward pass over natural language samples (Fig.~\ref{fig:pull_method} middle).
    However, to keep our training footprint low, we freeze the original pretrained attention layer's parameters and 
    simply \textit{insert} new $\phi_q, \phi_k$ after $\bm{W}_q, \bm{W}_k$ in each softmax attention (Fig.~\ref{fig:pull_method} left). We compute outputs $\bm{y},\hat{\bm{y}}$ with the same attention weights in separate passes (choosing either Eq.~\ref{eq:causal_attention} or Eq.~\ref{eq:linear_attention_2}; ``teacher-forcing'' by only sending softmax outputs $\bm{y}$ to future layers and preventing error propagation, Fig.~\ref{fig:pull_method} middle). 
    %
    %
    For Llama 3 8B or Mistral 7B, training $\phi_q,\phi_k$ with $d' = 64$ then only takes 32 layers $\times$ 32 heads $\times$ 2 feature maps $\times$ (128 $\times$ 64) weights $\approx$ 16.8M trainable weights (0.2\% of LLM sizes). 
    Furthermore, Eq.~\ref{eq:mse_loss} keeps attention transfer memory-efficient. While \cite{zhang2024the} also train layer-wise $\phi$ to approximate softmax attention, they supervise by matching on the $n^2$ \textit{attention weights}  computed for $n$-token inputs (App.~\ref{appendix:related_work}, Eq.~\ref{eq:hedgehog_loss}). This scales poorly for large $n$. Instead, by only needing attention outputs, we can reduce memory from $\mathcal{O}(n^2)$ to $\mathcal{O}(n)$, computing softmax attention $\bm{y}$ with FlashAttention~\citep{dao2022flashattention, dao2023flashattention} 
    and 
    linear attention $\hat{\bm{y}}$ with Eq.~\ref{eq:linear_attention_2} (or~\ref{eq:hybrid_linear_swa}) both in $\mathcal{O}(n)$ memory by not materializing attention weights. Despite these savings, we show similar attention weight recovery in App.~\ref{app:artifacts_attention_weights}.
    In Step 2, LoRA with $r=8$ on all attention projections suffices for state-of-the-art quality. This updates just $<$0.09\% of 7B parameters.


\subsection{Baseline Study: Attention Transfer and Low-rank Linearizing}
\label{sec:method_study}

As a first step, we aim to understand if attention transfer and low-rank adjusting as proposed are sufficient for linearizing LLMs. 
While simple,
it is unclear whether these steps can lead to high-quality linearizing, as all 
prior works default to more involved approaches~\citep{mercat2024linearizing,bick2024transformers,wang2024mamba}. They use linearizing layers featuring GroupNorms~\citep{wu2018group} and decay factors~\citep{sun2023retentive}, or alternate SSM-based architectures~\citep{gu2023mamba,dao2024transformers}. They also all use full-LLM training after swapping in the subquadratic layers. 
In contrast, as a first contribution we find that simple linear attentions \textit{can} lead to viable linearizing, with attention transfer + LoRA obtaining competitive quality on 4~/~6 popular LM Eval tasks. 

\begin{wraptable}{r}{0.5\textwidth}  
\vspace{-0.35cm}
\centering
\resizebox{1\linewidth}{!}{
\setlength{\tabcolsep}{2pt}
\begin{tabular}{@{}lcl@{}}
\toprule
Feature Map              & $\phi(\bm{q})$ (same for $\bm{k}$) & Weight Shapes     \\ \midrule
T2R  & $\text{ReLU}(\bm{q}\tilde{\bm{W}} + \tilde{\bm{b}})$ & 
$\tilde{\bm{W}}$: (128, 128), $\tilde{\bm{b}}$: (128,) \\
Hedgehog               & $[\text{SM}_d(\bm{q}\tilde{\bm{W})} \oplus \text{SM}_d(-\bm{q}\tilde{\bm{W})} ]$ &
$\tilde{\bm{W}}: (128, 64)$ 
\\ \bottomrule
\end{tabular}
}
\vspace{-0.25cm}
\caption{\small \textbf{Learnable feature maps}. Transformer to RNN (T2R) from~\cite{kasai-etal-2021-finetuning}, Hedgehog from~\cite{zhang2024the}, both $\oplus$ (concat) and $\text{SM}_d$ (softmax) apply over feature dimension.}
\label{table:linear_attentions_def}
\vspace{-1cm}
\end{wraptable}

\paragraph{Experimental Setup.} We test the \name{} framework by linearizing two popular base LLMs, Llama 3 8B~\citep{AI_2024} and Mistral 7B v0.1~\citep{jiang2023mistral}. For linearizing layers, we study two feature maps used in prior work 
(Table~\ref{table:linear_attentions_def}). 
To support the rotary positional embeddings (RoPE)~\citep{su2024roformer} in these LLMs, we apply the feature maps $\phi$ after RoPE,\footnote{
Unlike prior works that apply $\phi$ before RoPE~\citep{mercat2024linearizing, su2024roformer}, our choice preserves the linear attention kernel connection, where we can hope to learn $\phi_q, \phi_k$ for $\exp(\bm{q}_n^\top \bm{k}_i/\sqrt{d}) \approx \phi_q(\bm{q}_n)^\top \phi_k(\bm{k}_i)$. 
}
\ie{} computing query features $\phi_q(\bm{q}) = f(\text{RoPE}(\bm{q})\tilde{\bm{W}}_q + \tilde{\bm{b}}))$.
For linearizing data, we wish to see if \name{} with a small amount of data can recover general zero-shot and instruction-following LLM abilities. We use the 50K samples of a cleaned Alpaca dataset\footnote{\url{https://huggingface.co/datasets/yahma/alpaca-cleaned}}, due to its ability to improve general instruction-following in 7B LLMs despite its relatively small size~\citep{alpaca}. 
We train all feature maps jointly. 
Training code and implementation details are in App.~\ref{appendix:implementation_details}.

\begin{figure}[t]
    \centering
    \begin{subfigure}[b]{0.24\textwidth}
        \centering
        \includegraphics[width=\textwidth]{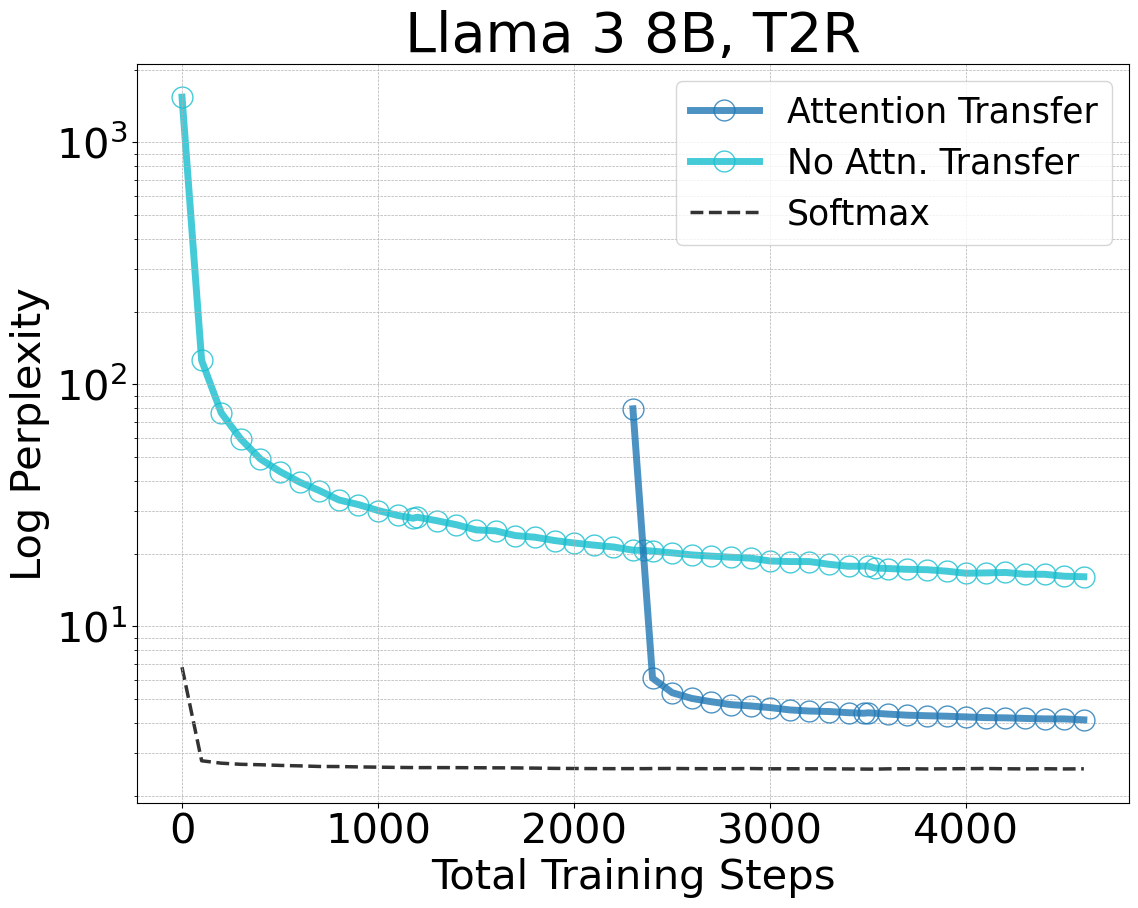}
    \label{fig:linearize_efficiency_llama3_t2r}
    \end{subfigure}
    \begin{subfigure}[b]{0.24\textwidth}
        \centering
        \includegraphics[width=\textwidth]{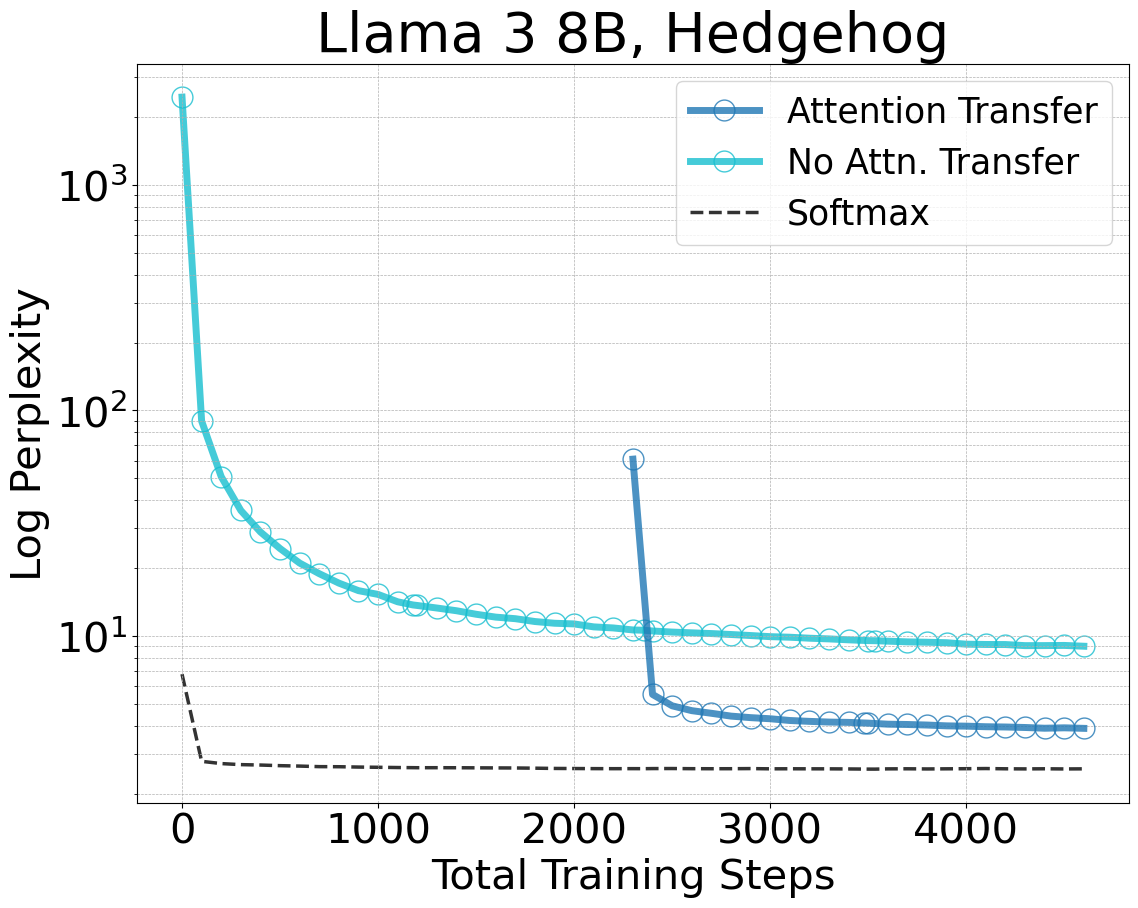}
        \label{fig:linearize_efficiency_llama3_hh}
    \end{subfigure}
    \begin{subfigure}[b]{0.24\textwidth}
        \centering
        \includegraphics[width=\textwidth]{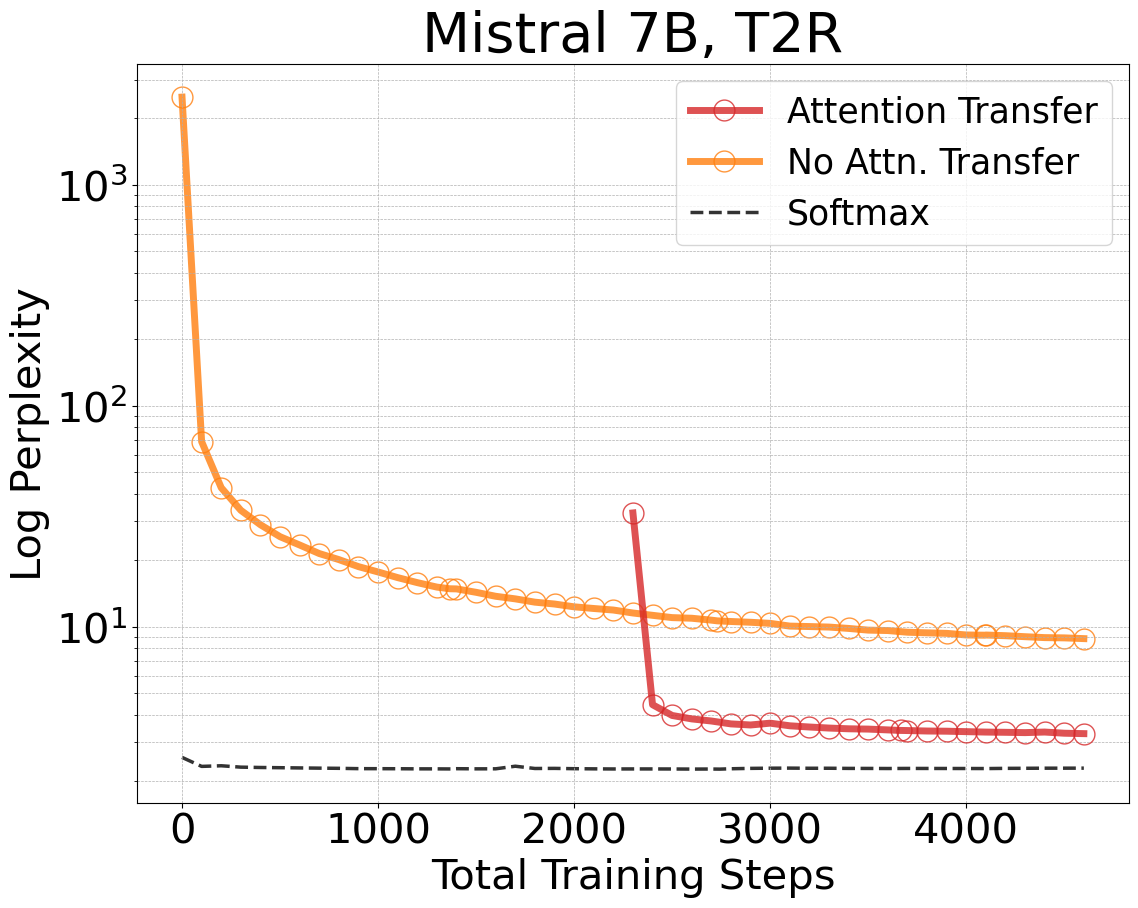}
\label{fig:linearize_efficiency_mistral_7b_t2r}
    \end{subfigure}
    \begin{subfigure}[b]{0.24\textwidth}
        \centering
        \includegraphics[width=\textwidth]{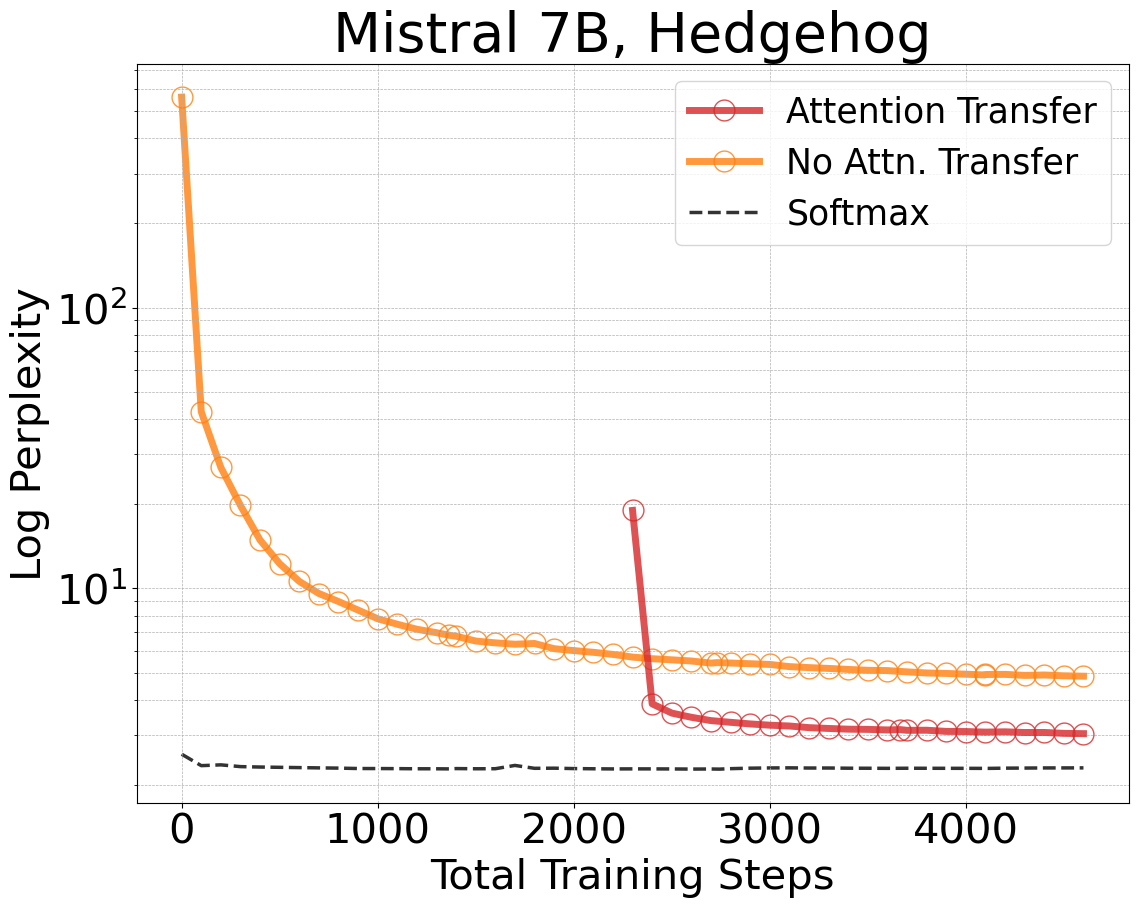}
    \label{fig:linearize_efficiency_mistral_7b_hh}
    \end{subfigure}
    \vspace{-0.45cm}
    \caption{\small \textbf{Attention transfer training efficiency}. Even accounting for initial training steps, low-rank linearizing with attention transfer still consistently achieves lower perplexity faster across feature maps and LLMs.}
    \label{fig:train_efficiency}
    \vspace{-0.5cm}
\end{figure}

To study the effects of attention transfer and low-rank linearizing across LLMs and linear attention architectures, we evaluate their validation set perplexity (Table~\ref{table:train_efficiency}, Fig.~\ref{fig:train_efficiency}) and downstream LM Eval zero-shot quality (Table~\ref{table:prelim_mmlu}).
We use the same data for both stages, early stopping, and either 2 epochs for attention transfer and LoRA adjusting or 4 epochs for either alone ($\approx$~40M total training tokens). We use LoRA $r=8$ by popular default~\citep{hu2021lora}, training 0.2\% of LLM parameter counts. 

\begin{table}[h]
\vspace{-0.25cm}
\begin{center}
\resizebox{0.96\linewidth}{!}{
\begin{tabular}{@{}ccccccccc@{}}
\toprule
& \multicolumn{4}{c}{Llama 3 8B}                                  & \multicolumn{4}{c}{Mistral 7B}                                  \\ 
\cmidrule(lr){2-5} \cmidrule(lr){6-9}
  \multicolumn{1}{c}{Attention}                       & \multicolumn{2}{c}{T2R}        & \multicolumn{2}{c}{Hedgehog}   & \multicolumn{2}{c}{T2R}        & \multicolumn{2}{c}{Hedgehog}   \\ \cmidrule(lr){2-3} \cmidrule(lr){4-5} \cmidrule(lr){6-7} \cmidrule(lr){8-9}
\multicolumn{1}{c}{Transfer?} & \cellcolor[HTML]{ECF4FF} PPL@0      &  PPL@2/4     & \cellcolor[HTML]{ECF4FF} PPL@0      &  PPL@2/4       & \cellcolor[HTML]{FFDECA}  PPL@0      &  PPL@2/4      & \cellcolor[HTML]{FFDECA} PPL@0      &  PPL@2/4     \\ 
\cmidrule(lr){1-1}
\cmidrule(lr){2-3}\cmidrule(lr){4-5} \cmidrule(lr){6-7} \cmidrule(lr){8-9}
No\; \xmark               & \cellcolor[HTML]{ECF4FF} 1539.39        &  16.05         & \cellcolor[HTML]{ECF4FF} 2448.01        &  9.02          & \cellcolor[HTML]{FFDECA} 2497.13        &  8.85          & \cellcolor[HTML]{FFDECA} 561.47         &  4.87          \\
Yes\; \cmark               & \cellcolor[HTML]{ECF4FF} \textbf{79.33} & \textbf{4.11} & \cellcolor[HTML]{ECF4FF} \textbf{60.86} &  \textbf{3.90} & \cellcolor[HTML]{FFDECA} \textbf{32.78} &  \textbf{3.29} & \cellcolor[HTML]{FFDECA} \textbf{18.94} &  \textbf{3.04} \\ \bottomrule
\end{tabular}
}
\end{center}
\vspace{-0.25cm}
\caption{\small Alpaca validation set perplexity (PPL) of linearized LLMs, comparing attention transfer, no LoRA adjusting (PPL@0) and PPL after training (PPL@2/4; 2 with attention transfer, 4 without, for equal total steps).
}
\label{table:train_efficiency}
\vspace{-0.25cm}
\end{table}

\paragraph{Attention Transfer + LoRA Enables Fast LLM Linearizing.} In Table~\ref{table:train_efficiency} and Fig.~\ref{fig:train_efficiency}, we report the validation PPL of linearized LLMs, ablating attention transfer and LoRA adjusting. We find that while attention transfer alone is often insufficient (\textit{c.f.,} PPL@0, Table~\ref{table:train_efficiency}), a single low-rank update rapidly recovers performance by 15--75 PPL (Fig.~\ref{fig:train_efficiency}), 
where training to approximate softmax leads to up to 11.9 lower PPL than no attention transfer.
%
Somewhat surprisingly, this translates to performing competitively with prior linearizing methods that train \textit{all} model parameters~\citep{mercat2024linearizing,wang2024mamba} (within 5 accuracy points on 4~/~6 popular LM Eval tasks; Table~\ref{table:prelim_mmlu}), while \textit{only} training with 0.04$\%$ of their token counts and 0.2\% of their parameter counts. The results suggest we can linearize 7B LLMs at orders-of-magnitude less training costs than previously shown.


\begin{figure}[b]
\vspace{-0.25cm}
    \centering
    \begin{minipage}{0.45\textwidth}  
        \centering
\resizebox{1\linewidth}{!}{
\setlength{\tabcolsep}{2pt}
\begin{tabular}{@{}lccccccc@{}}
\toprule
Model               & Tokens (B) & PiQA          & ARC-E      & ARC-C & HS     & WG    & MMLU\\ \midrule
\rowcolor[HTML]{ECF4FF} 
Llama 3 8B          & -          & 79.9          & 80.1          & 53.3          & 79.1          & 73.1          & 66.6          \\
$\rightarrow$ Mamba2      & 100        & 76.8          & \textbf{74.1} & \textbf{48.0} & \textbf{70.8} & \textbf{58.6} & \textbf{43.2} \\
$\rightarrow$ LoRA Hedgehog & 0.04       & \textbf{77.4} & 71.1          & 40.6          & 66.5          & 54.3          & 24.2          \\ \midrule
\rowcolor[HTML]{FFDECA} 
Mistral 7B          & -          & 82.1          & 80.9          & 53.8          & 81.0          & 74.0          & 62.4          \\
$\rightarrow$ SUPRA    & 100        & \textbf{80.4} & 75.9          & \textbf{45.8} & \textbf{77.1} & \textbf{70.3} & \textbf{34.2} \\
$\rightarrow$ LoRA Hedgehog & 0.04       & 79.3          & \textbf{76.4} & 45.1          & 73.1          & 57.5          & 28.2          \\ \bottomrule
\vspace{-0.5cm}
\end{tabular}
}
\caption{\small \textbf{Linearizing comparison on LM Eval}. Task names in Table~\ref{table:main_results}.~Acc.~norm:~ARC-C, HS. Acc.~otherwise.~5-shot MMLU. 0-shot otherwise.}
\label{table:prelim_mmlu}
    \end{minipage}
    \hfill
    \begin{minipage}{0.54\textwidth}  
    \vspace{-0.125cm}
        \centering
        \begin{subfigure}{0.48\textwidth}
            \centering
            \includegraphics[width=\textwidth]{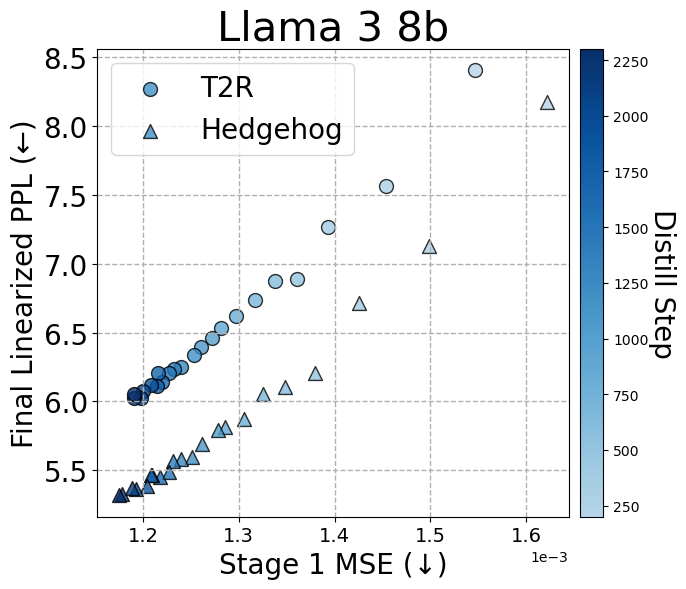} 
        \end{subfigure}
        \hfill
        \begin{subfigure}{0.48\textwidth}
            \centering
            \includegraphics[width=\textwidth]{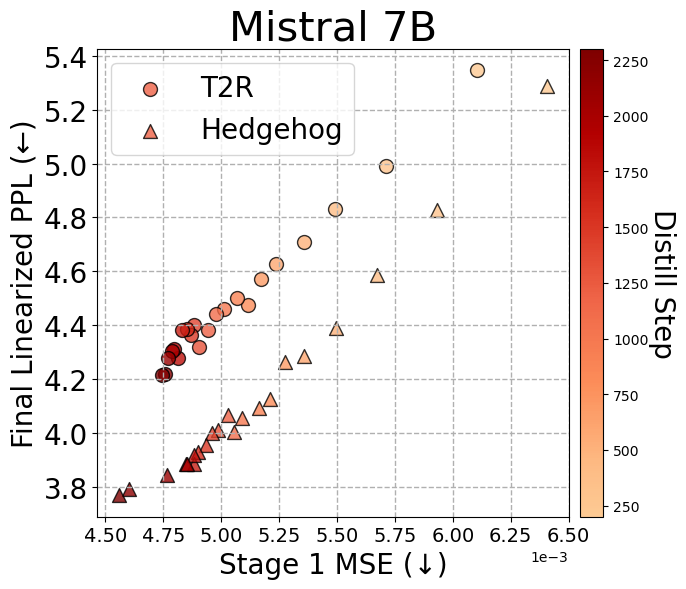} 
        \end{subfigure}
\vspace{-0.25cm}
\caption{\small \textbf{Attention~MSE vs.~PPL}. Across feature maps, LLMs; lower MSE coincides with better linearized quality.}
\label{fig:mse_vs_ppl}
    \end{minipage}
\end{figure}


\begin{figure}[t]
\vspace{-0.25cm}
    \centering
    \begin{subfigure}[b]{0.44\textwidth}
        \centering
        \includegraphics[width=\textwidth]{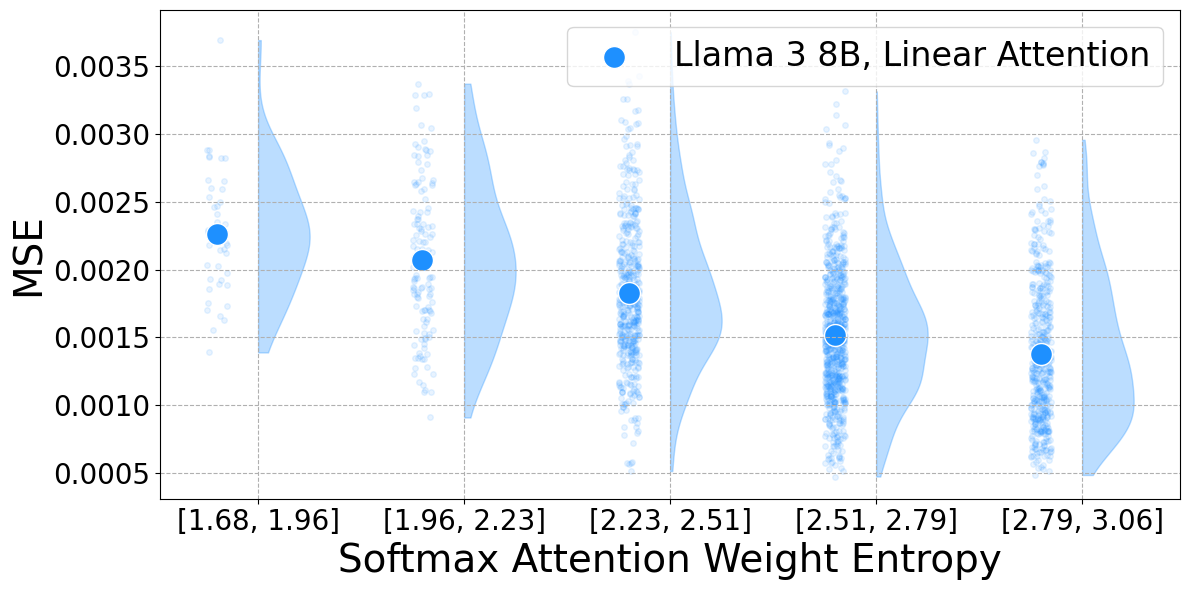}
        \caption{Attention Entropy}
        \label{fig:mse_error_llama3_entropy}
    \end{subfigure}
    \begin{subfigure}[b]{0.54\textwidth}
        \centering
        \includegraphics[width=\textwidth]{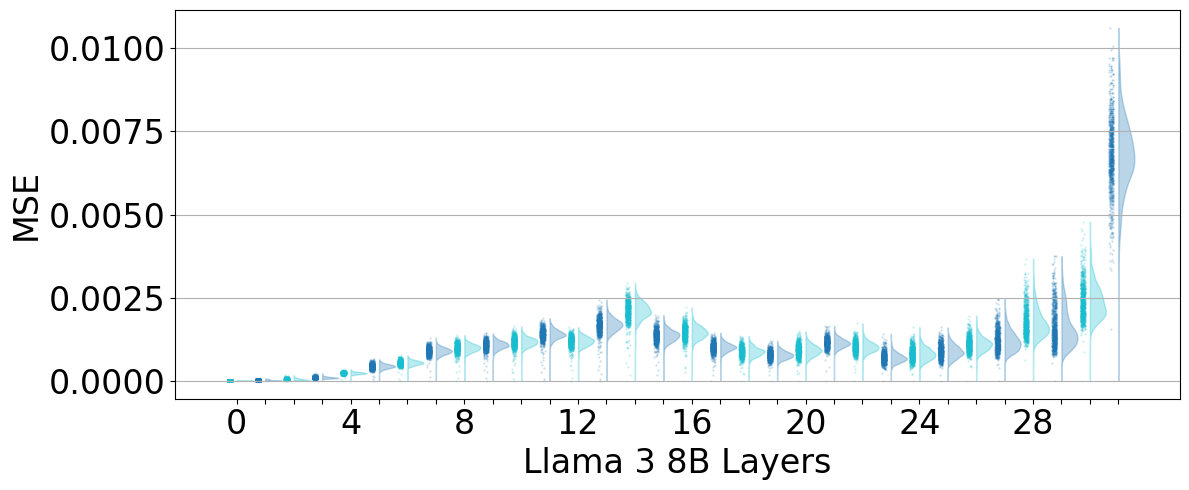}
        \caption{Model Layer}
        \label{fig:mse_error_llama3_layer}
    \end{subfigure}
    \vspace{-0.25cm}
    \caption{\textbf{Sources of Attention Transfer Error} with Llama 3 8B. We find two potential sources of attention transfer difficulty: (a) low softmax attention entropy and (b) attentions in later layers. 
    }
    \label{fig:mse_error}
    \vspace{-0.25cm}
\end{figure}


\paragraph{\textsc{LoL SAD}: Limitations of Low-Rank Linearizing.} At the same time, we note quality limitations with the present framework. 
While sometimes close, low-rank linearized LLMs perform worse than full-parameter alternatives and original Transformers on 5~/~6 LM Eval tasks (up to 42.4 points on 5-shot MMLU; Table~\ref{table:prelim_mmlu}). 
To understand the issue, we study if the attention transfer stage can produce close linear attention approximations of LLM softmax attentions. We note 
three
observations:
\begin{enumerate}[leftmargin=*]
    \item Attention transfer quality (via output MSE) strongly ties to final linearized LLM quality (via PPL) (Fig.~\ref{fig:mse_vs_ppl}), suggesting we can improve quality by reducing MSE with softmax attentions.
    \item However, larger MSEs coincide with lower softmax attention weight entropies (Fig.~\ref{fig:mse_error_llama3_entropy}). 
    ~\cite{zhang2024the} find  linear attentions struggle to approximate such ``spikier'' distributions, suggesting we may need better attention-matching layers
    to reduce MSE and improve final linearized quality.
    \item When training layers jointly like in prior work~\citep{zhang2024the}, larger MSEs also heavily concentrate in later layers (Fig.~\ref{fig:mse_error_llama3_layer}). To bring the MSE in these layers down, we may thus need more fine-grained layer-wise supervision, rather than the objective over all layers in Eq.~\ref{eq:mse_loss}.
\end{enumerate}

\subsection{\name{} Part 2: Components to Improve Low-rank Linearizing}
\label{sec:method_new_stuff}
Following our motivating hypotheses, framework, and observations, we now introduce two simple improvements to improve linearized LLM quality by reducing MSE: (1) better attention-matching architectures (\Cref{sec:method_analogs}), and (2) finer-grained layer-wise attention transfer (\Cref{sec:method_training}).

\begin{figure}[b]
    \centering
    \begin{subfigure}[b]{0.48\textwidth}
        \centering
        \includegraphics[width=\textwidth]{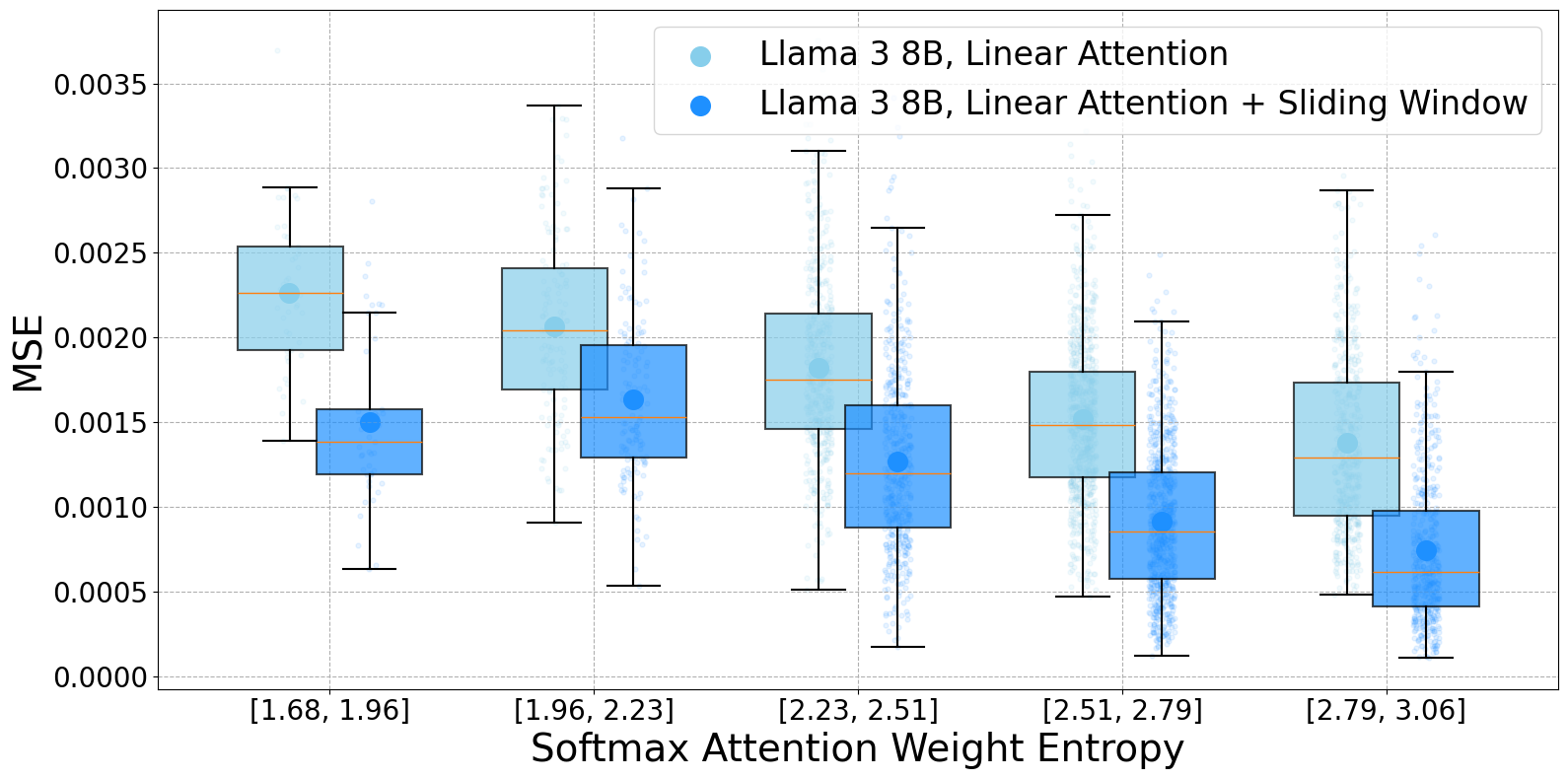}
        \label{fig:mse_analogs_mistral7b}
    \end{subfigure}
    \begin{subfigure}[b]{0.48\textwidth}
        \centering
        \includegraphics[width=\textwidth]{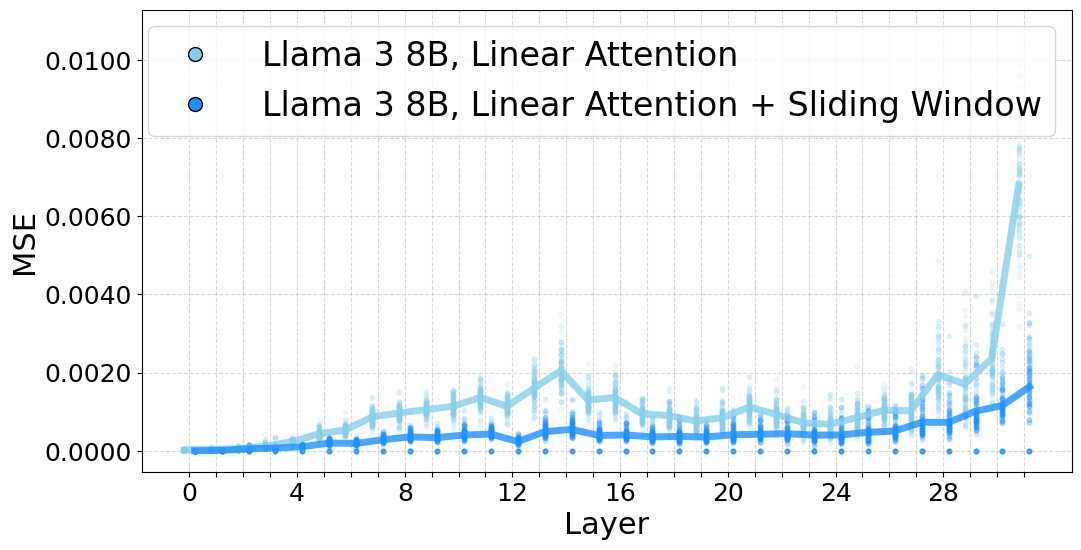}
        \label{fig:mse_analogs_llama3}
    \end{subfigure}
    \vspace{-0.25cm}
    \caption{\textbf{Improving Attention matching MSE}. Linearizing with linear + sliding window attention better matches LLM softmax attentions (lower MSE) over attention entropy values and LLM layers.}
    \label{fig:mse_analogs}
    \vspace{-0.25cm}
\end{figure}


\subsubsection{Architecture: Generalizing Learnable Linear Attentions}
\label{sec:method_analogs}
As described, we can apply our framework with any linear attentions with learnable $\phi$ (\textit{e.g.}, T2R and Hedgehog, Figure~\ref{fig:train_efficiency}). 
%
However, to improve attention-matching quality, we introduce a hybrid $\phi$ parameterization combining linear attention and \textit{sliding window} attention. Motivated by prior works that show quality improvements when combining attention layers with linear attentions~\citep{arora2024simple,Munkhdalai2024-kx}, 
we combine \textcolor{blue}{short sliding windows} of softmax attention~\citep{beltagy2020longformer,zhu2021long} (size 64 in experiments) followed by linear attention in a single layer. This allows attending to all prior tokens for each layer while keeping the entire LLM subquadratic. For window size $w$ and token indices $[1, \ldots, n-w, \ldots, n]$, we apply the softmax attention over the $w$ most recent tokens, 
and compute attention outputs $\hat{\bm{y}}_n$ as
\begin{equation}
    \hat{\bm{y}}_n =  
    \frac{ \textcolor{blue}{\sum_{i=n-w+1}^n \gamma \exp(\bm{q}_n^\top \bm{k}_i / \sqrt{d} - \bm{c}_n)\bm{v}_i} + \phi_q(\bm{q}_n)^\top \big( \sum_{j=1}^{n-w}\phi_k(\bm{k}_j)\bm{v}_j^\top \big)}{ \textcolor{blue}{\sum_{i = {n-w+1}}^n\gamma \exp(\bm{q}_n^\top \bm{k}_i / \sqrt{d} - \bm{c}_n)}\; +  \phi_q(\bm{q}_n)^\top \big( \sum_{j=1}^{n-w}\phi_k(\bm{k}_j)^\top \big)} 
\label{eq:hybrid_linear_swa}
\end{equation}
$\gamma$ is a learnable mixing term, and $\bm{c}_n$ is a stabilizing constant as in log-sum-exp calculations ($\bm{c}_n = \max_i\big\{\bm{q}_n^\top \bm{k}_i / \sqrt{d} : i \in [n-w+1, \ldots, n]\big\}$).
Like before, we can pick any learnable $\phi$.


\paragraph{Subquadratic efficiency.} The hybrid layer retains linear time and constant memory generation. For $n$-token prompts, we initially require $\mathcal{O}(w^2 d)$ and $\mathcal{O}((n-w)dd')$ time and space for window and linear attention respectively, attending over a $w$-sized KV-cache and computing KV and K-states (Eq.~\ref{eq:linear_attention_recurrent}). 
For generation, we only need $\mathcal{O}(w^2 d + dd')$ time and space for every token. We evict the KV-cache's first $\bm{k}$, $\bm{v}$, compute $\phi_k(\bm{k})$, and add $\phi_k(\bm{k})\bm{v}^\top$ and  $\phi_k(\bm{k})$ to KV and K-states respectively.

\paragraph{Hardware-aware implementation.} To make Eq.~\ref{eq:hybrid_linear_swa} competitive with modern softmax attentions like FlashAttention-2~\citep{dao2023flashattention}, we provide a ``hardware-aware'' Eq.~\ref{eq:hybrid_linear_swa} with the Hedgehog feature map. For space, we defer implementation details to App.~\ref{appendix:experimental_details},~\ref{app:tk_implementation}. We evaluate this version by default. 

\subsubsection{Training: Layer (or Block)-wise Attention Transfer}
\label{sec:method_training}
We describe our training approach and provide a simplified model to show its cost-quality tradeoffs.
%
Based on the layer-wise MSE differences in Sec.~\ref{sec:method_study} from training all layers jointly (Eq.~\ref{eq:mse_loss}),
we instead generalize to training over finer-grained $b$-layer blocks, and train each block independently:
\begin{equation}
        \ell_\text{MSE}^{\text{block}} = \frac{1}{bH}\sum_{m = i}^{i + b} \sum_{h=1}^H
        \ell_\text{MSE}^{h, m}\;\;\;\; 
        (\text{for blocks starting at layers } i = 0, b, 2b, \ldots)
        \label{eq:mse_loss_blocks}
    \end{equation}
We choose $b$ to balance quality and cost in both memory and training time, which we find particularly helpful for linearizing larger LLMs. Several block-wise training approaches exist,
including joint training with separate optimizer groups per block, sequentially training separate blocks, or precomputing hidden states and training blocks in parallel across GPUs. For space, we report results on LLM quality trade-offs in App.~\ref{app:blockwise_study}, where we find smaller $b$ improves \name{} Llama 3.1 405B PPL by 1.02 points (Table~\ref{table:block_wise_ppl}). Below we discuss primary cost trade-offs:
\begin{itemize}[leftmargin=*]
    \item \textbf{Compute}: 
    While the joint training of Llama 3.1 405B in 16-bit precision uses \textit{multiple nodes} (e.g., NVIDIA H100 8$ \times$80GB nodes), an individual block of $b=9$ or fewer layers can be trained on a \textit{single GPU} (e.g., H100 80GB GPU) at sequence length 1024. 
    \item \textbf{Time}: However, doing so on a single GPU multiples total training time by $M / b$. With multiple GPUs, we can distribute training different blocks on difference devices in parallel. To train blocks at layers $b, 2b, ...$, we need the outputs from each prior block. 
    As we teacher-force with the ``true'' softmax attention outputs, we can simply precompute these with the original Transformer.
    \item \textbf{Memory}: Lastly, we need to save each precomputed block's outputs 
    to disk.
    The total disk space required is 
\(
2 \times T \times d \times \frac{L}{k}
\)
for total training tokens $T$, model dimension $d$, number of layers $L$ and 2-byte (16-bit) precision. 
For Llama 3.1 405B, saving states per-layer ($b=1$) for just 50M tokens requires over 200TB of disk space. Larger $b$ divides this storage, potentially at the cost of quality.

\end{itemize}
For each target LLM, we thus aim to make low-rank linearizing feasible by first doing block-wise attention transfer, adjusting parameters based on linearizing quality and what is feasible in compute, time, and memory. We summarize \name{} with Alg.~\ref{alg:lolcats_stage_1},~\ref{alg:lolcats_stage_2}, providing pseudocode in App.~\ref{appendix:code}.

\begin{minipage}{0.49\textwidth}
\begin{algorithm}[H]
\scriptsize 
\caption{\name{} Step 1: Attn. Transfer}
\label{alg:lolcats_stage_1}
\begin{algorithmic}[1]
\Require{Pretrained Transformer with $M$ attn.~layers; input tokens $\bm{u}$}
\Require{Linear attn.~feature map params.~$\{\phi_q^m, \phi_k^m, \gamma^m : m \in [M]\}$, window size $w$; layer block-size $b$, learning rate $\alpha$}
    \State Freeze all Transformer parameters
    \State \textbf{Initialize} block-wise losses $\{\ell_i \gets 0 : i \in [M /\!\!/ b] \}$ 
    \State Compute initial attn. input $\bm{x}^1 \gets \texttt{embed}(\bm{u})$
    \For{attn. layer {$m \in [M]$}}  \Comment \textbf{Compute attentions}
        \State $\bm{q},\bm{k},\bm{v} = \bm{x}^m\bm{W}_q^m, \bm{x}^m\bm{W}_k^m, 
        \bm{x}^m\bm{W}_v^m$
        \State $\bm{y}^m = \texttt{softmax\_attn}(\bm{q},\bm{k},\bm{v})$ (Eq.~\ref{eq:causal_attention}) \Comment (No grad)
        \State $\hat{\bm{y}}^m = \texttt{linear\_attn}(\bm{q},\bm{k},\bm{v}, w, \phi_q^m, \phi_k^m, \gamma^m)$ \hfill (Eq.~\ref{eq:hybrid_linear_swa})
        \State $\ell_{m/\!\!/b} \gets \ell_{m/\!\!/b} + \ell_{\text{MSE}}(\hat{\bm{y}}^m, \bm{y}^m)$ \hfill (Eq.~\ref{eq:mse_loss_blocks})
        \State $\bm{x}^{m+1} = \texttt{mlp}^m(\bm{y}^m\bm{W}_o^m)$ \Comment (Teacher-force next layer)
    \EndFor
    \For{attn. layer {$m \in [M]$}} \Comment \textbf{Update feature maps}
        \For{weights {$\theta \in \{\phi_q^m, \phi_k^m, \gamma^m\}$}}
            \State Update $\theta \gets \theta - \alpha \frac{\partial}{ \partial \theta}\ell_{m/\!\!/b}$ 
        \EndFor
    \EndFor
\end{algorithmic}
\end{algorithm}
\end{minipage}
\hfill
\begin{minipage}{0.49\textwidth}
\begin{algorithm}[H]
\scriptsize 
\caption{\name{} Step 2: LoRA Adjust}
\label{alg:lolcats_stage_2}
\begin{algorithmic}[1]
\Require{\textit{Linearized} Transformer with $M$ attn. layers; input tokens $\bm{u}$}
\Require{\textit{Trained} feature map params.~$\{\phi_q^m, \phi_k^m, \gamma^m : m \in [M]\}$, window size $w$; learning rate $\alpha$}
    \For{attn. layer {$m \in [M]$}} \Comment (Add LoRA weights)
        \Statex\;\;\;\; \textbf{Init.}~LoRA $\bm{A}^m, \bm{B}^m$ for each $\bm{W}_q^m,\bm{W}_k^m,\bm{W}_v^m,\bm{W}_o^m$
    \EndFor
    \State Compute initial attn. input $\bm{x}^1 \gets \texttt{embed}(\bm{u})$
    \For{attn. layer {$m \in [M]$}}  \Comment \textbf{Compute attentions}
        \State $\bm{q} = \bm{x}^m (\bm{W}_q^m + \bm{B}_q^m\bm{A}_q^m)$ \Comment (LoRA forward pass)
        \State $\bm{k} = \bm{x}^m (\bm{W}_k^m + \bm{B}_k^m\bm{A_k}^m)$
        \State $\bm{v} = \bm{x}^m (\bm{W}_v^m + \bm{B}_v^m\bm{A}_v^m)$
        
        \State $\hat{\bm{y}}^m = \texttt{linear\_attn}(\bm{q},\bm{k},\bm{v}, w, \phi_q^m, \phi_k^m, \gamma^m)$ \hfill (Eq.~\ref{eq:hybrid_linear_swa})
        \State $\bm{x}^{m+1} = \texttt{mlp}^m(\bm{y}^m (\bm{W}_o^m + \bm{B}_o^m\bm{A}_o^m))$ 
    \EndFor
    \State Compute next-token pred $\hat{\bm{u}} = \texttt{lm\_head}(\bm{x}^M)$ \Comment \textbf{Train LoRA}
    \State Compute sample loss $\ell = \ell_{\text{CrossEnt}}(\hat{\bm{u}}_{0:n-1}, \bm{u}_{1:n})$ 
    
        \State Update $\theta \gets \theta - \alpha \frac{\partial \ell}{ \partial \theta}$ for all $\theta$ in all $\bm{A},\bm{B}$ 
\end{algorithmic}
\end{algorithm}
\end{minipage}

\section{Experiments}
\label{sec:results}
\vspace{-0.25cm}

Through experiments, we study: (1) if \name{} linearizes LLMs with higher quality than existing subquadratic alternatives and linearizations, and higher generation efficiency than original Transformers (Sec.~\ref{sec:main_results});  (2) how ablations on attention transfer loss, subquadratic architecture, and parameter and token counts impact LLM quality (Sec.~\ref{sec:results_ablations}); and (3) how \name{}' quality and efficiency holds up to 70B and 405B LLMs by linearizing the complete Llama 3.1 family
(Sec.~\ref{sec:chungus_amongus}).

\subsection{Main Results: \name{} Efficiently Recovers Quality in Linearized LLMs}
\label{sec:main_results}
\vspace{-0.125cm}

In our main evaluation, we linearize the popular base Llama 3 8B~\citep{AI_2024} and Mistral 7B~\citep{jiang2023mistral} LLMs. We first test if \name{} can efficiently create high-quality subquadratic LLMs from strong base Transformers, comparing to existing linearized LLMs from prior methods. We also test if \name{} can create subquadratic LLMs that outperform modern Transformer alternatives pretrained from scratch. For space, we defer linearizing training details to Appendix~\ref{appendix:experimental_details}.



\begin{table}[t]
\begin{center}
\resizebox{1\linewidth}{!}{
\begin{tabular}{@{}l|c|cccccc|cc@{}}
\toprule
\multicolumn{1}{c|}{Model}                       & \begin{tabular}[c]{@{}c@{}}Training\\ Tokens (B)\end{tabular} & PiQA                                         & ARC-e                                        & \begin{tabular}[c]{@{}c@{}}ARC-c\\ (norm)\end{tabular} & \begin{tabular}[c]{@{}c@{}}HellaSwag\\ (norm)\end{tabular} & \begin{tabular}[c]{@{}c@{}}Wino-\\ grande\end{tabular} & \begin{tabular}[c]{@{}c@{}}MMLU \\ (5-shot)\end{tabular} & Avg.                                 & \begin{tabular}[c]{@{}c@{}}Avg. \\ (no MMLU)\end{tabular} \\ 
\midrule 

\rowcolor[HTML]{FFDECA} 
Mistral 7B                & -                    & 82.1          & 80.9          & 53.8          & 81.0          & 74.0          & 62.4          & 72.4          & 74.4          \\
Mistral 7B SUPRA          & 100                  & 80.4          & 75.9          & 45.8          & 77.1          & 70.3          & 34.2          & 64.0          & 69.9          \\

Mistral 7B LoLCATs (Ours) & \textbf{0.04} &	\textbf{81.5}	& \textbf{81.7}	& \textbf{54.9}	& \textbf{80.7}	& \textbf{74.0}	&\textbf{51.4}	& \textbf{70.7}	& \textbf{74.5} \\

\midrule
\rowcolor[HTML]{ECF4FF} 
Llama 3 8B                & -                    & 79.9          & 80.1          & 53.3          & 79.1          & 73.1          & 66.6          & 72.0          & 73.1          \\
Mamba2-Llama 3            & 20                  & 76.8          & 74.1          & 48.0          & 70.8          & 58.6          & 43.2          & 61.9          & 65.6          \\
Mamba2-Llama 3, 50\% Attn.       & 20 & \textbf{81.5} & 78.8          & \textbf{58.2} & 79.5          & 71.5          & \textbf{56.7} & \textbf{71.0}          & 73.9          \\
Llama 3 8B Hedgehog       & \textbf{0.04}                 & 77.4          & 71.1          & 40.6          & 66.5          & 54.3          & 24.2          & 55.7          & 62.0          \\


Llama 3 8B LoLCATs (Ours) & \textbf{0.04}                 & 80.9          & \textbf{81.7} & 54.9          & \textbf{79.7} & \textbf{74.1} & 52.8          & 70.7 & \textbf{74.2} \\ \bottomrule

\end{tabular}
}
\end{center}
\vspace{-0.25cm}
\caption{\small \textbf{\name{} comparison among linearized 7B+ LLMs}. Among linearized 7B+ LLMs, \name{}-linearized Mistral 7B and Llama 3 8B consistently achieve best or 2nd-best performance on LM Eval tasks.
\name{} closes the Transformer quality gap by 
79.8\% (Mistral 7B) and 86.6\% (Llama 3 8B) 
(average over all tasks; numbers except Hedgehog cited from original works), despite only using 40M tokens to linearize.
}
\label{table:main_results_linearize}
\vspace{-0.25cm}
\end{table}

\begin{table}[t]
\begin{center}
\resizebox{1\linewidth}{!}{
\begin{tabular}{@{}l|c|cccccc|cc@{}}
\toprule
Model                     & Tokens (B)            & PiQA                 & ARC-e             & \begin{tabular}[c]{@{}c@{}}ARC-c\\ (acc. norm)\end{tabular} & \begin{tabular}[c]{@{}c@{}}HellaSwag\\ (acc. norm)\end{tabular} & Winogrande           & \begin{tabular}[c]{@{}c@{}}MMLU \\ (5-shot)\end{tabular} & \begin{tabular}[c]{@{}c@{}}Avg. \\ (w MMLU)\end{tabular} & \begin{tabular}[c]{@{}c@{}}Avg. \\ (no MMLU)\end{tabular} \\ \midrule

\rowcolor[HTML]{EFEFEF} \textbf{Transformer}      & \multicolumn{1}{l|}{} & \multicolumn{1}{l}{} & \multicolumn{1}{l}{} & \multicolumn{1}{l}{}                                        & \multicolumn{1}{l}{}                                            & \multicolumn{1}{l}{} & \multicolumn{1}{l|}{}                                    & \multicolumn{1}{l}{}                                     & \multicolumn{1}{l}{}                                      \\
Gemma 7B                  & 6000                  & 81.9                 & 81.1                 & 53.2                                                        & 80.7                                                            & 73.7                 & 62.9                                                     & 72.3                                                     & 74.1                                                      \\
Mistral 7B                & 8000*                 & 82.1                 & 80.9                 & 53.8                                                        & 81.0                                                            & 74.0                 & 62.4                                                     & 72.4                                                     & 74.4                                                      \\
Llama 3 8B                & 15000                 & 79.9                 & 80.1                 & 53.3                                                        & 79.1                                                            & 73.1                 & 66.6                                                     & 72.0                                                     & 73.1                                                      \\ \midrule
\rowcolor[HTML]{EFEFEF} \textbf{Subquadratic}     & \multicolumn{1}{l|}{} & \multicolumn{1}{l}{} & \multicolumn{1}{l}{} & \multicolumn{1}{l}{}                                        & \multicolumn{1}{l}{}                                            & \multicolumn{1}{l}{} & \multicolumn{1}{l|}{}                                    & \multicolumn{1}{l}{}                                     & \multicolumn{1}{l}{}                                      \\
Mamba 7B                  & 1200                  & 81.0                 & 77.5                 & 46.7                                                        & 77.9                                                            & 71.8                 & 33.3                                                     & 64.7                                                     & 71.0                                                      \\
RWKV-6 World v2.1 7B      & 1420                  & 78.7                 & 76.8                 & 46.3                                                        & 75.1                                                            & 70.0                 & -                                                        & 69.4                                                     & 69.4                                                      \\
TransNormerLLM 7B         & 1400                  & 80.1                 & 75.4                 & 44.4                                                        & 75.2                                                            & 66.1                 & 43.1                                                     & 64.1                                                     & 68.2                                                      \\
Hawk 7B                   & 300                   & 80.0                 & 74.4                 & 45.9                                                        & 77.6                                                            & 69.9                 & 35.0                                                     & 63.8                                                     & 69.6                                                      \\
Griffin 7B                & 300                   & 81.0                 & 75.4                 & 47.9                                                        & 78.6                                                            & 72.6                 & 39.3                                                     & 65.8                                                     & 71.1                                                      \\ 
\rowcolor[HTML]{EFEFEF} \textbf{Hybrid Softmax}   & \multicolumn{1}{l|}{} & \multicolumn{1}{l}{} & \multicolumn{1}{l}{} & \multicolumn{1}{l}{}                                        & \multicolumn{1}{l}{}                                            & \multicolumn{1}{l}{} & \multicolumn{1}{l|}{}                                    & \multicolumn{1}{l}{}                                     & \multicolumn{1}{l}{}                                      \\
StripedHyena-Nous-7B      & -                     & 78.8                 & 77.2                 & 40.0                                                        & 76.4                                                            & 66.4                 & 26.0                                                     & 60.8                                                     & 67.8                                                      \\
Zamba 7B                  & 1000                  & 81.4                 & 74.5                 & 46.6                                                        & 80.2                                                            & \textbf{76.4}        & \textbf{57.7}                                            & 69.5                                                     & 71.8                                                      \\ 
\rowcolor[HTML]{ECF4FF} \textbf{Linearized}       & \multicolumn{1}{l|}{} & \multicolumn{1}{l}{} & \multicolumn{1}{l}{} & \multicolumn{1}{l}{}                                        & \multicolumn{1}{l}{}                                            & \multicolumn{1}{l}{} & \multicolumn{1}{l|}{}                                    & \multicolumn{1}{l}{}                                     & \multicolumn{1}{l}{}                                      \\
Mistral 7B LoLCATs (Ours) & \textbf{0.04}         & \textbf{81.5}        & \textbf{81.7}        & \textbf{54.9}                                               & \textbf{80.7}                                                   & 74.0                 & 51.4                                                     & \textbf{70.7}                                            & \textbf{74.5}                                             \\
Llama 3 8B LoLCATs (Ours) & \textbf{0.04}         & 80.9                 & \textbf{81.7}        & \textbf{54.9}                                               & 79.7                                                            & 74.1                 & 52.8                                                     & \textbf{70.7}                                            & 74.2                                                      \\ \bottomrule
\end{tabular}
}
\end{center}
\vspace{-0.25cm}
\caption{\small \textbf{\name{} comparison to pretrained subquadratic LLMs}. \name{}-linearized Mistral 7B and Llama 3 8B outperform pretrained Transformer alternatives by
1.2 to 9.9 points (Avg.), 
only training 0.2\% of their parameter counts on 0.013 to 0.003\% of their training token counts. 
$^*$Reported in \citet{mercat2024linearizing}.}
\label{table:main_results}
\vspace{-0.5cm}
\end{table}

In Table~\ref{table:main_results}, we report results on six popular LM Evaluation Harness (LM Eval) tasks~\citep{eval-harness}.
Compared to recent linearizing methods, \name{} significantly improves quality and training efficiency across tasks and LLMs. On quality, \name{} closes 
79.8\% and 86.6\%
of the Transformer-linearizing gap for Mistral 7B and Llama 3 8B respectively, notably improving 5-shot MMLU by 
60.9\% and 40.9\% 
over next best fully subquadratic models (17.2 and 9.6 points). On efficiency, we achieve these results while only training $<$0.2\% of model parameters via LoRA versus prior full-parameter training. We also only use 40M tokens versus the prior 20~--~100B (a 500~--~2500$\times$ improvement in ``tokens-to-model'' efficiency).
Among all 7B LLMs, \name{}-linearized LLMs further outperform strong subquadratic Transformer alternatives, representing RNNs or linear attentions (RWKV-v6~\citep{peng2024eagle}, Hawk, Griffin~\citep{de2024griffin}, TransNormer~\citep{qin2023scaling}), state-space models (Mamba~\citep{gu2023mamba}), and hybrid architectures with some full attention (StripedHyena~\citep{stripedhyena}, Zamba~\citep{glorioso2024zamba}). 

\vspace{-0.125cm}
\subsection{\name{} Component Properties and Ablations}
\vspace{-0.125cm}
\label{sec:results_ablations}
We next validate that \name{} enables subquadratic efficiency, and study how each of \name{}' components contribute to these quality gains. We include additional ablations in Appendix~\ref{appendix:additional_experiments}.

\begin{wrapfigure}{r}{0.55\textwidth}
\vspace{-0.75cm}
    \centering
    \begin{subfigure}[b]{0.27\textwidth}
        \centering
        \includegraphics[width=\textwidth]{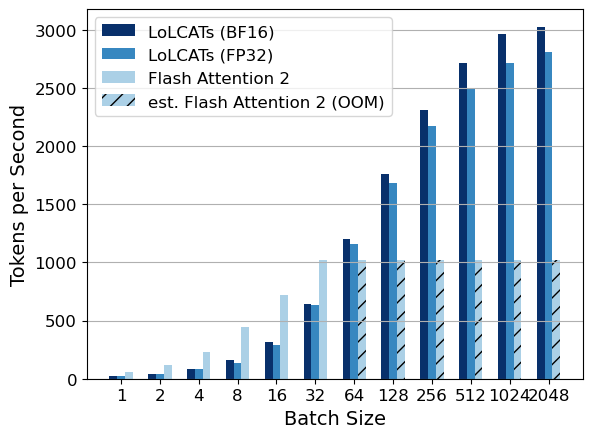}
        \label{fig:generation_throughput}
    \end{subfigure}
    \vspace{-0.12cm}
    \begin{subfigure}[b]{0.27\textwidth}
        \centering
        \includegraphics[width=\textwidth]{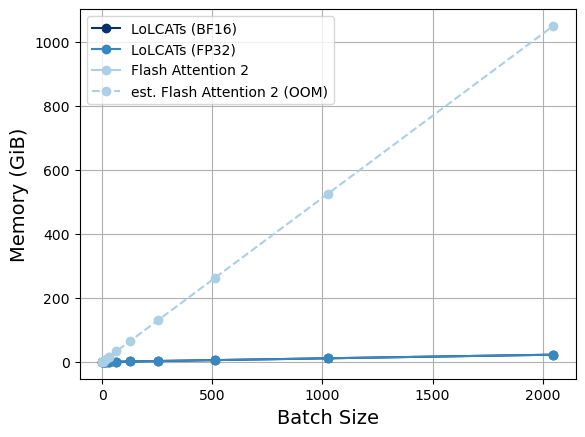}
        \label{fig:generation_memory}
    \end{subfigure}
    \vspace{-1cm}
    \caption{\small \textbf{ Generation Efficiency}, Llama 3 8B.}
    \label{fig:bm_efficiency}
    \vspace{-0.5cm}
\end{wrapfigure}

\paragraph{Subquadratic Generation Throughput and Memory.} We measure generation throughput and memory of \name{} LLMs, validating that linearizing LLMs can significantly improve their generation efficiency. We use the popular Llama 3 8B HuggingFace model\footnote{\url{https://huggingface.co/meta-llama/Meta-Llama-3-8B}}, and compare \name{} implemented in HuggingFace Transformers with the supported FlashAttention-2 (FA2)~\citep{dao2023flashattention}. We benchmark \name{} with Hedgehog feature map and linear + sliding window attention in FP32 and BF16 on one 80GB H100.
Fig.~\ref{fig:bm_efficiency} reports scaling batch size on throughput (left) and memory (right). We measure throughput as $($generated tokens $\times$ batch size $/$ total time$)$, with 128-token prompts and 4096-token generations. With larger batch size, \name{}-linearized LLMs achieve higher throughput than FA2. This corresponds with lower memory, where FA2 exceeds memory at batch size 64. Meanwhile, \name{} supports 3000 tokens / second at batch size 2048 
(Fig.~\ref{fig:bm_efficiency} left),
 with fixed ``KV state'' vs.~softmax attention's growing KV cache  
(Fig.~\ref{fig:bm_efficiency} right). 

\paragraph{Ablations.}
We study how attention transfer and linear + sliding window attention in \name{} contribute to downstream linearized Llama 3 8B performance  
(Table~\ref{table:main_results_ablations}). We start with prior linear attentions (Hedgehog,~\cite{zhang2024the}; T2R,~\cite{kasai-etal-2021-finetuning}), using the prior linearizing approach that swaps attentions and finetunes the model to predict next tokens~\citep{mercat2024linearizing}. We then add (i) attention transfer, (ii) sliding window attentions, or (iii) both. On average LM Eval score and 5-shot MMLU accuracy, \name{}' default performs best across feature maps.

\begin{table}[t]
\begin{center}
\resizebox{\linewidth}{!}{
\begin{tabular}{@{}lccccc@{}}
\toprule
Feature Map                & \begin{tabular}[c]{@{}c@{}}LM Eval \\ Metric\end{tabular} & \begin{tabular}[c]{@{}c@{}}Swap \&\\ Finetune\end{tabular} & \begin{tabular}[c]{@{}c@{}}+Attention \\ Transfer\end{tabular} & \cellcolor[HTML]{ECF4FF}\begin{tabular}[c]{@{}c@{}}+Sliding Window,\\ +Attention Transfer \end{tabular} & \begin{tabular}[c]{@{}c@{}}+ Sliding Window, \\ No Attention Transfer\end{tabular} \\ \midrule
                           & Average    & 44.20                                                      & 55.32                                                          & \cellcolor[HTML]{ECF4FF}\textbf{70.66}                                                      & {\ul 68.78}                                                                       \\
\multirow{-2}{*}{Hedgehog} & MMLU (5-shot)  & 23.80                                                      & 23.80                                                          & \cellcolor[HTML]{ECF4FF}\textbf{52.77}                                                      & 45.80                                                                             \\ \midrule
                           & Average    & 38.84                                                      & {\ul 54.83}                                                    & \textbf{68.28}                                                                              & 39.52                                                                             \\
\multirow{-2}{*}{T2R}      & MMLU (5-shot)   & 23.20                                                      & 23.10                                                          & \textbf{40.70}                                                                              & 23.80                                                                             \\ \bottomrule
\end{tabular}
}
\end{center}
\vspace{-0.25cm}
\caption{\textbf{\name{} component ablations}, linearizing Llama 3 8B over 1024-token sequences. \name{} default shaded. Across Hedgehog and T2R feature maps, \name{}' attention transfer and sliding window increasingly improve linearized LLM quality. 
Full task results in App.~\ref{appendix:add_exp_main_ablations_full}.
}
\label{table:main_results_ablations}
\vspace{-0.25cm}
\end{table}
\vspace{-0.125cm}
\subsection{Scaling Up Linearizing to 70B and 405B LLMs}
\vspace{-0.125cm}
\label{sec:chungus_amongus}
We finally use \name{} to scale up linearizing to Llama 3.1 70B and 405B models. 
In Table~\ref{table:chungus_results}, \name{} provides the first practical solution for linearizing larger LLMs, achieving significant quality improvements over prior linearizing approaches of swapping in attentions and finetuning~\citep{mercat2024linearizing}. With the same linear + sliding window layer, \name{} gets +39.0 points in 5-shot MMLU accuracy on Llama 3.1 70B, and +38.3 on Llama 3.1 405B.
These results highlight \name{}' ability to linearize large-scale models with greater efficiency and improved performance, showing for the first time we can scale up linearizing to 70B+ LLMs.

\begin{table}[t]
\begin{center}
\resizebox{0.99\linewidth}{!}{
\begin{tabular}{@{}lcccccc@{}}
\toprule
                       & PiQA           & ARC Easy       & ARC Challenge  & HellaSwag      & WinoGrande     & MMLU (5-shot)  \\ 
                       & acc            & acc            & (acc norm)     & (acc norm)     & acc            & acc            \\ \midrule
\rowcolor[HTML]{EFEFEF} 
Llama 3.1 8B          & 79.87          & 81.52          & 53.58          & 79.01          & 73.48          & 66.11          \\
Linearized, no attn. transfer & 78.67	& 78.11	& 49.83	& 77.83 & 	68.51 & 	51.44          \\

\name{} (Ours)                & \textbf{80.96} & \textbf{82.37} & \textbf{54.44} & \textbf{79.07} & \textbf{69.69} & \textbf{54.88} \\ \midrule

\rowcolor[HTML]{EFEFEF} 
Llama 3.1 70B          & 83.10          & 87.30          & 60.60          & 85.00          & 79.60          & 78.80          \\

Linearized, no attn. transfer & 81.99 &	80.89 &	54.44 &	82.29 &	71.19 &	28.74 \\
\name{} (Ours)                & \textbf{82.10} & \textbf{84.98} & \textbf{60.50} & \textbf{84.62} & \textbf{73.72} & \textbf{67.70} \\ \midrule
\rowcolor[HTML]{EFEFEF} 
Llama 3.1 405B         & 85.58          & 87.58          & 66.21          & 87.13          & 79.40          & 82.98          \\
Linearized, no attn. transfer & 84.44          & 86.62          & 64.33          & 86.19          & 79.87          & 33.86          \\
\name{} (Ours)                & \textbf{85.58} & \textbf{88.80} & \textbf{67.75} & \textbf{87.41} & \textbf{80.35} & \textbf{72.20} \\ \bottomrule
\end{tabular}
}
\end{center}
\vspace{-0.25cm}
\caption{\textbf{Linearizing Llama 3.1 8B, 70B, and 405B}. Among the first linearized 70B and 405B LLMs (via low-rank linearizing), \name{} significantly improves zero- and few-shot quality.
}
\label{table:chungus_results}
\vspace{-0.5cm}
\end{table}

\vspace{-3mm}
\section{Conclusion}
\label{sec:conclusion}
We propose \name{}, an efficient LLM linearizing method that (1) trains attention analogs---such as linear attentions and linear attention + sliding window hybrids---to approximate an LLM's self-attentions, before (2) swapping the attentions and only finetuning the replacing attentions with LoRA. 
We exploit the fidelity between these attention analogs and softmax attention, where we reduce the problem of linearizing LLMs to learning to approximate softmax attention in a subquadratic analog. 
Furthermore, we demonstrate that via an MSE-based attention output-matching loss, we \emph{are able} to train such attention analogs to approximate the ``ground-truth'' softmax attentions in practice. 
On popular zero-shot LM Evaluation harness benchmarks and 5-shot MMLU, we find this enables producing high-quality, high-inference efficiency LLMs that outperform prior Transformer alternatives while only updating 0.2\% of model parameters and requiring 0.003\% of the training tokens to achieve similar quality with LLM pretraining. Our findings significantly improve linearizing quality and accessibility, allowing us to create the first linearized 70B and 405B LLMs.

\subsubsection*{Limitations and Future Work}

While we focus on studying how to enable high quality yet highly efficient LLM linearizing with simple linear attentions, we note several areas for additional evaluation in both subquadratic capabilities and architectures. On subquadratic capabilities,  by replacing each attention layer alternative, we eliminate the need to manage growing key-value (KV) caches and their associated memory overheads. However, it remains to be seen what kinds of capabilities we can enable with this cheaper inference, \eg{} if linearized models can exploit  quality-improving inference scaling laws suggested by recent works~\citep{brown2024large,snell2024scaling}. Under a different motivation, while layers like linear attention achieve greater efficiency gains over softmax attention  when processing longer contexts, we leave studying how low-rank linearizing applies to such long context scenarios as a motivated direction for future work. Finally, while we stick to ``vanilla'' linear + sliding window attentions in \name{}, many more recent subquadratic architectures improve linear attention quality with additional factors such as decay terms~\citep{sun2023retentive} and additional gating~\citep{yang2023gated}. Studying whether attention transfer and low-rank linearizing can help scale up these additional attention analogs is an interesting line of future work.

\subsubsection*{Ethics Statement}

Our work deals with improving the efficiency of open-weight models. While promising for beneficial applications, increasing their accessibility also raises concerns about potential misuse. Bad actors could leverage our technique to develop LLMs capable of generating harmful content, spreading misinformation, or enabling other malicious activities. We focus primarily on base models, but acknowledge that linearizing could also be used on instruction-tuned LLMs; research on whether linearizing preserves guardrails is still an open question. We acknowledge the risks and believe in the responsible development and deployment of efficient and widely accessible models.

\subsubsection*{Reproducibility}
We include experimental details in Appendix~\ref{appendix:experimental_details}, and further implementation details with sample code for linearizing architectures and training in Appendix~\ref{appendix:code}. Our code is also available at \url{https://github.com/HazyResearch/lolcats}

\subsection*{Acknowledgements}
We thank Mayee Chen, Ben Viggiano, Gautam Machiraju, Dan Fu, Sabri Eyuboglu, Tri Dao, and anonymous reviewers for helpful discussions on linear attention and paper feedback. 
We gratefully acknowledge the support of NIH under No. U54EB020405 (Mobilize), NSF under Nos. CCF2247015 (Hardware-Aware), CCF1763315 (Beyond Sparsity), CCF1563078 (Volume to Velocity), and 1937301 (RTML); US DEVCOM ARL under Nos. W911NF-23-2-0184 (Long-context) and W911NF-21-2-0251 (Interactive Human-AI Teaming); ONR under Nos. N000142312633 (Deep Signal Processing); Stanford HAI under No. 247183; NXP, Xilinx, LETI-CEA, Intel, IBM, Microsoft, NEC, Toshiba, TSMC, ARM, Hitachi, BASF, Accenture, Ericsson, Qualcomm, Analog Devices, Google Cloud, Salesforce, Total, the HAI-GCP Cloud Credits for Research program,  the Stanford Data Science Initiative (SDSI), and members of the Stanford DAWN project: Meta, Google, and VMWare. The U.S. Government is authorized to reproduce and distribute reprints for Governmental purposes notwithstanding any copyright notation thereon. Any opinions, findings, and conclusions or recommendations expressed in this material are those of the authors and do not necessarily reflect the views, policies, or endorsements, either expressed or implied, of NIH, ONR, or the U.S. Government.

\bibliography{main}

\begin{thebibliography}{75}
\providecommand{\natexlab}[1]{#1}
\providecommand{\url}[1]{\texttt{#1}}
\expandafter\ifx\csname urlstyle\endcsname\relax
  \providecommand{\doi}[1]{doi: #1}\else
  \providecommand{\doi}{doi: \begingroup \urlstyle{rm}\Url}\fi

\bibitem[kai()]{kaiokendev.github.io}
URL \url{https://kaiokendev.github.io/context}.

\bibitem[AI(2024)]{AI_2024}
Mistral AI.
\newblock Mixtral of experts.
\newblock \emph{Mistral AI | Frontier AI in your hands}, May 2024.
\newblock URL \url{https://mistral.ai/news/mixtral-of-experts/}.

\bibitem[AI@Meta(2024{\natexlab{a}})]{llama3modelcard}
AI@Meta.
\newblock Llama 3 model card.
\newblock 2024{\natexlab{a}}.
\newblock URL \url{https://github.com/meta-llama/llama3/blob/main/MODEL_CARD.md}.

\bibitem[AI@Meta(2024{\natexlab{b}})]{meta2024llama32}
AI@Meta.
\newblock Llama 3.2: Revolutionizing edge ai and vision with open, customizable models, Sep 2024{\natexlab{b}}.
\newblock URL \url{https://ai.meta.com/blog/llama-3-2-connect-2024-vision-edge-mobile-devices}.

\bibitem[Arora et~al.(2024)Arora, Eyuboglu, Zhang, Timalsina, Alberti, Zinsley, Zou, Rudra, and R{\'e}]{arora2024simple}
Simran Arora, Sabri Eyuboglu, Michael Zhang, Aman Timalsina, Silas Alberti, Dylan Zinsley, James Zou, Atri Rudra, and Christopher R{\'e}.
\newblock Simple linear attention language models balance the recall-throughput tradeoff.
\newblock \emph{arXiv preprint arXiv:2402.18668}, 2024.

\bibitem[Beck et~al.(2024)Beck, P{\"o}ppel, Spanring, Auer, Prudnikova, Kopp, Klambauer, Brandstetter, and Hochreiter]{beck2024xlstm}
Maximilian Beck, Korbinian P{\"o}ppel, Markus Spanring, Andreas Auer, Oleksandra Prudnikova, Michael Kopp, G{\"u}nter Klambauer, Johannes Brandstetter, and Sepp Hochreiter.
\newblock xlstm: Extended long short-term memory.
\newblock \emph{arXiv preprint arXiv:2405.04517}, 2024.

\bibitem[Beltagy et~al.(2020)Beltagy, Peters, and Cohan]{beltagy2020longformer}
Iz~Beltagy, Matthew~E Peters, and Arman Cohan.
\newblock Longformer: The long-document transformer.
\newblock \emph{arXiv preprint arXiv:2004.05150}, 2020.

\bibitem[Bick et~al.(2024)Bick, Li, Xing, Kolter, and Gu]{bick2024transformers}
Aviv Bick, Kevin~Y Li, Eric~P Xing, J~Zico Kolter, and Albert Gu.
\newblock Transformers to ssms: Distilling quadratic knowledge to subquadratic models.
\newblock \emph{arXiv preprint arXiv:2408.10189}, 2024.

\bibitem[Biderman et~al.(2023)Biderman, Schoelkopf, Anthony, Bradley, O’Brien, Hallahan, Khan, Purohit, Prashanth, Raff, et~al.]{biderman2023pythia}
Stella Biderman, Hailey Schoelkopf, Quentin~Gregory Anthony, Herbie Bradley, Kyle O’Brien, Eric Hallahan, Mohammad~Aflah Khan, Shivanshu Purohit, USVSN~Sai Prashanth, Edward Raff, et~al.
\newblock Pythia: A suite for analyzing large language models across training and scaling.
\newblock In \emph{International Conference on Machine Learning}, pp.\  2397--2430. PMLR, 2023.

\bibitem[Brown et~al.(2024)Brown, Juravsky, Ehrlich, Clark, Le, R{\'e}, and Mirhoseini]{brown2024large}
Bradley Brown, Jordan Juravsky, Ryan Ehrlich, Ronald Clark, Quoc~V Le, Christopher R{\'e}, and Azalia Mirhoseini.
\newblock Large language monkeys: Scaling inference compute with repeated sampling.
\newblock \emph{arXiv preprint arXiv:2407.21787}, 2024.

\bibitem[Chen et~al.(2021{\natexlab{a}})Chen, Dao, Winsor, Song, Rudra, and R{\'e}]{chen2021scatterbrain}
Beidi Chen, Tri Dao, Eric Winsor, Zhao Song, Atri Rudra, and Christopher R{\'e}.
\newblock Scatterbrain: Unifying sparse and low-rank attention approximation.
\newblock \emph{arXiv preprint arXiv:2110.15343}, 2021{\natexlab{a}}.

\bibitem[Chen et~al.(2023{\natexlab{a}})Chen, Wong, Chen, and Tian]{chen2023extending}
Shouyuan Chen, Sherman Wong, Liangjian Chen, and Yuandong Tian.
\newblock Extending context window of large language models via positional interpolation.
\newblock \emph{arXiv preprint arXiv:2306.15595}, 2023{\natexlab{a}}.

\bibitem[Chen et~al.(2021{\natexlab{b}})Chen, Zeng, Ji, and Yang]{chen2021skyformer}
Yifan Chen, Qi~Zeng, Heng Ji, and Yun Yang.
\newblock Skyformer: Remodel self-attention with gaussian kernel and nystr{\textbackslash}''om method.
\newblock In A.~Beygelzimer, Y.~Dauphin, P.~Liang, and J.~Wortman Vaughan (eds.), \emph{Advances in Neural Information Processing Systems}, 2021{\natexlab{b}}.
\newblock URL \url{https://openreview.net/forum?id=pZCYG7gjkKz}.

\bibitem[Chen et~al.(2023{\natexlab{b}})Chen, Qian, Tang, Lai, Liu, Han, and Jia]{chen2023longlora}
Yukang Chen, Shengju Qian, Haotian Tang, Xin Lai, Zhijian Liu, Song Han, and Jiaya Jia.
\newblock Longlora: Efficient fine-tuning of long-context large language models.
\newblock \emph{arXiv preprint arXiv:2309.12307}, 2023{\natexlab{b}}.

\bibitem[Choromanski et~al.(2020)Choromanski, Likhosherstov, Dohan, Song, Gane, Sarlos, Hawkins, Davis, Mohiuddin, Kaiser, et~al.]{choromanski2020rethinking}
Krzysztof Choromanski, Valerii Likhosherstov, David Dohan, Xingyou Song, Andreea Gane, Tamas Sarlos, Peter Hawkins, Jared Davis, Afroz Mohiuddin, Lukasz Kaiser, et~al.
\newblock Rethinking attention with performers.
\newblock \emph{arXiv preprint arXiv:2009.14794}, 2020.

\bibitem[Computer(2023)]{together2023redpajama}
Together Computer.
\newblock Redpajama: An open source recipe to reproduce llama training dataset, 2023.
\newblock URL \url{https://github.com/togethercomputer/RedPajama-Data}.

\bibitem[Dao(2023)]{dao2023flashattention}
Tri Dao.
\newblock Flashattention-2: Faster attention with better parallelism and work partitioning.
\newblock \emph{arXiv preprint arXiv:2307.08691}, 2023.

\bibitem[Dao \& Gu(2024)Dao and Gu]{dao2024transformers}
Tri Dao and Albert Gu.
\newblock Transformers are ssms: Generalized models and efficient algorithms through structured state space duality.
\newblock \emph{arXiv preprint arXiv:2405.21060}, 2024.

\bibitem[Dao et~al.(2022)Dao, Fu, Ermon, Rudra, and R{\'e}]{dao2022flashattention}
Tri Dao, Dan Fu, Stefano Ermon, Atri Rudra, and Christopher R{\'e}.
\newblock Flashattention: Fast and memory-efficient exact attention with io-awareness.
\newblock \emph{Advances in Neural Information Processing Systems}, 35:\penalty0 16344--16359, 2022.

\bibitem[De et~al.(2024)De, Smith, Fernando, Botev, Cristian-Muraru, Gu, Haroun, Berrada, Chen, Srinivasan, et~al.]{de2024griffin}
Soham De, Samuel~L Smith, Anushan Fernando, Aleksandar Botev, George Cristian-Muraru, Albert Gu, Ruba Haroun, Leonard Berrada, Yutian Chen, Srivatsan Srinivasan, et~al.
\newblock Griffin: Mixing gated linear recurrences with local attention for efficient language models.
\newblock \emph{arXiv preprint arXiv:2402.19427}, 2024.

\bibitem[Devlin et~al.(2018)Devlin, Chang, Lee, and Toutanova]{devlin2018bert}
Jacob Devlin, Ming-Wei Chang, Kenton Lee, and Kristina Toutanova.
\newblock Bert: Pre-training of deep bidirectional transformers for language understanding.
\newblock \emph{arXiv preprint arXiv:1810.04805}, 2018.

\bibitem[Dubey et~al.(2024)Dubey, Jauhri, Pandey, Kadian, Al-Dahle, Letman, Mathur, Schelten, Yang, Fan, et~al.]{dubey2024llama}
Abhimanyu Dubey, Abhinav Jauhri, Abhinav Pandey, Abhishek Kadian, Ahmad Al-Dahle, Aiesha Letman, Akhil Mathur, Alan Schelten, Amy Yang, Angela Fan, et~al.
\newblock The llama 3 herd of models.
\newblock \emph{arXiv preprint arXiv:2407.21783}, 2024.

\bibitem[Gao et~al.(2023)Gao, Tow, Abbasi, Biderman, Black, DiPofi, Foster, Golding, Hsu, Le~Noac'h, Li, McDonell, Muennighoff, Ociepa, Phang, Reynolds, Schoelkopf, Skowron, Sutawika, Tang, Thite, Wang, Wang, and Zou]{eval-harness}
Leo Gao, Jonathan Tow, Baber Abbasi, Stella Biderman, Sid Black, Anthony DiPofi, Charles Foster, Laurence Golding, Jeffrey Hsu, Alain Le~Noac'h, Haonan Li, Kyle McDonell, Niklas Muennighoff, Chris Ociepa, Jason Phang, Laria Reynolds, Hailey Schoelkopf, Aviya Skowron, Lintang Sutawika, Eric Tang, Anish Thite, Ben Wang, Kevin Wang, and Andy Zou.
\newblock A framework for few-shot language model evaluation, 12 2023.
\newblock URL \url{https://zenodo.org/records/10256836}.

\bibitem[Gliwa et~al.(2019)Gliwa, Mochol, Biesek, and Wawer]{gliwa-etal-2019-samsum}
Bogdan Gliwa, Iwona Mochol, Maciej Biesek, and Aleksander Wawer.
\newblock {SAMS}um corpus: A human-annotated dialogue dataset for abstractive summarization.
\newblock In Lu~Wang, Jackie Chi~Kit Cheung, Giuseppe Carenini, and Fei Liu (eds.), \emph{Proceedings of the 2nd Workshop on New Frontiers in Summarization}, pp.\  70--79, Hong Kong, China, November 2019. Association for Computational Linguistics.
\newblock \doi{10.18653/v1/D19-5409}.
\newblock URL \url{https://aclanthology.org/D19-5409}.

\bibitem[Glorioso et~al.(2024)Glorioso, Anthony, Tokpanov, Whittington, Pilault, Ibrahim, and Millidge]{glorioso2024zamba}
Paolo Glorioso, Quentin Anthony, Yury Tokpanov, James Whittington, Jonathan Pilault, Adam Ibrahim, and Beren Millidge.
\newblock Zamba: A compact 7b ssm hybrid model.
\newblock \emph{arXiv preprint arXiv:2405.16712}, 2024.

\bibitem[Gu \& Dao(2023)Gu and Dao]{gu2023mamba}
Albert Gu and Tri Dao.
\newblock Mamba: Linear-time sequence modeling with selective state spaces.
\newblock \emph{arXiv preprint arXiv:2312.00752}, 2023.

\bibitem[Gu et~al.(2021)Gu, Goel, and R{\'e}]{gu2021efficiently}
Albert Gu, Karan Goel, and Christopher R{\'e}.
\newblock Efficiently modeling long sequences with structured state spaces.
\newblock \emph{arXiv preprint arXiv:2111.00396}, 2021.

\bibitem[Hendrycks et~al.(2020)Hendrycks, Burns, Basart, Zou, Mazeika, Song, and Steinhardt]{hendrycks2020measuring}
Dan Hendrycks, Collin Burns, Steven Basart, Andy Zou, Mantas Mazeika, Dawn Song, and Jacob Steinhardt.
\newblock Measuring massive multitask language understanding.
\newblock \emph{arXiv preprint arXiv:2009.03300}, 2020.

\bibitem[Hu et~al.(2021)Hu, Shen, Wallis, Allen-Zhu, Li, Wang, Wang, and Chen]{hu2021lora}
Edward~J Hu, Yelong Shen, Phillip Wallis, Zeyuan Allen-Zhu, Yuanzhi Li, Shean Wang, Lu~Wang, and Weizhu Chen.
\newblock Lora: Low-rank adaptation of large language models.
\newblock \emph{arXiv preprint arXiv:2106.09685}, 2021.

\bibitem[Jiang et~al.(2023)Jiang, Sablayrolles, Mensch, Bamford, Chaplot, Casas, Bressand, Lengyel, Lample, Saulnier, et~al.]{jiang2023mistral}
Albert~Q Jiang, Alexandre Sablayrolles, Arthur Mensch, Chris Bamford, Devendra~Singh Chaplot, Diego de~las Casas, Florian Bressand, Gianna Lengyel, Guillaume Lample, Lucile Saulnier, et~al.
\newblock Mistral 7b.
\newblock \emph{arXiv preprint arXiv:2310.06825}, 2023.

\bibitem[Kasai et~al.(2021)Kasai, Peng, Zhang, Yogatama, Ilharco, Pappas, Mao, Chen, and Smith]{kasai-etal-2021-finetuning}
Jungo Kasai, Hao Peng, Yizhe Zhang, Dani Yogatama, Gabriel Ilharco, Nikolaos Pappas, Yi~Mao, Weizhu Chen, and Noah~A. Smith.
\newblock Finetuning pretrained transformers into {RNN}s.
\newblock In \emph{Proceedings of the 2021 Conference on Empirical Methods in Natural Language Processing}, pp.\  10630--10643, Online and Punta Cana, Dominican Republic, November 2021. Association for Computational Linguistics.
\newblock \doi{10.18653/v1/2021.emnlp-main.830}.
\newblock URL \url{https://aclanthology.org/2021.emnlp-main.830}.

\bibitem[Katharopoulos et~al.(2020)Katharopoulos, Vyas, Pappas, and Fleuret]{katharopoulos2020transformers}
Angelos Katharopoulos, Apoorv Vyas, Nikolaos Pappas, and Fran{\c{c}}ois Fleuret.
\newblock Transformers are rnns: Fast autoregressive transformers with linear attention.
\newblock In \emph{International conference on machine learning}, pp.\  5156--5165. PMLR, 2020.

\bibitem[Keles et~al.(2023)Keles, Wijewardena, and Hegde]{keles2023computational}
Feyza~Duman Keles, Pruthuvi~Mahesakya Wijewardena, and Chinmay Hegde.
\newblock On the computational complexity of self-attention.
\newblock In \emph{International Conference on Algorithmic Learning Theory}, pp.\  597--619. PMLR, 2023.

\bibitem[Kwon et~al.(2023)Kwon, Li, Zhuang, Sheng, Zheng, Yu, Gonzalez, Zhang, and Stoica]{kwon2023efficient}
Woosuk Kwon, Zhuohan Li, Siyuan Zhuang, Ying Sheng, Lianmin Zheng, Cody~Hao Yu, Joseph Gonzalez, Hao Zhang, and Ion Stoica.
\newblock Efficient memory management for large language model serving with pagedattention.
\newblock In \emph{Proceedings of the 29th Symposium on Operating Systems Principles}, pp.\  611--626, 2023.

\bibitem[Li et~al.(2023)Li, Bubeck, Eldan, Del~Giorno, Gunasekar, and Lee]{li2023textbooks}
Yuanzhi Li, S{\'e}bastien Bubeck, Ronen Eldan, Allie Del~Giorno, Suriya Gunasekar, and Yin~Tat Lee.
\newblock Textbooks are all you need ii: phi-1.5 technical report.
\newblock \emph{arXiv preprint arXiv:2309.05463}, 2023.

\bibitem[Loshchilov \& Hutter(2017)Loshchilov and Hutter]{loshchilov2017decoupled}
Ilya Loshchilov and Frank Hutter.
\newblock Decoupled weight decay regularization.
\newblock \emph{arXiv preprint arXiv:1711.05101}, 2017.

\bibitem[Mao(2022)]{mao-2022-fine}
Huanru~Henry Mao.
\newblock Fine-tuning pre-trained transformers into decaying fast weights.
\newblock In \emph{Proceedings of the 2022 Conference on Empirical Methods in Natural Language Processing}, pp.\  10236--10242, Abu Dhabi, United Arab Emirates, December 2022. Association for Computational Linguistics.
\newblock \doi{10.18653/v1/2022.emnlp-main.697}.
\newblock URL \url{https://aclanthology.org/2022.emnlp-main.697}.

\bibitem[Mercat et~al.(2024)Mercat, Vasiljevic, Keh, Arora, Dave, Gaidon, and Kollar]{mercat2024linearizing}
Jean Mercat, Igor Vasiljevic, Sedrick Keh, Kushal Arora, Achal Dave, Adrien Gaidon, and Thomas Kollar.
\newblock Linearizing large language models.
\newblock \emph{arXiv preprint arXiv:2405.06640}, 2024.

\bibitem[Merity et~al.(2017)Merity, Xiong, Bradbury, and Socher]{merity2017pointer}
Stephen Merity, Caiming Xiong, James Bradbury, and Richard Socher.
\newblock Pointer sentinel mixture models.
\newblock In \emph{International Conference on Learning Representations}, 2017.
\newblock URL \url{https://openreview.net/forum?id=Byj72udxe}.

\bibitem[Mohtashami \& Jaggi(2023)Mohtashami and Jaggi]{mohtashami2023landmark}
Amirkeivan Mohtashami and Martin Jaggi.
\newblock Landmark attention: Random-access infinite context length for transformers.
\newblock \emph{arXiv preprint arXiv:2305.16300}, 2023.

\bibitem[Munkhdalai et~al.(2024)Munkhdalai, Faruqui, and Gopal]{Munkhdalai2024-kx}
Tsendsuren Munkhdalai, Manaal Faruqui, and Siddharth Gopal.
\newblock Leave no context behind: Efficient infinite context transformers with infini-attention.
\newblock April 2024.

\bibitem[Penedo et~al.(2023)Penedo, Malartic, Hesslow, Cojocaru, Cappelli, Alobeidli, Pannier, Almazrouei, and Launay]{penedo2023refinedweb}
Guilherme Penedo, Quentin Malartic, Daniel Hesslow, Ruxandra Cojocaru, Alessandro Cappelli, Hamza Alobeidli, Baptiste Pannier, Ebtesam Almazrouei, and Julien Launay.
\newblock The refinedweb dataset for falcon llm: outperforming curated corpora with web data, and web data only.
\newblock \emph{arXiv preprint arXiv:2306.01116}, 2023.

\bibitem[Peng et~al.(2023{\natexlab{a}})Peng, Alcaide, Anthony, Albalak, Arcadinho, Cao, Cheng, Chung, Grella, GV, et~al.]{peng2023rwkv}
Bo~Peng, Eric Alcaide, Quentin Anthony, Alon Albalak, Samuel Arcadinho, Huanqi Cao, Xin Cheng, Michael Chung, Matteo Grella, Kranthi~Kiran GV, et~al.
\newblock Rwkv: Reinventing rnns for the transformer era.
\newblock \emph{arXiv preprint arXiv:2305.13048}, 2023{\natexlab{a}}.

\bibitem[Peng et~al.(2024)Peng, Goldstein, Anthony, Albalak, Alcaide, Biderman, Cheah, Ferdinan, Hou, Kazienko, et~al.]{peng2024eagle}
Bo~Peng, Daniel Goldstein, Quentin Anthony, Alon Albalak, Eric Alcaide, Stella Biderman, Eugene Cheah, Teddy Ferdinan, Haowen Hou, Przemys{\l}aw Kazienko, et~al.
\newblock Eagle and finch: Rwkv with matrix-valued states and dynamic recurrence.
\newblock \emph{arXiv preprint arXiv:2404.05892}, 2024.

\bibitem[Peng et~al.(2023{\natexlab{b}})Peng, Quesnelle, Fan, and Shippole]{peng2023yarn}
Bowen Peng, Jeffrey Quesnelle, Honglu Fan, and Enrico Shippole.
\newblock Yarn: Efficient context window extension of large language models.
\newblock \emph{arXiv preprint arXiv:2309.00071}, 2023{\natexlab{b}}.

\bibitem[Poli et~al.(2023{\natexlab{a}})Poli, Massaroli, Nguyen, Fu, Dao, Baccus, Bengio, Ermon, and R{\'e}]{poli2023hyena}
Michael Poli, Stefano Massaroli, Eric Nguyen, Daniel~Y Fu, Tri Dao, Stephen Baccus, Yoshua Bengio, Stefano Ermon, and Christopher R{\'e}.
\newblock Hyena hierarchy: Towards larger convolutional language models.
\newblock \emph{arXiv preprint arXiv:2302.10866}, 2023{\natexlab{a}}.

\bibitem[Poli et~al.(2023{\natexlab{b}})Poli, Wang, Massaroli, Quesnelle, Carlow, Nguyen, and Thomas]{stripedhyena}
Michael Poli, Jue Wang, Stefano Massaroli, Jeffrey Quesnelle, Ryan Carlow, Eric Nguyen, and Armin Thomas.
\newblock {StripedHyena: Moving Beyond Transformers with Hybrid Signal Processing Models}, 12 2023{\natexlab{b}}.
\newblock URL \url{https://github.com/togethercomputer/stripedhyena}.

\bibitem[Qin et~al.(2022)Qin, Han, Sun, Li, Kong, Barnes, and Zhong]{qin-etal-2022-devil}
Zhen Qin, Xiaodong Han, Weixuan Sun, Dongxu Li, Lingpeng Kong, Nick Barnes, and Yiran Zhong.
\newblock The devil in linear transformer.
\newblock In \emph{Proceedings of the 2022 Conference on Empirical Methods in Natural Language Processing}, pp.\  7025--7041, Abu Dhabi, United Arab Emirates, December 2022. Association for Computational Linguistics.
\newblock \doi{10.18653/v1/2022.emnlp-main.473}.
\newblock URL \url{https://aclanthology.org/2022.emnlp-main.473}.

\bibitem[Qin et~al.(2023)Qin, Li, Sun, Sun, Shen, Han, Wei, Lv, Yuan, Luo, et~al.]{qin2023scaling}
Zhen Qin, Dong Li, Weigao Sun, Weixuan Sun, Xuyang Shen, Xiaodong Han, Yunshen Wei, Baohong Lv, Fei Yuan, Xiao Luo, et~al.
\newblock Scaling transnormer to 175 billion parameters.
\newblock \emph{arXiv preprint arXiv:2307.14995}, 2023.

\bibitem[Radford et~al.(2019)Radford, Wu, Child, Luan, Amodei, and Sutskever]{Radford2019LanguageMA}
Alec Radford, Jeff Wu, Rewon Child, David Luan, Dario Amodei, and Ilya Sutskever.
\newblock Language models are unsupervised multitask learners.
\newblock 2019.
\newblock URL \url{https://api.semanticscholar.org/CorpusID:160025533}.

\bibitem[Raffel et~al.(2020)Raffel, Shazeer, Roberts, Lee, Narang, Matena, Zhou, Li, and Liu]{raffel2020exploring}
Colin Raffel, Noam Shazeer, Adam Roberts, Katherine Lee, Sharan Narang, Michael Matena, Yanqi Zhou, Wei Li, and Peter~J Liu.
\newblock Exploring the limits of transfer learning with a unified text-to-text transformer.
\newblock \emph{Journal of machine learning research}, 21\penalty0 (140):\penalty0 1--67, 2020.

\bibitem[Raschka(2023)]{Raschka_2023}
Sebastian Raschka.
\newblock Finetuning llms with lora and qlora: Insights from hundreds of experiments, 2023.
\newblock URL \url{https://lightning.ai/pages/community/lora-insights/}.

\bibitem[Schlag et~al.(2021)Schlag, Irie, and Schmidhuber]{schlag2021linear}
Imanol Schlag, Kazuki Irie, and J{\"u}rgen Schmidhuber.
\newblock Linear transformers are secretly fast weight programmers.
\newblock In \emph{International Conference on Machine Learning}, pp.\  9355--9366. PMLR, 2021.

\bibitem[Shaham et~al.(2022)Shaham, Segal, Ivgi, Efrat, Yoran, Haviv, Gupta, Xiong, Geva, Berant, and Levy]{shaham-etal-2022-scrolls}
Uri Shaham, Elad Segal, Maor Ivgi, Avia Efrat, Ori Yoran, Adi Haviv, Ankit Gupta, Wenhan Xiong, Mor Geva, Jonathan Berant, and Omer Levy.
\newblock {SCROLLS}: Standardized {C}ompa{R}ison over long language sequences.
\newblock In \emph{Proceedings of the 2022 Conference on Empirical Methods in Natural Language Processing}, pp.\  12007--12021, Abu Dhabi, United Arab Emirates, December 2022. Association for Computational Linguistics.
\newblock URL \url{https://aclanthology.org/2022.emnlp-main.823}.

\bibitem[Shen et~al.(2024)Shen, Li, Leng, Qin, Sun, and Zhong]{shen2024scaling}
Xuyang Shen, Dong Li, Ruitao Leng, Zhen Qin, Weigao Sun, and Yiran Zhong.
\newblock Scaling laws for linear complexity language models.
\newblock \emph{arXiv preprint arXiv:2406.16690}, 2024.

\bibitem[Snell et~al.(2024)Snell, Lee, Xu, and Kumar]{snell2024scaling}
Charlie Snell, Jaehoon Lee, Kelvin Xu, and Aviral Kumar.
\newblock Scaling llm test-time compute optimally can be more effective than scaling model parameters.
\newblock \emph{arXiv preprint arXiv:2408.03314}, 2024.

\bibitem[Spector et~al.(2024)Spector, Singhal, Arora, and Ré]{Spector_Singhal_Arora_Ré_2024}
Benjamin Spector, Aaryan Singhal, Simran Arora, and Christopher Ré.
\newblock Gpus go brrr, May 2024.
\newblock URL \url{https://hazyresearch.stanford.edu/blog/2024-05-12-tk}.

\bibitem[Su et~al.(2024)Su, Ahmed, Lu, Pan, Bo, and Liu]{su2024roformer}
Jianlin Su, Murtadha Ahmed, Yu~Lu, Shengfeng Pan, Wen Bo, and Yunfeng Liu.
\newblock Roformer: Enhanced transformer with rotary position embedding.
\newblock \emph{Neurocomputing}, 568:\penalty0 127063, 2024.

\bibitem[Sun et~al.(2023)Sun, Dong, Huang, Ma, Xia, Xue, Wang, and Wei]{sun2023retentive}
Yutao Sun, Li~Dong, Shaohan Huang, Shuming Ma, Yuqing Xia, Jilong Xue, Jianyong Wang, and Furu Wei.
\newblock Retentive network: A successor to transformer for large language models.
\newblock \emph{arXiv preprint arXiv:2307.08621}, 2023.

\bibitem[Talmor et~al.(2019)Talmor, Herzig, Lourie, and Berant]{talmor-etal-2019-commonsenseqa}
Alon Talmor, Jonathan Herzig, Nicholas Lourie, and Jonathan Berant.
\newblock {C}ommonsense{QA}: A question answering challenge targeting commonsense knowledge.
\newblock In Jill Burstein, Christy Doran, and Thamar Solorio (eds.), \emph{Proceedings of the 2019 Conference of the North {A}merican Chapter of the Association for Computational Linguistics: Human Language Technologies, Volume 1 (Long and Short Papers)}, pp.\  4149--4158, Minneapolis, Minnesota, June 2019. Association for Computational Linguistics.
\newblock \doi{10.18653/v1/N19-1421}.
\newblock URL \url{https://aclanthology.org/N19-1421}.

\bibitem[Taori et~al.(2023)Taori, Gulrajani, Zhang, Dubois, Li, Guestrin, Liang, and Hashimoto]{alpaca}
Rohan Taori, Ishaan Gulrajani, Tianyi Zhang, Yann Dubois, Xuechen Li, Carlos Guestrin, Percy Liang, and Tatsunori~B. Hashimoto.
\newblock Stanford alpaca: An instruction-following llama model.
\newblock \url{https://github.com/tatsu-lab/stanford_alpaca}, 2023.

\bibitem[Touvron et~al.(2023{\natexlab{a}})Touvron, Lavril, Izacard, Martinet, Lachaux, Lacroix, Rozi{\`e}re, Goyal, Hambro, Azhar, et~al.]{touvron2023llama}
Hugo Touvron, Thibaut Lavril, Gautier Izacard, Xavier Martinet, Marie-Anne Lachaux, Timoth{\'e}e Lacroix, Baptiste Rozi{\`e}re, Naman Goyal, Eric Hambro, Faisal Azhar, et~al.
\newblock Llama: Open and efficient foundation language models.
\newblock \emph{arXiv preprint arXiv:2302.13971}, 2023{\natexlab{a}}.

\bibitem[Touvron et~al.(2023{\natexlab{b}})Touvron, Martin, Stone, Albert, Almahairi, Babaei, Bashlykov, Batra, Bhargava, Bhosale, et~al.]{touvron2023llama2}
Hugo Touvron, Louis Martin, Kevin Stone, Peter Albert, Amjad Almahairi, Yasmine Babaei, Nikolay Bashlykov, Soumya Batra, Prajjwal Bhargava, Shruti Bhosale, et~al.
\newblock Llama 2: Open foundation and fine-tuned chat models.
\newblock \emph{arXiv preprint arXiv:2307.09288}, 2023{\natexlab{b}}.

\bibitem[Tworkowski et~al.(2024)Tworkowski, Staniszewski, Pacek, Wu, Michalewski, and Mi{\l}o{\'s}]{tworkowski2024focused}
Szymon Tworkowski, Konrad Staniszewski, Miko{\l}aj Pacek, Yuhuai Wu, Henryk Michalewski, and Piotr Mi{\l}o{\'s}.
\newblock Focused transformer: Contrastive training for context scaling.
\newblock \emph{Advances in Neural Information Processing Systems}, 36, 2024.

\bibitem[Vaswani et~al.(2017)Vaswani, Shazeer, Parmar, Uszkoreit, Jones, Gomez, Kaiser, and Polosukhin]{vaswani2017attention}
Ashish Vaswani, Noam Shazeer, Niki Parmar, Jakob Uszkoreit, Llion Jones, Aidan~N Gomez, {\L}ukasz Kaiser, and Illia Polosukhin.
\newblock Attention is all you need.
\newblock \emph{Advances in neural information processing systems}, 30, 2017.

\bibitem[Waleffe et~al.(2024)Waleffe, Byeon, Riach, Norick, Korthikanti, Dao, Gu, Hatamizadeh, Singh, Narayanan, et~al.]{waleffe2024empirical}
Roger Waleffe, Wonmin Byeon, Duncan Riach, Brandon Norick, Vijay Korthikanti, Tri Dao, Albert Gu, Ali Hatamizadeh, Sudhakar Singh, Deepak Narayanan, et~al.
\newblock An empirical study of mamba-based language models.
\newblock \emph{arXiv preprint arXiv:2406.07887}, 2024.

\bibitem[Wang et~al.(2018)Wang, Singh, Michael, Hill, Levy, and Bowman]{wang2018glue}
Alex Wang, Amanpreet Singh, Julian Michael, Felix Hill, Omer Levy, and Samuel~R Bowman.
\newblock Glue: A multi-task benchmark and analysis platform for natural language understanding.
\newblock \emph{arXiv preprint arXiv:1804.07461}, 2018.

\bibitem[Wang et~al.(2024)Wang, Paliotta, May, Rush, and Dao]{wang2024mamba}
Junxiong Wang, Daniele Paliotta, Avner May, Alexander~M Rush, and Tri Dao.
\newblock The mamba in the llama: Distilling and accelerating hybrid models.
\newblock \emph{arXiv preprint arXiv:2408.15237}, 2024.

\bibitem[Wu \& He(2018)Wu and He]{wu2018group}
Yuxin Wu and Kaiming He.
\newblock Group normalization.
\newblock In \emph{Proceedings of the European conference on computer vision (ECCV)}, pp.\  3--19, 2018.

\bibitem[Xiong et~al.(2021)Xiong, Zeng, Chakraborty, Tan, Fung, Li, and Singh]{xiong2021nystromformer}
Yunyang Xiong, Zhanpeng Zeng, Rudrasis Chakraborty, Mingxing Tan, Glenn Fung, Yin Li, and Vikas Singh.
\newblock Nystr{\"o}mformer: A nystr{\"o}m-based algorithm for approximating self-attention.
\newblock In \emph{Proceedings of the AAAI Conference on Artificial Intelligence}, volume~35, pp.\  14138--14148, 2021.

\bibitem[Yang et~al.(2023)Yang, Wang, Shen, Panda, and Kim]{yang2023gated}
Songlin Yang, Bailin Wang, Yikang Shen, Rameswar Panda, and Yoon Kim.
\newblock Gated linear attention transformers with hardware-efficient training.
\newblock \emph{arXiv preprint arXiv:2312.06635}, 2023.

\bibitem[Yang et~al.(2024)Yang, Wang, Zhang, Shen, and Kim]{yang2024parallelizing}
Songlin Yang, Bailin Wang, Yu~Zhang, Yikang Shen, and Yoon Kim.
\newblock Parallelizing linear transformers with the delta rule over sequence length.
\newblock \emph{arXiv preprint arXiv:2406.06484}, 2024.

\bibitem[Zaheer et~al.(2020)Zaheer, Guruganesh, Dubey, Ainslie, Alberti, Ontanon, Pham, Ravula, Wang, Yang, et~al.]{zaheer2020big}
Manzil Zaheer, Guru Guruganesh, Kumar~Avinava Dubey, Joshua Ainslie, Chris Alberti, Santiago Ontanon, Philip Pham, Anirudh Ravula, Qifan Wang, Li~Yang, et~al.
\newblock Big bird: Transformers for longer sequences.
\newblock \emph{Advances in neural information processing systems}, 33:\penalty0 17283--17297, 2020.

\bibitem[Zhang et~al.(2024)Zhang, Bhatia, Kumbong, and Re]{zhang2024the}
Michael Zhang, Kush Bhatia, Hermann Kumbong, and Christopher Re.
\newblock The hedgehog \& the porcupine: Expressive linear attentions with softmax mimicry.
\newblock In \emph{The Twelfth International Conference on Learning Representations}, 2024.
\newblock URL \url{https://openreview.net/forum?id=4g02l2N2Nx}.

\bibitem[Zhu et~al.(2021)Zhu, Ping, Xiao, Shoeybi, Goldstein, Anandkumar, and Catanzaro]{zhu2021long}
Chen Zhu, Wei Ping, Chaowei Xiao, Mohammad Shoeybi, Tom Goldstein, Anima Anandkumar, and Bryan Catanzaro.
\newblock Long-short transformer: Efficient transformers for language and vision.
\newblock \emph{Advances in neural information processing systems}, 34:\penalty0 17723--17736, 2021.

\end{thebibliography}
\bibliographystyle{iclr2025_conference}

\appendix
\clearpage

\section{Experimental Details}
\label{appendix:experimental_details}

\subsection{Main Results, Linearizing 7B and 8B LLMs}
\label{app:experimenta_details_linearizing_7b}

\paragraph{Setup.} We describe our setup for linearizing Mistral 7B (v0.1)~\citep{jiang2023mistral}, Llama 3 8B~\citep{llama3modelcard}, and Llama 3.1 8B~\citep{dubey2024llama}. 

For linearizing layers, we replace softmax attentions with hybrid linear + sliding window analogs (Section~\ref{sec:method_analogs}), using Hedgehog's feature map for its prior quality~\citep{zhang2024the}.  
%

\begin{wrapfigure}{r}{0.5\textwidth}
    \centering
    \vspace{-0.5cm}
    \begin{subfigure}[b]{0.235\textwidth}
        \centering
        \includegraphics[width=\textwidth]{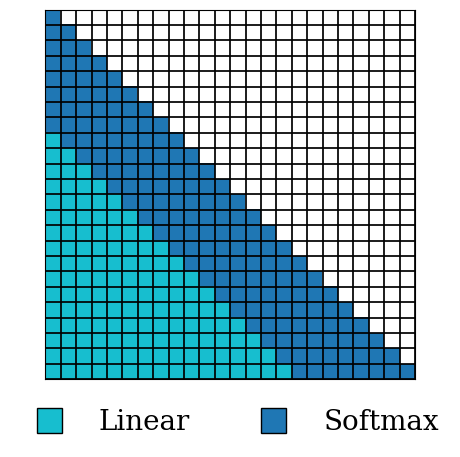}
        \vspace{-0.5cm}
        \caption{\footnotesize Sliding window}
        \label{fig:hybrid_attn_sw}
    \end{subfigure}
    \begin{subfigure}[b]{0.235\textwidth}
        \centering
        \includegraphics[width=\textwidth]{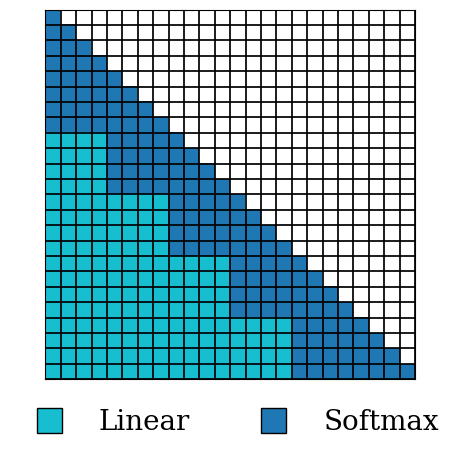}
        \vspace{-0.5cm}
        \caption{\footnotesize TK ``terrace''}
        \label{fig:hybrid_attn_tk}
    \end{subfigure}
    \caption{We apply softmax attention locally and attend to all past tokens with linear attention. 
    }
    \label{fig:hybrid_attn}
\end{wrapfigure}

For the sliding window implementation, we considered two options: a standard sliding window where $w$ is the same for all tokens, and a ``terraced'' window where $w$ changes based on token index (Figure~\ref{fig:hybrid_attn}). While we found both comparable in quality (Table~\ref{table:app_ablate_window_size}), the latter lets us exploit the new ThunderKittens (TK) DSL's~\citep{Spector_Singhal_Arora_Ré_2024} primitives for implementing fast CUDA kernels. Here we prefer contiguous blocks of size $w = 64$, which can quickly 
be computed in parallel on modern GPUs. We use this ``terrace'' implementation in our main results, and include further implementation details in Appendix~\ref{app:tk_implementation}.

For linearizing data, we use the Alpaca linearizing data setup in Section~\ref{sec:method_study} unless otherwise noted. We also tried a more typical pretraining corpus (a subset\footnote{\url{https://huggingface.co/datasets/togethercomputer/RedPajama-Data-1T-Sample}} of RedPajama~\citep{together2023redpajama}), but found comparable performance when controlling for number of token updates (Appendix~\ref{app:ablate_alpaca_vs_rp}). 
%
To linearize, we simply train all feature maps in parallel for two epochs with learning rate 1e-2, before applying LoRA on
the attention projection layers for two epochs with learning rate 1e-4. By default, we use LoRA rank $r = 8$, and scale LoRA updates by $2$ ($\alpha$ = 16 in HuggingFace PEFT\footnote{\url{https://huggingface.co/docs/peft/en/index}}), amounting to training $<$0.09\% of all model parameters.
For both
stages, we train with early stopping, AdamW optimizer~\citep{loshchilov2017decoupled}, and packing into 1024-token sequences with batch size 8. We evaluate the best checkpoints based on validation set perplexity.

\paragraph{Hyperparameters.} We list all model and training hyperparameters in Table~\ref{tab:8b-training-details}. For learning rates, we did an initial sweep over \{1e-2, 1e-3, 1e-4\}, choosing the best based on final validation set perplexity during step 2: low-rank adjusting, and checkpointing with early stopping. We did not tune batch size or choice of optimizer, and used default values informed by prior work for other design parameters such as sliding window size~\citep{arora2024simple}, LoRA rank, and LoRA projection layers~\citep{hu2021lora}. In Appendix~\ref{appendix:additional_experiments}, we study the effect of sweeping various values such as window sizes, ranks, and LoRA modules as ablations.
\begin{table}[h!]
    \centering
    \begin{tabular}{rccc}
    \toprule
    {} & Hedgehog & \name{} \\
    \midrule
     \textbf{Model} & & \\
    Precision & \multicolumn{2}{c}{16-bit (bfloat16)} \\
    Sequence length & \multicolumn{2}{c}{1024} \\
    Linearizing attention & (Linear) & (Linear + Sliding Window) \\
    Linear attn feature map & \multicolumn{2}{c}{Hedgehog} \\
    Linear attn feature dimension & \multicolumn{2}{c}{64 (effectively 128, see Table~\ref{table:linear_attentions_def})} \\
    Linear attn feature activation & \multicolumn{2}{c}{Softmax (across feature dim)} \\
    Sliding window implementation & N/A & Terrace \\
    Sliding window attn size & N/A & 64 \\
    \midrule
    \textbf{Optimizer and LR Schedule} & & \\
    Optimizer & \multicolumn{2}{c}{AdamW} \\
    Global batch size & \multicolumn{2}{c}{8} \\ 
    Gradient accumulation & \multicolumn{2}{c}{8} \\
    Gradient clipping threshold & \multicolumn{2}{c}{$1.0$} \\
    Learning rate schedule & \multicolumn{2}{c}{Reduce LR on Plateau} \\
    \midrule 
    \textbf{Step 1: Attention Transfer} & & \\
    Number of epochs & \multicolumn{2}{c}{2}\\
    Tokens per epoch & \multicolumn{2}{c}{10M} \\
    Learning rate & \multicolumn{2}{c}{0.01} \\
    \midrule
    \textbf{Step 2: Low-rank Adjusting} & & \\
    Number of epochs & \multicolumn{2}{c}{2} \\
    Tokens per epoch & \multicolumn{2}{c}{20M} \\
    Learning rate & \multicolumn{2}{c}{1e-4} \\
    LoRA rank and alpha & \multicolumn{2}{c}{$r=$8, $\alpha=$16} \\
    LoRA dropout & \multicolumn{2}{c}{$0.0$} \\
    LoRA projections & \multicolumn{2}{c}{$\bm{W}_q, \bm{W}_k, \bm{W}_v, \bm{W}_o$} \\
    \bottomrule 
    \end{tabular}
    \caption{\small Hyperparameters for Mistral 7B, Llama 3 8B, and Llama 3.1 8B experiments.}
    \label{tab:8b-training-details}
\end{table}

\paragraph{Compute Resources.} For each linearizing run we use one NVIDIA 40GB A100 GPU. With batch size 1 and gradient accumulation over 8 batches, attention transfer takes $\approx$~2 hours and post-swap finetuning takes $\approx$~4.5 hours,~\ie{} 6.5 total GPU hours to linearize an 8B LLM.

\subsection{Linearizing Llama 3.1 70B}
\label{app:experimenta_details_linearizing_70b}
We provide experimental details corresponding to the $70$B parameter results reported in \Cref{table:chungus_results}.

\paragraph{Setup.} We compare the quality of two linearization approaches to the quality of the original Llama 3.1 70B model, including (1) the baseline linearization \textit{without} attention transfer, which is representative of the approach used in prior work~\citep{mercat2024linearizing, yang2024parallelizing, wang2024mamba} and (2) our approach, \name{}. For both the baseline and \name{}, we start with Llama 3.1 70B and replace the softmax attentions with the linear attention architecture defined in \Cref{sec:method_analogs}, \Cref{eq:hybrid_linear_swa}. The training procedure involves:
\begin{itemize}[leftmargin=*]
    \item Baseline: We introduce LoRA parameters to the attention $\bm{W}_q, \bm{W}_k, \bm{W}_v, \bm{W}_o$ projection matrices.  We train the linear attention feature maps, learnable mixing term $\gamma$, and the LoRA parameters during the fine-tuning adjustment stage of the linearization process.
    \item \name{}: We first perform layer-wise attention transfer following \Cref{eq:mse_loss_blocks}, with $k=80$ (i.e., we optimize over all layers together). We then introduce LoRA parameters to the attention $\bm{W}_q, \bm{W}_k, \bm{W}_v, \bm{W}_o$ projection matrices. During fine-tuning we only train the LoRA parameters, freezing the linear attention map and $\gamma$ weights.
\end{itemize}
We use an MSE loss for the layer-wise attention transfer stage and cross-entropy loss for the next-token prediction fine-tuning (Low-rank Finetuning) stage.


\paragraph{Hyperparameters.} We include the hyperparameters for the baseline and \name{} approaches in Table~\ref{tab:70b-training-details}, following the same sweep as the 8B models.  Since the baseline does not use attention transfer, we mark these values with ``N/A''.
We linearize each model using the same randomly sampled $20$M tokens of the RedPajama pre-training corpus~\citep{together2023redpajama}. We pack the sequences to fill full context length, and evaluate the best checkpoints based on validation set perplexity.

\begin{table}[h!]
    \centering
    \begin{tabular}{rccc}
    \toprule
    {} & Baseline & \name{} \\
    \midrule
     \textbf{Model} & & \\
    Precision & \multicolumn{2}{c}{16-bit (bfloat16)} \\
    Sequence length & \multicolumn{2}{c}{1024} \\
    Linearizing attention & \multicolumn{2}{c}{Linear + Sliding Window} \\
    Linear attn feature map & \multicolumn{2}{c}{Hedgehog} \\
    Linear attn feature dimension & \multicolumn{2}{c}{64 (effectively 128, see Table~\ref{table:linear_attentions_def})} \\
    Linear attn feature activation & \multicolumn{2}{c}{Softmax (across feature dim)} \\
    Sliding window implementation & \multicolumn{2}{c}{Terrace} \\
    Sliding window attn size & \multicolumn{2}{c}{64} \\
    \midrule
    \textbf{Optimizer and LR Schedule} & & \\
    Optimizer & \multicolumn{2}{c}{AdamW} \\
    Global batch size & \multicolumn{2}{c}{8} \\ 
    Gradient accumulation & \multicolumn{2}{c}{8} \\
    Gradient clipping threshold & \multicolumn{2}{c}{$1.0$} \\
    Learning rate schedule & \multicolumn{2}{c}{Reduce LR on Plateau} \\
    \midrule 
    \textbf{Stage 1: Attention Transfer} & & \\
    Number of epochs & N/A & 1 \\
    Tokens per epoch & N/A & 20M \\
    Learning rate & N/A & 0.01 \\
    \midrule
    \textbf{Stage 2: Low-rank Adjusting} & & \\
    Number of epochs & \multicolumn{2}{c}{1} \\
    Tokens per epoch & 20M & 20M \\
    Learning rate & \multicolumn{2}{c}{1e-4} \\
    LoRA rank and alpha & \multicolumn{2}{c}{$r=$8, $\alpha=$16} \\
    LoRA dropout & \multicolumn{2}{c}{$0.0$} \\
    LoRA projections & \multicolumn{2}{c}{$\bm{W}_q, \bm{W}_k, \bm{W}_v, \bm{W}_o$} \\
    \bottomrule 
    \end{tabular}
    \caption{\small Hyperparameters for Llama 3.1 70B experiments.}
    \label{tab:70b-training-details}
\end{table}

\paragraph{Compute Resources.} We linearize using a single NVIDIA $8 \times 80$GB H100 node. Attention transfer takes $4$ hours and fine-tuning takes $14$ hours. We use PyTorch FSDP with activation checkpointing for distributed training.

\subsection{Linearizing Llama 3.1 405B}
\label{app:experimenta_details_linearizing_405b}

\paragraph{Setup.} We compare the quality of two linearization approaches to the quality of the original Llama 3.1 405B model, including (1) the baseline linearization \textit{without} attention transfer, which is representative of the approach used in prior work~\citep{mercat2024linearizing} and (2) our approach, \name{}. 
For both the baseline and \name{}, we start with Llama 3.1 405B and replace the softmax attentions with the linear attention architecture defined in \Cref{sec:method_analogs}, \Cref{eq:hybrid_linear_swa}. 
The training procedure involves:
\begin{itemize}[leftmargin=*]
    \item Baseline: We introduce LoRA parameters to the attention $\bm{W}_q, \bm{W}_k, \bm{W}_v, \bm{W}_o$ projection matrices.  
    We train the linear attention feature maps, learnable mixing term $\gamma$, and the LoRA parameters during the fine-tuning adjustment stage of the linearization process.
    \item \name{}: 
    We first perform block-wise attention transfer following \Cref{eq:mse_loss_blocks}, with $k=9$ as the block size. To perform attention transfer for block $i$, we save the hidden states outputted by  block $i-1$ to disk and then use this as training data for block $i$.
    We then introduce LoRA parameters to the attention $\bm{W}_q, \bm{W}_k, \bm{W}_v, \bm{W}_o$ projection matrices. 
    During fine-tuning we only train the LoRA parameters, freezing the linear attention map and $\gamma$ weights.
\end{itemize}
For the reported checkpoints, we train the layer-wise attention transfer stage using use a weighted combination of the MSE loss on attention outputs plus a cross-entropy loss between the softmax and linear attention maps.  We use cross-entropy loss for Stage 2 Low-rank Linearizing. 


\paragraph{Hyperparameters.} We include the hyperparameters for the baseline and \name{} approaches in Table~\ref{tab:405b-training-details}, following the same sweep as the 8B models. Since the baseline does not use attention transfer, we mark it with ``N/A''.
We linearize each model using the same randomly sampled $20$M tokens of the RedPajama pre-training corpus~\citep{together2023redpajama}. We pack the sequences to fill the full context length, and evaluate the best checkpoints based on validation set perplexity.
\begin{table}[h!]
    \centering
    \begin{tabular}{rccc}
    \toprule
    {} & Baseline & \name{} \\
    \midrule
     \textbf{Model} & & \\
    Precision & \multicolumn{2}{c}{16-bit (FP16)} \\
    Sequence length & \multicolumn{2}{c}{1024} \\
    Linearizing attention & \multicolumn{2}{c}{Linear + Sliding Window} \\
    Linear attn feature map & \multicolumn{2}{c}{Hedgehog} \\
    Linear attn feature dimension & \multicolumn{2}{c}{64 (effectively 128, see Table~\ref{table:linear_attentions_def})} \\
    Linear attn feature activation & \multicolumn{2}{c}{Softmax (across feature dim)} \\
    Sliding window implementation & \multicolumn{2}{c}{Terrace} \\
    Sliding window attn size & \multicolumn{2}{c}{64} \\
    \midrule
    \textbf{Optimizer and LR Schedule} & & \\
    Optimizer & \multicolumn{2}{c}{AdamW} \\
    Global batch size & \multicolumn{2}{c}{8} \\ 
    Gradient accumulation & \multicolumn{2}{c}{8} \\
    Gradient clipping threshold & \multicolumn{2}{c}{$1.0$} \\
    Learning rate schedule & \multicolumn{2}{c}{Reduce LR on Plateau} \\
    \midrule 
    \textbf{Stage 1: Attention Transfer} & & \\
    Number of epochs & N/A & 1 \\
    Tokens per epoch & N/A & 20M \\
    Learning rate & N/A & 0.01 \\
    MSE and X-ent weights & N/A & 1000, 1 \\
    \midrule
    \textbf{Stage 2: Low-rank Adjusting} & & \\
    Number of epochs & \multicolumn{2}{c}{1} \\
    Tokens per epoch & 20M & 20M \\
    Learning rate & \multicolumn{2}{c}{1e-4} \\
    LoRA rank and alpha & \multicolumn{2}{c}{$r=4$, $\alpha=8$} \\
    LoRA dropout & \multicolumn{2}{c}{$0.5$} \\
    LoRA projections & \multicolumn{2}{c}{$\bm{W}_q, \bm{W}_k, \bm{W}_v, \bm{W}_o$} \\
    \bottomrule 
    \end{tabular}
    \caption{\small Hyperparameters for Llama 3.1 405B experiments.}
    \label{tab:405b-training-details}
\end{table}

\paragraph{Compute Resources.} We linearize the baseline using three NVIDIA $8 \times 80$GB H100 nodes, evaluating the best validation checkpoint after 19.5 hours. 
For \name{}, we perform attention transfer with 1 $80$GB H100 GPU for 5 hours per block, and we finetune with 3 NVIDIA $8 \times 80$GB H100 nodes for 16 hours. We use PyTorch FSDP with activation checkpointing for distributed training.

\section{Additional Experiments and Results}
\label{appendix:additional_experiments}

To better understand \name{}' properties and performance,
we now report extended results on \name{}. We first report expanded results of our main paper, including task-specific ablation numbers, multiple seeds, and additional comparison at the 1B LLM scale. We then extend our ablations by studying how different amounts of parameter updates, different data sources, and different amounts of training data affect \name{} quality for various tasks such as zero-shot LM Eval, 5-shot MMLU, and passkey retrieval. 
We finally expand on the layer-wise training dynamics of \name{}: how layer-wise MSE between learned and softmax attentions changes during LoRA adjusting (App.~\ref{app:layerwise_mse_study}), how LoRA weights update over time depending on attention transfer and softmax attention-matching quality (App.~\ref{app:layerwise_lora_train}), and how block-wise attention transfer both is motivated by layer-wise MSE and improves linearized LLM quality for larger LLMs (App.~\ref{app:blockwise_study}).

\subsection{Expanded Results}
\label{appendix:add_exp_ext_results}

\subsubsection{Task-specific Results for \name{} Component Ablations}
\label{appendix:add_exp_main_ablations_full}

We report the task-specific LM Eval results when ablating the attention transfer and linear + sliding window attention in \name{}, expanding on Table~\ref{table:main_results_ablations} in Table~\ref{table:main_results_ablations_full}. Across all but one task (ARC-easy), we validate that the \name{} proposed combination leads to best performance.

\begin{table}[h]
\begin{center}
\resizebox{1.0\linewidth}{!}{
\begin{tabular}{@{}c|cc|ccccccc@{}}
\toprule
Feature Map                                        & \begin{tabular}[c]{@{}c@{}}+Attention\\ Transfer\end{tabular} & \begin{tabular}[c]{@{}c@{}}+Sliding \\ Window\end{tabular} & PiQA          & ARC-e         & \begin{tabular}[c]{@{}c@{}}ARC-c \\ (acc. norm)\end{tabular} & \begin{tabular}[c]{@{}c@{}}HellaSwag \\ (acc. norm)\end{tabular} & Winogrande    & \begin{tabular}[c]{@{}c@{}}MMLU \\ (5-shot)\end{tabular} & Average       \\ \midrule
\cellcolor[HTML]{00FFFF}                           & \xmark                                        & \xmark                                     & 67.8          & 58.1          & 28.9                                                         & 35.8                                                             & 50.8          & 23.8                                                     & 44.2          \\
\cellcolor[HTML]{00FFFF}                           & \cmark                                        & \xmark                                     & 76.5          & 72.6          & 40.1                                                         & 65.6                                                             & 53.3          & 23.8                                                     & 55.3          \\
\cellcolor[HTML]{00FFFF}                           & \xmark                                        & \cmark                                     & 80.5          & 80.6          & 53.4                                                         & 78.8                                                             & 73.6          & 45.8                                                     & 68.8          \\
\rowcolor[HTML]{00FFFF} 
\multirow{-4}{*}{\cellcolor[HTML]{00FFFF}Hedgehog} & \cmark                                        & \cmark                                     & \textbf{80.9} & 81.7          & \textbf{54.9}                                                & \textbf{79.7}                                                    & \textbf{74.1} & \textbf{52.8}                                            & \textbf{70.7} \\ \midrule
                                                   & \xmark                                        & \xmark                                     & 62.0          & 42.1          & 24.7                                                         & 32.7                                                             & 48.3          & 23.2                                                     & 38.8          \\
                                                   & \cmark                                        & \xmark                                     & 76.1          & 72.8          & 40.8                                                         & 63.6                                                             & 52.6          & 23.1                                                     & 54.8          \\
                                                   & \xmark                                        & \cmark                                     & 54.8          & 26.3          & 26.2                                                         & 56.4                                                             & 49.6          & 23.8                                                     & 39.5          \\
\multirow{-4}{*}{T2R}                              & \cmark                                        & \cmark                                     & 80.7          & \textbf{82.0} & 54.6                                                         & 79.5                                                             & 72.2          & 40.7                                                     & 68.3          \\ \bottomrule
\end{tabular}
}
\end{center}
\vspace{-2mm}
\caption{\textbf{\name{} component ablations on individual LM Eval tasks}, linearizing Llama 3 8B, expanded view of Table~\ref{table:main_results_ablations}. \name{} default shaded. Across Hedgehog and T2R feature maps, \name{}' attention transfer and sliding window increasingly improve linearized LLM quality on popular LM Eval tasks both on average and specifically for 5-shot MMLU scores.
}
\label{table:main_results_ablations_full}
\end{table}

\subsubsection{Llama 3 8B \name{} Results Across Multiple Runs}
\label{appendix:add_exp_main_results_linearize_stdev}

In Table~\ref{table:main_results_linearize_stdev}, we report the LM Eval results for Llama 3 8B models linearized with two linear attention approaches, Hedgehog~\citep{zhang2024the} and \name{} (ours), after doing linearizing across three seeds (0, 1, 2). We report means and standard deviations (in parentheses). Across all tasks, variation across seeds is low (under 1 point, other than Winogrande), with the difference in task performance across methods being much higher (greater than 10 points).

\begin{table}[!h]
\begin{center}
\resizebox{1.0\linewidth}{!}{
\begin{tabular}{@{}lllllllll@{}}
\toprule
Llama 3 8B     & \multicolumn{1}{c}{PiQA} & \multicolumn{1}{c}{ARC-e} & \multicolumn{1}{c}{\begin{tabular}[c]{@{}c@{}}ARC-c \\ (norm)\end{tabular}} & \multicolumn{1}{c}{\begin{tabular}[c]{@{}c@{}}HellaSwag \\ (norm)\end{tabular}} & \multicolumn{1}{c}{Winogrande} & \multicolumn{1}{c}{\begin{tabular}[c]{@{}c@{}}MMLU \\ (5-shot)\end{tabular}} & \multicolumn{1}{c}{Average} & \multicolumn{1}{c}{\begin{tabular}[c]{@{}c@{}}Average \\ (no MMLU)\end{tabular}} \\ \midrule
Hedgehog       & 76.86 (0.32)             & 73.27 (0.67)              & 40.76 (0.69)                                                                & 65.77 (0.38)                                                                    & 53.42 (0.22)                   & 24.22 (0.62)                                                                 & 55.72 (0.35)                & 62.02 (0.35)                                                                     \\
LoLCATs (Ours) & \textbf{80.79 (0.11)}    & \textbf{81.62 (0.41)}     & \textbf{54.73 (0.41)}                                                       & \textbf{79.48 (0.07)}                                                           & \textbf{72.92 (1.02)}          & \textbf{52.74 (0.64)}                                                        & \textbf{70.38 (0.33)}       & \textbf{73.91 (0.29)}                                                            \\ \bottomrule
\end{tabular}
}
\end{center}
\vspace{-2mm}
\caption{\textbf{\name{} comparison with prior linearizing methods, multiple runs}; expanded view of linear attention Llama 3 8B results in  Table~\ref{table:main_results_linearize}. We report mean and (standard deviation) across three seeds, comparing \name{} and prior linear attention Hedgehog linearizing method.}
\label{table:main_results_linearize_stdev}
\end{table}

\subsection{Closing Performance Gaps with Hybridizing \name{} across Layers 
}
We now study how \name{} models perform while keeping some layers as full softmax attention. For Llama 3 8B, we keep the first 16 layers (half of all layers) as full softmax attention and only linearize the last half with the sliding window + linear attention layer for Llama 3 8B. We then just do attention transfer over Alpaca samples, instead of the two-stage attention transfer and subsequent LoRA finetuning as default in Table~\ref{tab:8b-training-details} (keeping Stage 1 details the same). In Table~\ref{table:results_hybrid_llama3_8b}, we find that keeping just 50\% softmax attention, while only doing Stage 1 transfer on 20 million tokens, substantially closes the MMLU performance gap by  94.2\% (65.8\% hybrid versus 52.8\% default \name{}, compared to Transformer 66.6\% accuracy). Furthermore, while just training on 20 million tokens (half that of the default protocol), we recover performance on all other tasks. 

\begin{table}[!h]
\begin{center}
\resizebox{1.0\linewidth}{!}{
\begin{tabular}{@{}lccccccc@{}}
\toprule
Softmax Attn. Layers                    & Tokens (B)    & PiQA          & ARC-E         & ARC-C         & HellaSwag     & WinoGrande    & MMLU          \\ \midrule
\rowcolor[HTML]{EFEFEF} 
All (Llama 3 8B baseline)               & 0             & 79.9          & 80.1          & 53.3          & 79.1          & 73.1          & 66.6          \\
0-15 (LoLCATS 50\%, Just Attn Transfer) & \textbf{0.02} & 79.5          & 80.2          & 53.4 & \textbf{79.2} & 73.6          & \textbf{65.8} \\
None (LoLCATs, Attn Transfer + LoRA)    & 0.04          & \textbf{80.9} & \textbf{81.7} & \textbf{54.9} & 79.0   & \textbf{74.1} & 52.8          \\ \bottomrule
\end{tabular}
}
\end{center}
\vspace{-2mm}
\caption{\small \textbf{Hybrid \name{}, Llama 3 8B, LM Eval}. With just 50\% softmax attention, training on 20 million tokens (half that of the default protocol) results in substantially closing the gap to full softmax attention Transformers on MMLU (+13 points absolute), and maintaining performance on other LM Eval tasks.}
\label{table:results_hybrid_llama3_8b}
\end{table}

\subsection{\name{} Evaluation for Linearizing 1B+ LLMs}
\label{appendix:add_exp_1b_eval}
We now study \name{}' performance when linearizing $\sim$ 1B parameter LLMs. We choose two popular models, Llama 3.2 1B~\citep{meta2024llama32} and Phi 1.5 1.3B~\citep{li2023textbooks}, and linearize with \name{} using the same hyperparameters, architectures, and training details (\eg{} 40M tokens over Alpaca) as Llama 3 8B (Table~\ref{tab:8b-training-details}) except for linear attention feature dimension, instead adjusting this to the 1B model head dimensions (\ie{} 0.5 $\times$ head dimension 64 = 32). To evaluate \name{} quality, we compare the LM Eval Harness performance of \name{} with other available linearizing methods on each Transformer LLM (Table~\ref{table:results_llama3_2_1b}, Table~\ref{table:results_phi1_5_1b}), and against competitive 1B subquadratic LLMs pretrained from scratch (Table~\ref{table:results_1b_llms}), reporting results from \cite{bick2024transformers}.

In all evaluation settings, \name{} demonstrates competitive or state-of-the-art linearizing quality. When first controlling for the original pretrained Transformer, \name{} is able to outperform both prior linear attentions, as well as pure and hybrid Phi-Mamba models created with MOHAWK, another method to distill Transformers into SSM-based Mamba architectures~\citep{bick2024transformers}. Notably, \name{} makes this possible while only using 1.33\% of the training tokens in MOHAWK (Table~\ref{table:results_phi1_5_1b}), and again only using parameter-efficient updates. Furthermore, by linearizing Llama 3.2 1B and Phi 1.5 1.3B, \name{} efficiently creates subquadratic 1B+ LLMs that outperform various LLMs pretrained from scratch (Table~\ref{table:results_1b_llms}). These results suggest \name{} remains a competitive option for linearizing LLMs at multiple parameter counts.

\begin{table}[t]
\begin{center}
\resizebox{1.0\linewidth}{!}{
\begin{tabular}{@{}lccccccccc@{}}
\toprule
Model                       & PiQA          & ARC-e         & \begin{tabular}[c]{@{}c@{}}ARC-c \\ (acc. norm)\end{tabular} & \begin{tabular}[c]{@{}c@{}}HellaSwag \\ (acc. norm)\end{tabular} & Winogrande    & \begin{tabular}[c]{@{}c@{}}MMLU \\ (5-shot)\end{tabular} & Avg.          & \begin{tabular}[c]{@{}c@{}}Avg. \\ (No MMLU)\end{tabular} & \begin{tabular}[c]{@{}c@{}}Avg. \% of \\ Transformer\end{tabular} \\ \midrule
\rowcolor[HTML]{EFEFEF} 
Llama 3.2 1B                & 74.4          & 65.5          & 35.8                                                         & 63.7                                                             & 60.5          & 31.9                                                     & 55.3          & 60.0                                                      & \cellcolor[HTML]{F3F3F3}-                                         \\
$\rightarrow$ T2R           & 69.2          & 58.2          & 29.9                                                         & 42.6                                                             & 54.1          & 23.3                                                     & 46.2          & 50.8                                                      & 84.7                                                              \\
$\rightarrow$ Hedgehog      & 70.1          & 55.8          & 29.8                                                         & 47.7                                                             & 50.7          & 23.0                                                     & 46.2          & 50.8                                                      & 84.7                                                              \\
$\rightarrow$ \name{} (Ours) & \textbf{74.6} & \textbf{63.0} & \textbf{35.1}                                                & \textbf{63.7}                                                    & \textbf{61.5} & \textbf{27.3}                                            & \textbf{54.2} & \textbf{59.6}                                             & \textbf{99.3}                                                     \\ \bottomrule
\end{tabular}
}
\end{center}
\vspace{-2mm}
\caption{\small \textbf{\name{} comparison with linearizing methods, Llama 3.2 1B, LM Eval}. Following the two-step linearizing procedure in \name{}, we compare the \name{} architecture against the prior Transformer-to-RNN (T2R) and Hedgehog linear attentions. At the 1B scale, \name{} substantially closes the performance gap with original Llama 3.2 1B Transformer.}
\label{table:results_llama3_2_1b}
\end{table}

\begin{table}[t]
\begin{center}
\resizebox{1.0\linewidth}{!}{
\begin{tabular}{@{}lccccccccc@{}}
\toprule
Model                       & \multicolumn{1}{l}{\begin{tabular}[c]{@{}l@{}}Training\\ Tokens (B)\end{tabular}} & PiQA          & ARC-e         & \begin{tabular}[c]{@{}c@{}}ARC-c \\ (acc. norm)\end{tabular} & \begin{tabular}[c]{@{}c@{}}HellaSwag \\ (acc. norm)\end{tabular} & Winogrande    & \begin{tabular}[c]{@{}c@{}}MMLU \\ (5-shot)\end{tabular} & \begin{tabular}[c]{@{}c@{}}Avg. \\ (No MMLU)\end{tabular} & \begin{tabular}[c]{@{}c@{}}Avg. \% \\ of Transformer\end{tabular} \\ \midrule
\rowcolor[HTML]{EFEFEF} 
Phi 1.5 1.3B (MOHAWK)       & 150                                                                               & 76.6          & 75.6          & 48.0                                                         & 62.6                                                             & 73.4          & -                                                        & 67.2                                                      & -                                                                 \\
\rowcolor[HTML]{EFEFEF} 
Phi 1.5 1.3B (Our run)      & 150                                                                               & 76.6          & 76.1          & 47.6                                                         & 62.6                                                             & 72.8          & 43.6                                                     & 67.1                                                      & -                                                                 \\
Phi-Mamba 1.5               & 3                                                                                 & 75.5          & 74.0          & 44.1                                                         & 60.2                                                             & 71.7          & -                                                        & 65.1                                                      & 96.8                                                              \\
Hybrid Phi-Mamba 1.5        & 3                                                                                 & {\ul 76.5}    & {\ul 75.3}    & {\ul 45.8}                                                   & {\ul 60.6}                                                       & {\ul 72.0}    & -                                                        & {\ul 66.0}                                                & 98.2                                                              \\
Phi 1.5 1.3B T2R            & \textbf{0.04}                                                                     & 71.0          & 69.1          & 36.6                                                         & 46.2                                                             & 53.6          & 24.3                                                     & 55.3                                                      & 82.4                                                              \\
Phi 1.5 1.3B Hedgehog       & \textbf{0.04}                                                                     & 72.7          & 70.9          & 38.0                                                         & 49.4                                                             & 54.1          & 23.5                                                     & 57.0                                                      & 85.0                                                              \\
Phi 1.5 1.3B LoLCATs (Ours) & \textbf{0.04}                                                                     & \textbf{76.9} & \textbf{77.0} & \textbf{46.9}                                                & \textbf{62.3}                                                    & \textbf{72.7} & \textbf{39.2}                                            & \textbf{67.2}                                             & 100.1                                                             \\ \bottomrule
\end{tabular}
}
\end{center}
\vspace{-2mm}
\caption{\small \textbf{\name{} comparison with linearizing methods, Phi 1.5 1.3B, LM Eval}. We compare \name{} with linearizing with prior linear attentions, and available results from~\cite{bick2024transformers}, who distill Phi 1.5B into Mamba and hybrid Mamba-Transformer architectures with their MOHAWK method (Phi-Mamba 1.5, Hybrid Phi-Mamba 1.5). \name{} similarly outperforms prior linearizing methods, closing the gap to the Transformer Phi 1.5 1.3B on PiQA, ARC-Easy and average zero-shot LM Eval (no MMLU).}
\label{table:results_phi1_5_1b}
\end{table}

\begin{table}[!h]
\begin{center}
\resizebox{1.0\linewidth}{!}{
\begin{tabular}{@{}lcccccccc@{}}
\toprule
Model                               & \begin{tabular}[c]{@{}c@{}}Training \\ Tokens (B)\end{tabular} & PiQA                                                                & ARC-e                                                               & \begin{tabular}[c]{@{}c@{}}ARC-c \\ (acc. norm)\end{tabular}        & \begin{tabular}[c]{@{}c@{}}HellaSwag \\ (acc. norm)\end{tabular}    & Winogrande                                                          & \begin{tabular}[c]{@{}c@{}}MMLU \\ (5-shot)\end{tabular}            & \begin{tabular}[c]{@{}c@{}}Avg. \\ (No MMLU)\end{tabular}           \\ \midrule
\rowcolor[HTML]{EFEFEF} 
Transformer                         & \multicolumn{1}{l}{\cellcolor[HTML]{EFEFEF}}                   & \multicolumn{1}{l}{\cellcolor[HTML]{EFEFEF}}                        & \multicolumn{1}{l}{\cellcolor[HTML]{EFEFEF}}                        & \multicolumn{1}{l}{\cellcolor[HTML]{EFEFEF}}                        & \multicolumn{1}{l}{\cellcolor[HTML]{EFEFEF}}                        & \multicolumn{1}{l}{\cellcolor[HTML]{EFEFEF}}                        & \multicolumn{1}{l}{\cellcolor[HTML]{EFEFEF}}                        & \multicolumn{1}{l}{\cellcolor[HTML]{EFEFEF}}                        \\
Pythia 1.4B                         & 300                                                            & 71.1                                                                & 60.6                                                                & 26.0                                                                & 52.1                                                                & 57.3                                                                & 26.6                                                                & 53.4                                                                \\
Llama 3.2 1B                        & 9000                                                           & 74.4                                                                & 65.5                                                                & 35.8                                                                & 63.7                                                                & 60.5                                                                & 31.9                                                                & 60.0                                                                \\
Phi 1.5 1B                          & 150                                                            & 76.6                                                                & 76.1                                                                & 47.6                                                                & 62.6                                                                & 72.8                                                                & 43.6                                                                & 67.1                                                                \\
\rowcolor[HTML]{EFEFEF} 
{\color[HTML]{333333} Subquadratic} & \multicolumn{1}{l}{\cellcolor[HTML]{EFEFEF}}                   & \multicolumn{1}{l}{\cellcolor[HTML]{EFEFEF}{\color[HTML]{333333} }} & \multicolumn{1}{l}{\cellcolor[HTML]{EFEFEF}{\color[HTML]{333333} }} & \multicolumn{1}{l}{\cellcolor[HTML]{EFEFEF}{\color[HTML]{333333} }} & \multicolumn{1}{l}{\cellcolor[HTML]{EFEFEF}{\color[HTML]{333333} }} & \multicolumn{1}{l}{\cellcolor[HTML]{EFEFEF}{\color[HTML]{333333} }} & \multicolumn{1}{l}{\cellcolor[HTML]{EFEFEF}{\color[HTML]{333333} }} & \multicolumn{1}{l}{\cellcolor[HTML]{EFEFEF}{\color[HTML]{333333} }} \\
xLSTM 1.4B                          & 300                                                            & {\ul 74.6}                                                          & 64.3                                                                & 32.6                                                                & 60.9                                                                & 60.6                                                                & -                                                                   & 58.6                                                                \\
Finch 1.6B (RWKV-v6)                & 1100                                                           & 72.6                                                                & 64.2                                                                & 34.1                                                                &     57.3                                                                 & 59.4                                                                & -                                                                   & 57.5                                                                \\
DeltaNet 1.3B                       & 100                                                            & 71.2                                                                & 57.2                                                                & 28.3                                                                & 50.2                                                                & 53.6                                                                & -                                                                   & 52.1                                                                \\
GLA 1.3B                            & 100                                                            & 71.8                                                                & 57.2                                                                & 26.6                                                                & 49.8                                                                & 53.9                                                                & -                                                                   & 51.9                                                                \\
Mamba 1 1.4B                        & 315                                                            & 74.2                                                                & {\ul 65.5}                                                          & 32.8                                                                & 59.1                                                                & 61.5                                                                & -                                                                   & 58.6                                                                \\
Mamba 2 1.3B                        & 315                                                            & 73.2                                                                & 64.3                                                                & 33.3                                                                & 59.9                                                                & 60.9                                                                & -                                                                   & 58.3                                                                \\
\rowcolor[HTML]{ECF4FF} 
Llama 3.2 1B LoLCATs (Ours)         & \textbf{0.04}                                                  & {\ul 74.6}                                                          & 63.0                                                                & {\ul 35.1}                                                          & \textbf{63.7}                                                       & {\ul 61.5}                                                          & 27.3                                                                & {\ul 59.6}                                                          \\
\rowcolor[HTML]{ECF4FF} 
Phi 1.5 1.3B LoLCATs (Ours)         & \textbf{0.04}                                                  & \textbf{76.9}                                                       & \textbf{77.0}                                                       & \textbf{46.9}                                                       & {\ul 62.3}                                                          & \textbf{72.7}                                                       & 39.2                                                                & \textbf{67.2}                                                       \\ \bottomrule
\end{tabular}
}
\end{center}
\vspace{-2mm}
\caption{\small \textbf{\name{} comparison to pretrained 1B LLMs}. \name{}-linearized Llama 3.2 1B and Phi 1.5 1.3B consistently outperform strong subquadratic 1B+ LLMs pretrained from scratch, achieving \textbf{best} or {\ul second-best} accuracy on all tasks other than ARC-easy. Subquadratic results reported from~\cite{bick2024transformers}.}
\label{table:results_1b_llms}
\end{table}

\begin{table}[t]
\begin{center}
\resizebox{1\linewidth}{!}{
\begin{tabular}{@{}c|cccccc|cc@{}}
\toprule
\multicolumn{1}{c|}{\begin{tabular}[c]{@{}c@{}}LoRA\\ Rank\end{tabular}} & PiQA          & ARC-e         & \begin{tabular}[c]{@{}c@{}}ARC-c\\ (norm)\end{tabular} & \begin{tabular}[c]{@{}c@{}}HellaSwag\\ (norm)\end{tabular} & \begin{tabular}[c]{@{}c@{}}Wino-\\ grande\end{tabular} & \begin{tabular}[c]{@{}c@{}}MMLU\\ (5-shot)\end{tabular} & Avg.          & \begin{tabular}[c]{@{}c@{}}Avg. \\ (no MMLU)\end{tabular} \\ \midrule
4                                                                        & 80.7          & \textbf{82.5} & \textbf{56.0}                                          & {\ul 79.6}                                                 & 73.6                                                   & 50.1                                                    & \textbf{71.3} & \textbf{74.5}                                             \\
8                                                                        & {\ul 80.9}    & 81.7          & {\ul 54.9}                                             & \textbf{79.7}                                              & \textbf{74.1}                                          & {\ul 52.8}                                              & 70.7          & {\ul 74.2}                                                \\
16                                                                       & 80.7          & {\ul 81.9}          & 54.5                                                   & \textbf{79.7}                                              & {\ul 73.8}                                             & 48.9                                                    & 69.9          & 74.1                                                      \\
32                                                                       & \textbf{81.1} & 81.5          & 54.5                                                   & \textbf{79.7}                                              & 72.8                                                   & 51.0                                                    & 70.1          & 73.9                                                      \\
64                                                                       & {\ul 80.9}    & 81.9          & 54.5                                                   & 79.3                                                       & 72.1                                                   & 51.7                                                    & {\ul 71.1}    & 73.8                                                      \\
128                                                                      & 80.5          & 80.9          & 52.8                                                   & 78.4                                                       & 72.2                                                   & \textbf{53.4}                                           & 69.7          & 73.0                                                      \\
256                                                                      & 80.7          & 80.2          & 52.1                                                   & 78.8                                                       & 71.4                                                   & 52.1                                                    & 69.2          & 72.7                                                      \\ \midrule
Full Finetune                                                                      & 80.6          & {\ul 81.9}          & 54.3                                                   & 79.4                                                       & 72.4                                                   & 52.1                                                    & 70.1          & 73.7                                                      \\ \bottomrule
\end{tabular}
}
\end{center}
\vspace{-2mm}
\caption{\small \textbf{LoRA rank $r$ comparison}. Evaluation on LM Evaluation Harness tasks. When adapting all projections $\bm{W}_q, \bm{W}_k, \bm{W}_v, \bm{W}_o$ after attention transfer, we find smaller ranks $r=4,8$ are surprisingly sufficient. Larger ranks with more parameter-heavy updates do not necessarily improve downstream performance.
}
\label{table:app_ablate_lora_r}
\vspace{-0.5cm}
\end{table}

\subsection{Study on Parameter-Efficient Training}
\label{appendix:add_exp_lora}
In our main results, we found that simple default initializations (\eg{} rank $8$, applied to all attention projection layers) could recover high quality linearizing while only updating $<$0.2\% of model parameters. In this section, we study how changing different aspects of low-rank adaptation and sliding window size impact linearizing performance. 

\subsubsection{Effect of LoRA Rank}
\label{appendix:add_exp_lora_rank}
We study the effect of LoRA rank in post-swap finetuning for zero-shot linearized LLM performance. Following standard implementations~\citep{hu2021lora}, we consider two factors. First: rank $r$, which determines the rank of the low-rank matrices $\bm{A},\bm{B}$ we decompose the weight deltas into. Second: alpha $\alpha$, where $\alpha / r$ is a scaling factor that controls the degree to which $\bm{B}\bm{A}$ affect the output (\ie{} LoRA output $y = \bm{W}x + \frac{\alpha}{r}\bm{B}\bm{A} x$).

\paragraph{Setup.} We sweep over ranks \{4, 8, 16, 32, 64, 128, 256\}, and adjust $\alpha$ such that $\alpha / r = 2$ as a default scaling factor~\citep{Raschka_2023}. For comparison, we also do a full finetuning run, where after attention transfer we do the stage 2 adjustment but training all Llama 3 8B parameters. For all runs, we linearize Llama 3 8B. For LoRA runs, we use the same default experimental setup as our main results, while for the full finetuning we decreased learning rate to 1e-5 due to training instability. We start with the linear + ``terrace'' window attentions, training the feature maps via attention transfer over 20M tokens over Alpaca (2 full epochs). We then freeze feature maps and apply LoRA with the above ranks on all attention weight projections (freezing all other parameters such as those in MLPs or GroupNorms), finetuning for two more epochs over Alpaca using the hyperparameters in Table~\ref{tab:8b-training-details}. We evaluate with LM Eval tasks.

\paragraph{Results.} We report results in  Table~\ref{table:app_ablate_lora_r}. We find that when applying LoRA to adjust in linearizing after attention transfer, larger rank updates,~\eg{} $r=$~128 or 256, do not necessarily lead to improved zero or few-shot downstream performance. Full finetuning also does not improve performance. Somewhat surprisingly, using just $r =$~4 leads to overall best performance, while $r =$~128 and $r =$~8 achieve best and second-best 5-shot MMLU accuracy. While we leave further exploration for future work, we hypothesize that low-rank updates may improve quality by preventing large and potentially harmful updates with the linearizing data to pretrained weights. During the second step of adjusting, if the linearizing data is not particularly diverse or large (\eg{} for efficient linearizing), LoRA can then reduce the risk of overfitting to the linearizing data and losing LLM pretrained generalization.

\subsubsection{Effect of LoRA Projection Layer}
\label{appendix:add_exp_lora_projection}

We next compare performance when applying LoRA to different weight matrices of the linear attention layers. With the same training and evaluation setup as Appendix~\ref{appendix:add_exp_lora_rank}, but fixing $r = 8, \alpha=16$, we now apply LoRA to different combinations of $\bm{W}_q, \bm{W}_k, \bm{W}_v, \bm{W}_o$ weights after swapping in attention-transferred linear attentions. We use the same combination for each layer.

We report results in Table~\ref{table:app_ablate_lora_proj}. Interestingly, when isolating for projections updated, LoRA on projections \textit{not involved} in computing layer-wise attention weights (value $\bm{W}_v$ and output $\bm{W}_o$ projections) improves quality compared to query $\bm{W}_q$ or key projections $\bm{W}_k$. Somewhat surprisingly, updating just $\bm{W}_v$ or $\bm{W}_o$ achieves comparable performance to updating all projections (\textit{c.f.}, average zeros-shot accuracy of 74.19\% when updating just $\{\bm{W}_v\}$, 74.09\% for $\{\bm{W}_o\}$ versus 74.24\% updating $\{\bm{W}_q,\bm{W}_k,\bm{W}_v,\bm{W}_o\}$). Meanwhile, updating just $\bm{W}_q$ or $\bm{W}_k$ performs significantly worse (72.68\%, 72.29\%; versus 74.24\%). This suggests much of the quality recovery in Stage-2 low-rank finetuning comes from adjusting values and outputs to the learned attention weights---as opposed to further refining attention weight computation. While best results do come from a combination of adapting either $\bm{W}_q$ or $\bm{W}_k$ with value and output projections, we may be able to achieve even more parameter-efficient linearizing---with comparable quality---by focusing on subsets with the latter.

\begin{table}[t]
\begin{center}
\resizebox{1\linewidth}{!}{
\begin{tabular}{@{}lcccccc@{}}
\toprule
LoRA Projection                          & PiQA                                   & ARC-e                                  & ARC-c (norm)                           & HellaSwag (norm)                       & Winogrande                             & Average                                \\ \midrule
$\bm{W}_q$                               & \cellcolor[HTML]{FFFFFF}79.49          & \cellcolor[HTML]{B9E3CE}81.06          & \cellcolor[HTML]{EAF7F1}51.45          & \cellcolor[HTML]{9FD8BC}79.18          & \cellcolor[HTML]{D9F0E5}72.22          & \cellcolor[HTML]{93D4B4}72.68          \\
$\bm{W}_k$                               & \cellcolor[HTML]{E1F3EA}79.82          & \cellcolor[HTML]{FFFFFF}79.80          & \cellcolor[HTML]{FFFFFF}50.68          & \cellcolor[HTML]{FFFFFF}78.54          & \cellcolor[HTML]{C1E6D4}72.61          & \cellcolor[HTML]{A0D9BD}72.29          \\
$\bm{W}_v$                               & \cellcolor[HTML]{8FD2B1}80.69          & \cellcolor[HTML]{68C296}82.49          & \cellcolor[HTML]{5CBD8E}56.66          & \cellcolor[HTML]{90D2B2}79.28          & \cellcolor[HTML]{F1FAF6}71.82          & \cellcolor[HTML]{62C092}74.19          \\
$\bm{W}_o$                               & \cellcolor[HTML]{C7E9D8}80.09          & \cellcolor[HTML]{97D5B7}81.65          & \cellcolor[HTML]{78C9A1}55.63          & \cellcolor[HTML]{85CEAA}79.35          & \cellcolor[HTML]{7ECBA5}73.72          & \cellcolor[HTML]{66C194}74.09          \\ \midrule
$\bm{W}_q, \bm{W}_k$                     & \cellcolor[HTML]{E6F5EE}79.76          & \cellcolor[HTML]{B1E0C9}81.19          & \cellcolor[HTML]{F6FCF9}51.02          & \cellcolor[HTML]{8ED2B0}79.29          & \cellcolor[HTML]{D9F0E5}72.22          & \cellcolor[HTML]{93D3B4}72.70          \\
$\bm{W}_q, \bm{W}_v$                     & \cellcolor[HTML]{76C89F}80.96          & \cellcolor[HTML]{89D0AD}81.90          & \cellcolor[HTML]{9BD7BA}54.35          & \cellcolor[HTML]{B8E3CE}79.01          & \cellcolor[HTML]{D4EEE1}72.30          & \cellcolor[HTML]{72C69D}73.70          \\
$\bm{W}_q, \bm{W}_o$                     & \cellcolor[HTML]{57BB8A}\textbf{81.28} & \cellcolor[HTML]{8CD1AF}81.86          & \cellcolor[HTML]{8FD2B1}54.78          & \cellcolor[HTML]{9CD7BA}79.20          & \cellcolor[HTML]{7ECBA5}73.72          & \cellcolor[HTML]{63C093}74.17          \\
$\bm{W}_k, \bm{W}_v$                     & \cellcolor[HTML]{8AD0AE}80.74          & \cellcolor[HTML]{61BF91}82.62          & \cellcolor[HTML]{57BB8A}\textbf{56.83} & \cellcolor[HTML]{93D3B4}79.26          & \cellcolor[HTML]{E8F6EF}71.98          & \cellcolor[HTML]{60BF90}74.28          \\
$\bm{W}_k, \bm{W}_o$                     & \cellcolor[HTML]{8FD2B1}80.69          & \cellcolor[HTML]{7BCAA3}82.15          & \cellcolor[HTML]{7DCBA5}55.46          & \cellcolor[HTML]{6AC397}79.53          & \cellcolor[HTML]{A9DCC3}73.01          & \cellcolor[HTML]{FFFFFF}69.33          \\
$\bm{W}_v, \bm{W}_o$                     & \cellcolor[HTML]{8FD2B1}80.69          & \cellcolor[HTML]{57BB8A}\textbf{82.79} & \cellcolor[HTML]{71C69C}55.89          & \cellcolor[HTML]{64C193}79.57          & \cellcolor[HTML]{FFFFFF}71.59          & \cellcolor[HTML]{65C194}74.10          \\ \midrule
$\bm{W}_q, \bm{W}_k, \bm{W}_v$           & \cellcolor[HTML]{7BCAA3}80.90          & \cellcolor[HTML]{78C9A1}82.20          & \cellcolor[HTML]{61BF91}56.48          & \cellcolor[HTML]{A6DBC1}79.13          & \cellcolor[HTML]{70C59B}73.95          & \cellcolor[HTML]{57BB8A}\textbf{74.53} \\
$\bm{W}_q, \bm{W}_k, \bm{W}_o$           & \cellcolor[HTML]{A9DDC3}80.41          & \cellcolor[HTML]{C2E7D5}80.89          & \cellcolor[HTML]{A0D9BD}54.18          & \cellcolor[HTML]{9FD8BC}79.18          & \cellcolor[HTML]{57BB8A}\textbf{74.35} & \cellcolor[HTML]{6FC59B}73.80          \\
$\bm{W}_k, \bm{W}_v, \bm{W}_o$           & \cellcolor[HTML]{AEDEC7}80.36          & \cellcolor[HTML]{72C69D}82.32          & \cellcolor[HTML]{94D4B5}54.61          & \cellcolor[HTML]{A6DBC1}79.13          & \cellcolor[HTML]{88CFAC}73.56          & \cellcolor[HTML]{69C296}74.00          \\
$\bm{W}_q, \bm{W}_k, \bm{W}_v, \bm{W}_o$ & \cellcolor[HTML]{80CCA7}80.85          & \cellcolor[HTML]{93D4B4}81.73          & \cellcolor[HTML]{8DD1B0}54.86          & \cellcolor[HTML]{57BB8A}\textbf{79.65} & \cellcolor[HTML]{66C195}74.11          & \cellcolor[HTML]{61BF91}74.24          \\ \bottomrule
\end{tabular}
}
\end{center}
\vspace{-2mm}
\caption{\small \textbf{LoRA projection comparison}. Evaluation on zero-shot LM Evaluation Harness tasks. We apply LoRA with the same rank $r = 8$ to different combinations of the attention projections, shading scores by increasing quality (darker is better). 
When isolating for projections updated, LoRA on projections \textit{not involved} in layer-wise attention weight computations (value $\bm{W}_v$ and output $\bm{W}_o$ projections) achieves higher quality over query $\bm{W}_q$ or key projections $\bm{W}_k$. This suggests $\bm{W}_v,\bm{W}_o$ may be more important to adapt after attention transfer, although best results involve a combination across attention weight and output projections. 
}
\label{table:app_ablate_lora_proj}
\end{table}



\begin{table}[t]
\begin{center}
\resizebox{1\linewidth}{!}{
\begin{tabular}{@{}c|cccccc|cc@{}}
\toprule
\begin{tabular}[c]{@{}c@{}}Window\\ Size\end{tabular} & PiQA          & ARC-e         & \begin{tabular}[c]{@{}c@{}}ARC-c\\ (norm)\end{tabular} & \begin{tabular}[c]{@{}c@{}}HellaSwag\\ (norm)\end{tabular} & \begin{tabular}[c]{@{}c@{}}Winogrande\end{tabular} & \begin{tabular}[c]{@{}c@{}}MMLU\\ (5-shot)\end{tabular} & Avg.          & \begin{tabular}[c]{@{}c@{}}Avg. \\ (no MMLU)\end{tabular} \\ \midrule
4                                                     & 80.7          & 81.4          & 55.8                                                   & 76.6                                                       & 72.1                                                   & 40.8                                                    & 67.9          & 73.3                                                      \\
16                                                    & 80.5          & \textbf{82.2} & \textbf{56.0}                                          & 78.2                                                       & 73.9                                                   & \textbf{50.3}                                           & 70.2          & 74.1                                                      \\
64                                                    & \textbf{80.7} & 81.7          & 54.7                                                   & \textbf{79.1}                                              & \textbf{75.3}                                          & \textbf{50.3}                                           & \textbf{70.3} & \textbf{74.3}                                             \\
256                                                   & 80.6          & 81.8          & 55.0                                                   & 75.6                                                       & 74.9                                                   & 41.5                                                    & 68.3          & 73.6                                                      \\ \midrule
TK 64                                                   & 80.9 &	81.7 &	54.9 &	79.7 &	74.1 &	52.8 &	70.7 &	74.2                                                    \\
\bottomrule
\end{tabular}
}
\end{center}
\vspace{-2mm}
\caption{\small \textbf{Window size comparison}. We ablate the window size in \name{} linear + sliding window attention. For each window size $w$, the layer applies softmax attention to the $w$-most recent positions, combined with Hedgehog linear attention applied for all prior positions (Eq.~\ref{eq:hybrid_linear_swa}, Figure~\ref{fig:hybrid_attn}). We compare $w \in \{4, 16, 64, 256\}$ using the standard sliding window implementation with our default $w = 64$ terraced window setup motivated by ThunderKittens (TK 64). Window size 64 performs best, where both implementations perform comparably. 
}
\label{table:app_ablate_window_size}
\end{table}

\subsubsection{Effect of Window Size}
We now compare model performance using different window sizes in the \name{} linear + sliding window attention layer. With the standard sliding window implementation (Fig~\ref{fig:hybrid_attn}), we compare LM Eval performance after linearizing with window sizes $w \in \{4, 16, 64, 256\}$. We also compare against the ThunderKittens-motivated ``terraced'' implementation used in our main experiments with window size $64$. 
In Table~\ref{table:app_ablate_window_size}, we find that in each of these settings, having more softmax attention generally improves performance (\textit{c.f.}, $w =$ 16,64 versus $w=$4). However, more softmax attention does not always lead to better quality. Window size 256 results in up to an 8.8 point drop in 5-shot MMLU accuracy, suggesting we may not necessarily trade-off more softmax attention for higher quality. Comparing the standard sliding window and terracing implementations, we find similar performance (70.3 versus 70.7 average accuracy across tasks).

\subsection{Study on Linearizing Data}
While most of our work focuses on architecture and training procedure for improving LLM linearizing quality, we now study how data selection affects \name{} performance.


\subsubsection{Data Source: Alpaca versus RedPajama}
\label{app:ablate_alpaca_vs_rp}

We study the effect of linearizing data for downstream LLM performance. While we initially found that just using the $\sim$~50K samples of a cleaned Alpaca dataset\footnote{\url{https://huggingface.co/datasets/yahma/alpaca-cleaned}}~\citep{alpaca} could lead to surprisingly high performance on popular zero-shot LM Eval tasks, prior linearizing works~\citep{mercat2024linearizing} use more typical pretraining datasets to linearize such as RefinedWeb~\citep{penedo2023refinedweb}. We thus also try linearizing with a random subset of RedPajama~\citep{together2023redpajama} to evaluate how \name{} works with pretraining data, albeit without any special curation. For both setups, we pack samples into 1024 token sequences and randomly subsample the RedPajama data so that we use the same number of training tokens (20M) for both attention transfer and finetune stages (40M tokens overall). We use the setup as described in Appendix~\ref{app:experimenta_details_linearizing_7b} for all other hyperparameters.

In Table~\ref{table:app_ablate_data}, we find that across Mistral 7B (v0.1) and Llama 3 8B, using the Alpaca cleaned dataset actually leads to better downstream task quality for all tasks except for 5-shot MMLU, where linearizing with RedPajama consistently leads to $\sim$~2 percentage point improvements. \name{} with both of these datasets leads to comparable or higher performance than prior methods trained on 2500$\times$ the data (\textit{c.f.}, Table~\ref{table:main_results_linearize}; SUPRA trained on 100B tokens gets Avg. accuracy of 64.0\%), suggesting that \name{} can robustly improve linearizing quality over different data sources.

\begin{table}[t]
\begin{center}
\resizebox{1\linewidth}{!}{
\begin{tabular}{@{}l|c|cccccc|cc@{}}
\toprule
Model                     & \begin{tabular}[c]{@{}c@{}}Linearizing \\ Dataset\end{tabular} & PiQA          & ARC-e         & \begin{tabular}[c]{@{}c@{}}ARC-c\\ (acc. norm)\end{tabular} & \begin{tabular}[c]{@{}c@{}}HellaSwag\\ (acc. norm)\end{tabular} & \begin{tabular}[c]{@{}c@{}}Wino-\\ grande\end{tabular} & \begin{tabular}[c]{@{}c@{}}MMLU\\ (5-shot)\end{tabular} & Avg.          & \begin{tabular}[c]{@{}c@{}}Avg. \\ (no MMLU)\end{tabular} \\ \midrule
\rowcolor[HTML]{FCE5CD} 
Mistral 7B (v0.1)                & -                                                           & 82.1          & 80.9          & 53.8                                                        & 81.0                                                            & 74.0                                                   & 62.4                                                    & 72.4          & 74.4                                                      \\
$\rightarrow$ LoLCATs  & Alpaca Clean                                                & \textbf{81.5} & \textbf{81.7} & \textbf{54.9}                                               & \textbf{80.7}                                                   & \textbf{74.0}                                          & 51.4                                                    & \textbf{70.7} & \textbf{74.5}                                             \\
$\rightarrow$ LoLCATs  & RedPajama                                                   & 80.1          & 77.6          & 49.0                                                        & 80.3                                                            & 71.7                                                   & \textbf{53.2}                                           & 68.6          & 71.7                                                      \\ \midrule
\rowcolor[HTML]{CFE2F3} 
Llama 3 8B                & -                                                           & 79.9          & 80.1          & 53.3                                                        & 79.1                                                            & 73.1                                                   & 66.6                                                    & 72.0          & 73.1                                                      \\

$\rightarrow$ LoLCATs  & Alpaca Clean                                                & \textbf{80.9} & \textbf{81.7} & \textbf{54.9}                                               & \textbf{79.7}                                                   & \textbf{74.1}                                          & 52.8                                                    & \textbf{70.7} & \textbf{74.2}                                             \\
$\rightarrow$ LoLCATs  & RedPajama                                                   & 78.9          & 79.0          & 52.0                                                        & 78.1                                                            & 72.6                                                   & \textbf{55.2}                                           & 69.3          & 72.1                                                      \\ \bottomrule
\end{tabular}
}
\end{center}
\vspace{-2mm}
\caption{\small \textbf{Linearizing data comparison}. For linearizing Mistral 7B (v0.1) and Llama 3 8B, \name{} with Alpaca and RedPajama subsets perform comparably (c.f. prior methods,~Table~\ref{table:main_results_linearize}), though we find that Alpaca actually leads to higher accuracy for most tasks other than 5-shot MMLU.
}
\label{table:app_ablate_data}
\vspace{-0.5cm}
\end{table}

\subsubsection{Matching Linearizing Data to Downstream (Retrieval) Task}
\label{app:ablate_alpaca_vs_commonsenseqa}

Among LM Eval tasks, we note a sizable gap between linearized and Transformer-based LLMs on MMLU. We hypothesize one source of this gap is due to MMLU's evaluation setup, which not only tests for knowledge recall in pretrained weights, but also \textit{retrieval} over the input context~\citep{hendrycks2020measuring}. In the default 5-shot multiple choice setup, models must be able to retrieve and produce the letter associated with the right answer choice in context. However, prior works have shown that linear attentions and non-softmax attention models perform worse on retrieval, both explicitly on MMLU~\citep{waleffe2024empirical} and in retrieval tasks at large~\citep{waleffe2024empirical,shen2024scaling}.

To counteract these effects, we study if linearizing LLMs with data that explicitly reflects the target downstream task can improve performance. We report two such settings next. 

\paragraph{Improving MMLU.} First, to test improving performance on MMLU, we linearize with additional data from CommonsenseQA~\citep{talmor-etal-2019-commonsenseqa} (CQA), another multiple-choice dataset. We construct a linearizing dataset by using the same 5-shot in-context template as in MMLU, \ie{} \\
\texttt{<bos><question 1><answer choices 1><answer 1>, <question 2><answer choices 2><answer 2>, \ldots, <question 5><answer choices 5>\_\_\_<eos>}, where the next token to predict \texttt{\_\_\_} corresponds to the correct answer among \texttt{<answer choices 5>}, using the $\sim$10k samples of the CQA training set. For both stages, we linearize Llama 3 8B over the combined set of 5-shot CQA samples and our default $\sim$50k Alpaca samples, comparing against just Alpaca or CQA samples. We pack all samples to context length 1024. 

In Table~\ref{table:ablate_alpaca_vs_cqa_mmlu}, we report the MMLU scores following each linearizing data choice. We find that adding a small amount of multiple-choice samples can substantially improve MMLU accuracy ($\sim$ 2 points). However, CQA alone performs $\sim$ 10 points worse than Alpaca. This suggests quantity and diversity of samples may still be necessary for linearizing; we break this down further and study the amount of linearizing needed during both attention transfer and low-rank adjusting in Appendix~\ref{app:add_exp_token_budget}.

\begin{table}[!h]
\centering
\begin{tabular}{@{}c|ccc|c@{}}
\toprule
Model              & \multicolumn{3}{c}{Llama 3 8B LoLCATs} & \multicolumn{1}{|l}{Llama 3 8B} \\ 
Linearizing Data               & Alpaca     & Alpaca + CQA    & CQA     & N/A                              \\ \midrule
MMLU (5-shot) Acc. & 52.8       & \textbf{54.5}            & 43.9    & 66.6                           \\ \bottomrule
\end{tabular}
\caption{\small \textbf{MMLU comparison with task-specific linearizing data}. Adding multiple-choice samples (CommonsenseQA; CQA) to linearizing training data modestly improves downstream MMLU performance. However, CQA alone is insufficient to achieve competitive quality.}
\label{table:ablate_alpaca_vs_cqa_mmlu}
\vspace{-0.25cm}
\end{table}

\begin{table}[b]
\vspace{-0.25cm}
\begin{center}
\resizebox{1.0\linewidth}{!}{
\begin{tabular}{@{}lcccccccccc@{}}
\toprule
Passkey Placement            & 0-10\%          & 10-20\%         & 20-30\%         & 30-40\%         & 40-50\%         & 50-60\%         & 60-70\%         & 70-80\%         & 80-90\%         & 90-100\%        \\ \midrule
Llama 3 8B                   & \textbf{100.00} & \textbf{100.00} & \textbf{100.00} & \textbf{100.00} & 92.86           & \textbf{100.00} & 94.44           & 94.12           & 92.31           & \textbf{100.00} \\
Llama 3 8B Instruct          & \textbf{100.00} & \textbf{100.00} & \textbf{100.00} & \textbf{100.00} & \textbf{100.00} & \textbf{100.00} & \textbf{100.00} & \textbf{100.00} & \textbf{100.00} & \textbf{100.00} \\
Llama 3 8B (Alpaca)          & \textbf{100.00} & \textbf{100.00} & \textbf{100.00} & \textbf{100.00} & \textbf{100.00} & \textbf{100.00} & \textbf{100.00} & \textbf{100.00} & \textbf{100.00} & \textbf{100.00} \\
LoLCATs Llama 3 8B (Alpaca)  & 0.00            & 0.00            & 0.00            & 0.00            & 0.00            & 0.00            & 0.00            & 0.00            & 0.00            & 0.00            \\
LoLCATs Llama 3 8B (Passkey) & \textbf{100.00} & \textbf{100.00} & \textbf{100.00} & \textbf{100.00} & \textbf{100.00} & \textbf{100.00} & \textbf{100.00} & \textbf{100.00} & \textbf{100.00} & \textbf{100.00} \\ \bottomrule
\end{tabular}
}
\end{center}
\vspace{-2mm}
\caption{\small \textbf{8K Context Passkey Retrieval}. We report passkey retrieval accuracy for various Llama 3 8B models, using 8192-token samples and binning by passkey placement decile. Linearizing with just packed Alpaca samples fails, but using passkey samples as the linearizing data recovers 100\% retrieval performance.}
\label{table:passkey_8k}
\vspace{-0.25cm}
\end{table}

\paragraph{Evaluating and improving needle-in-a-haystack retrieval.} Next, to further test linearizing performance on downstream tasks, we evaluate the \name{}-linearized Llama 3 8B LLM on  ``needle-in-a-haystack'' tasks. We use the passkey-retrieval task setup introduced in \cite{mohtashami2023landmark} and evaluated in various prior works~\citep{chen2023extending,chen2023longlora,tworkowski2024focused}, where a model must retrieve a hidden passkey
uniform randomly placed inside a text span. We use randomly generated prompts and 5-digit passkeys (Listing~\ref{lst:passkey_template}), and test correctness by whether the model outputs the passkey exactly (exact-match). We evaluate over various text lengths (2048 to 10240 tokens), and report accuracy binned by which decile the passkey occurs in. 

Like in our MMLU study, we evaluate \name{} with two linearizing data setups. First, we use the default Alpaca data, but this time test linearizing Llama 3 8B at its max context length by packing the 50K Alpaca samples into 8192-token chunks. Second, to see if linearizing explicitly with retrieval data helps, we generate 10K passkey retrieval samples with $\sim$8192 tokens, and use these samples for both stages of \name{} linearizing. For both, we use the same hyperparameters as Table~\ref{tab:8b-training-details}.

\begin{lstlisting}[caption=Passkey Retrieval Prompt Template,label={lst:passkey_template},float=tp,
  floatplacement=tbp,]
There is an important piece of info hidden inside a lot of irrelevant 
text. Find it and memorize it. I will quiz you about the important 
information there.
The grass is green. The sky is blue. The sun is yellow. Here we go. There 
and back again. The grass is green. The sky is blue. The sun is yellow. 
Here we go. There and back again. The grass is green. The sky is blue. 
The sun is yellow. Here we go. There and back again...
...  <EVEN MORE FILLER>
The pass key is <PASSKEY>. Remember it. <PASSKEY> is the pass key. 
The grass is green. The sky is blue. The sun is yellow. Here we go. There 
and back again. The grass is green. The sky is blue. The sun is yellow. 
Here we go. There and back again...
...  <MORE FILLER>
What is the pass key? The pass key is
\end{lstlisting}
\vspace{-0.125cm}

In Table~\ref{table:passkey_8k}, we first report the results for Llama models evaluated on 8192-token prompts. We compare the \name{} Llama models against the base Llama 3 8B, the instruction-tuned version (Llama 3 8B Instruct), and a Transformer Llama 3 8B LoRA-finetuned on Alpaca data (Llama 3 8B (Alpaca); using the same LoRA parameters and second stage training procedure as in Table~\ref{tab:8b-training-details}). We find that with \name{} models, linearizing data plays a particularly strong role in downstream performance. As a potential drawback of \name{}, when just using packed Alpaca samples, the linearized Llama 3 8B fails to get even a single passkey retrieval sample correct (\name{} Llama 3 8B (Alpaca)). Meanwhile, LoRA-finetuning the non-linearized Llama 3 8B maintains high passkey retrieval accuracy. However, linearizing with passkey samples (\name{} Llama 3 8B (Passkey)) recovers 100\% accuracy. This suggests that with \name{} linearizing, the linear + sliding window attention is able to do passkey retrieval with similar performance to full softmax attention models. 

\begin{figure}[t]
    \centering
    \begin{subfigure}[b]{0.49\textwidth}
        \centering
        \includegraphics[width=\textwidth]{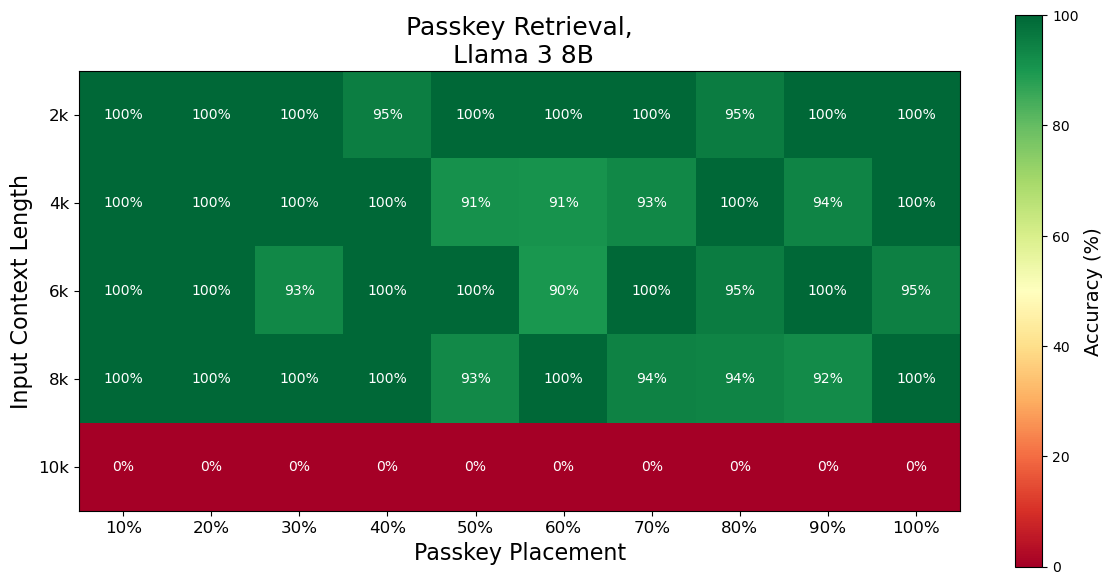}
        \label{fig:passkey_llama3_8}
    \end{subfigure}
    \begin{subfigure}[b]{0.49\textwidth}
        \centering
        \includegraphics[width=\textwidth]{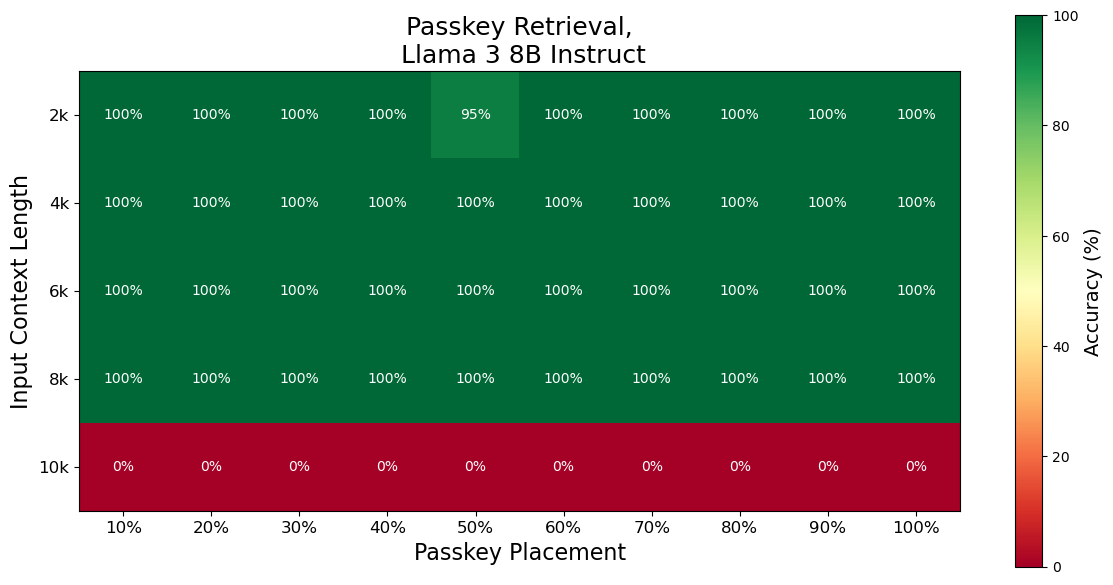}
        \label{fig:passkey_llama3_8b_inst}
    \end{subfigure}
    \begin{subfigure}[b]{0.49\textwidth}
        \centering
        \includegraphics[width=\textwidth]{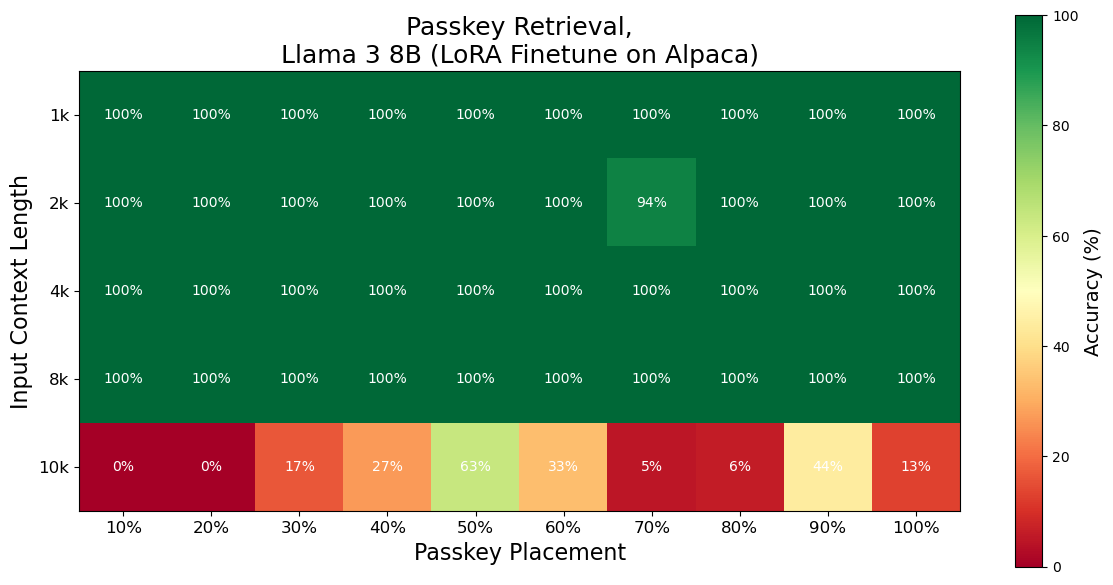}
        \label{fig:passkey_llama3_8_alpaca}
    \end{subfigure}
    \begin{subfigure}[b]{0.49\textwidth}
        \centering
        \includegraphics[width=\textwidth]{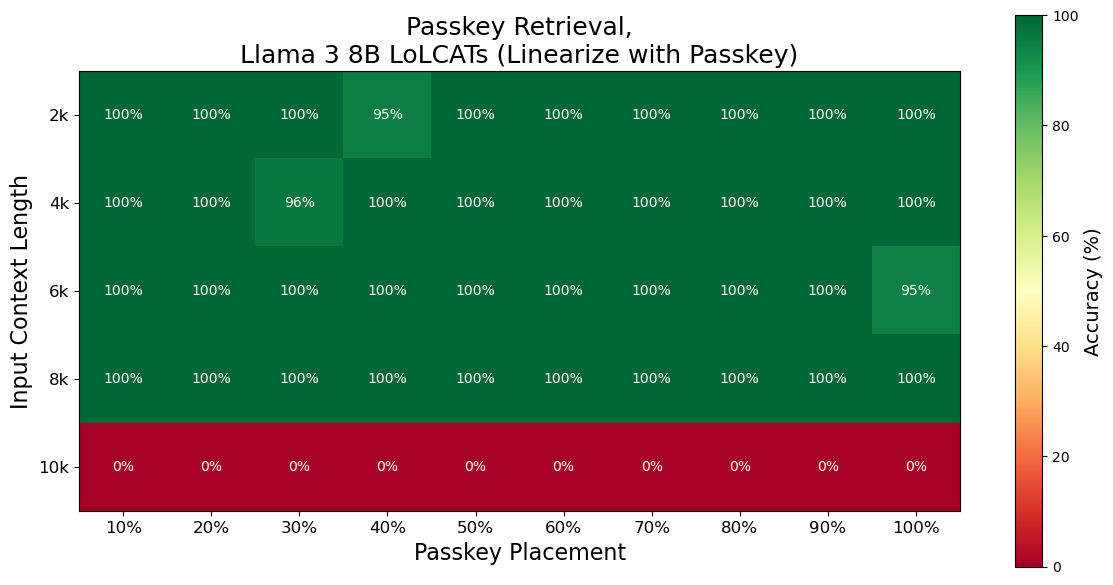}
        \label{fig:passkey_llama3_8_lolcats}
    \end{subfigure}
    \vspace{-0.5cm}
    \caption{\small \textbf{Needle-in-a-haystack with passkey retrieval}. \name{}-linearized Llama 3 8B with passkey retrieval linearizing data results in comparable retrieval to original Llama 3 8B and instruction-tuned variants.}
    \label{fig:passkey_niah}
    \vspace{-0.25cm}
\end{figure}

We further test the robustness of this retrieval across various context lengths (2048 to 10240 tokens) in Figure~\ref{fig:passkey_niah}. We report accuracies across input context lengths and passkey placements, finding similar strong retrieval performance for samples under Llama 3 8B's 8192 context length. Interestingly, only the Transformer Llama 3 8B LoRA-finetuned on Alpaca is able to do retrieval over longer context samples (10240 tokens).

\subsubsection{Sample Lengths: Effect of Effective Sequence Lengths}
We further study the impact of sample sequence length for linearizing quality. 
By default, for linearizing data we pack original data samples into sequences of consistent length, \eg{} 1024 tokens. As done in prior work~\citep{raffel2020exploring}, this allows us to pack multiple short data samples together into longer training sequences, improving training efficiency and removing any padding tokens. However, it may also introduce situations where our linearizing sequences only carry short-context dependencies, \ie{} because we pack together many samples with few tokens, or split longer samples into multiple sequences. Especially with attention transfer, linearized LLMs may model longer samples less well (\eg{} the 5-example in-context samples in 5-shot MMLU) because we never learn to approximate attentions over ``long enough'' sequence lengths. 

\paragraph{Effective sequence length.} To study this data effect, we define an ``effective sequence length'' (ESL) metric. This roughly captures for each query how far back a layer needs to attend to capture all non-zero softmax attention weights. For query at position $i$, we define the ESL per query as
\begin{equation}
    \text{ESL}({\bm{q}_i}) := \sum_{j=1}^{i} (i - j) \frac{\exp(\bm{q}_i ^\top \bm{k}_j / \sqrt{d})}{\sum_{m\leq j} \exp(\bm{q}_i ^\top \bm{k}_m / \sqrt{d})}
\end{equation}
We compute a sample's ESL per head as the sum over all query ESLs, \textit{i.e.,} $\sum_{i=1}^n\text{ESL}(\bm{q}_i)$ for a sample with $n$ tokens. We average this over all heads and layers to measure a sample's overall ESL.

We hypothesize that if our linearizing data only has samples with shorter ESL than those encountered at test time, then we would poorly model these test samples. Conversely, we may be able to improve linearizing quality by specifically filtering for samples with longer ESL. We report two findings next.

\begin{figure}[h]
    \centering
    \begin{subfigure}[b]{0.49\textwidth}
        \centering
        \includegraphics[width=\textwidth]{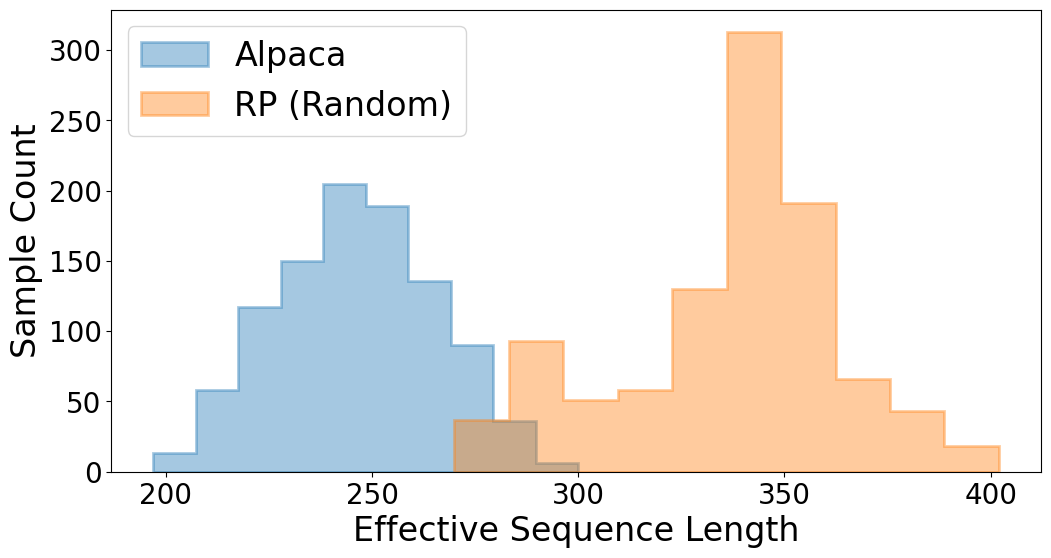}
        \label{fig:attn_maps_llama_softmax}
    \end{subfigure}
    \begin{subfigure}[b]{0.49\textwidth}
        \centering
        \includegraphics[width=\textwidth]{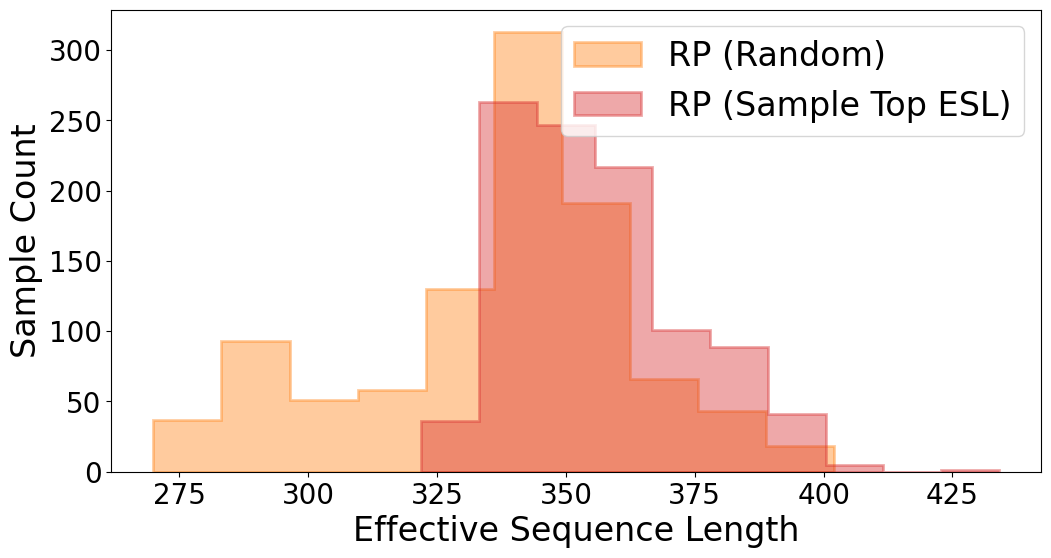}
        \label{fig:attn_maps_mistral_softmax}
    \end{subfigure}
    \vspace{-0.5cm}
    \caption{\small \textbf{Effective sequence length distributions}. Although we pack data samples into 1024-token training sequences, different data sources can vary in effective sequence lengths (left). Furthermore, we can selectively filter for longer ESL samples (right). We find linearizing with longer ESLs coincides with improved MMLU scores, albeit at the cost of other LM Eval tasks (Table~\ref{table:app_esl}).}
    \label{fig:esl}
\end{figure}

\begin{table}[b]
\begin{center}
\resizebox{1\linewidth}{!}{
\begin{tabular}{@{}lccccccc@{}}
\toprule
Linearizing Data           & PiQA          & ARC-easy      & ARC-challenge & \begin{tabular}[c]{@{}c@{}}HellaSwag\\ (acc. norm)\end{tabular} & \begin{tabular}[c]{@{}c@{}}Winogrande\\ (acc. norm)\end{tabular} & \begin{tabular}[c]{@{}c@{}}MMLU \\ (5-shot)\end{tabular} & \begin{tabular}[c]{@{}c@{}}Avg.\\ (no MMLU)\end{tabular} \\ \midrule
Alpaca                     & \textbf{80.9} & \textbf{81.7} & \textbf{54.9} & \textbf{79.7}                                                   & \textbf{74.1}                                                    & 52.8                                                     & \textbf{74.2}                                            \\
RedPajama                  & 78.9          & 79.0          & 52.0          & 78.1                                                            & 72.6                                                             & 55.2                                                     & 72.1                                                     \\
RedPajama (Sample Top ESL) & 78.4          & 77.0          & 49.8          & 78.0                                                            & 71.4                                                             & \textbf{56.5}                                            & 70.9                                                     \\ \bottomrule
\end{tabular}
}
\end{center}
\vspace{-2mm}
\caption{\small \textbf{Effect of ESL on linearized LLM quality}, Llama 3 8B. While linearizing with longer ESLs---\eg{} RedPajama samples or specifically filtering for top ESLs in RedPajama (Sample Top ESL)---improves 5-shot MMLU accuracy up to 3.7 points, it reduces quality on all other evaluated LM Eval tasks by 1.7 to 5.0 points.  
}
\label{table:app_esl}
\vspace{-0.5cm}
\end{table}


\paragraph{Finding 1: ESL corresponds with RedPajama versus Alpaca performance.} In Figure~\ref{fig:esl}, we first plot the distribution of sample ESLs computed with Llama 3 8B on Alpaca and RedPajama linearizing data subsets. We find RedPajama samples on average display longer ESLs, which conincides with improved MMLU score (\textit{c.f.}, Table~\ref{app:ablate_alpaca_vs_rp}). 

\paragraph{Finding 2: Filtering for higher ESL improves MMLU.} Furthermore, we can increase the ESLs in linearizing data by actively filtering for high ESL samples (Figure~\ref{fig:esl} right). Here we actively filter for the top 20,000 packed RedPajama samples with the highest ESLs, amounting to 20M tokens. When doing one epoch of attention transfer and low-rank linearizing with this subset, we further improve MMLU accuracy by 1.3 points (Table~\ref{table:app_esl}). However, this comes at a cost for all other LM Eval tasks, dropping quality compared to random RedPajama packing by 0.2 to 2.1 points.






\subsection{Study on Linearizing Token Budget}
\label{app:add_exp_token_budget}

We further study how varying the number of tokens used for both attention transfer and low-rank adaptation impacts \name{} linearizing quality. 

\begin{figure}[h]
    \centering
    \begin{subfigure}[b]{0.32\textwidth}
        \centering
        \includegraphics[width=1\textwidth]{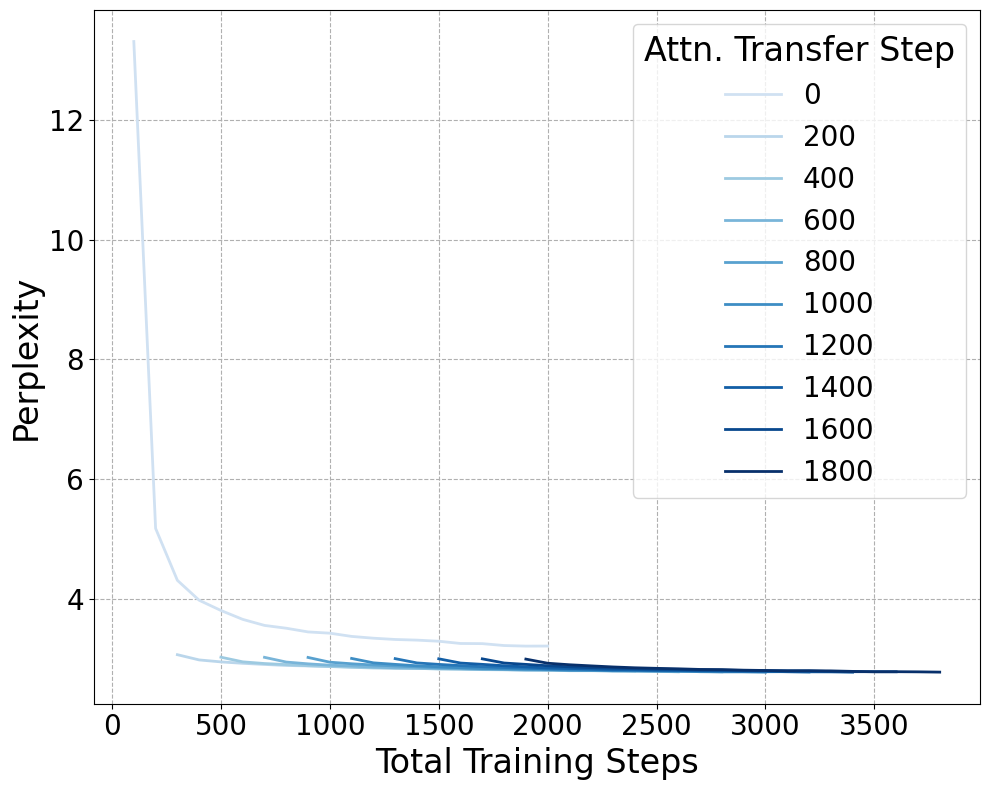}
        \caption{Alpaca PPL}
        \label{fig:app_ablate_transfer_steps_ppl}
    \end{subfigure}
    \begin{subfigure}[b]{0.32\textwidth}
        \centering
        \includegraphics[width=1\textwidth]{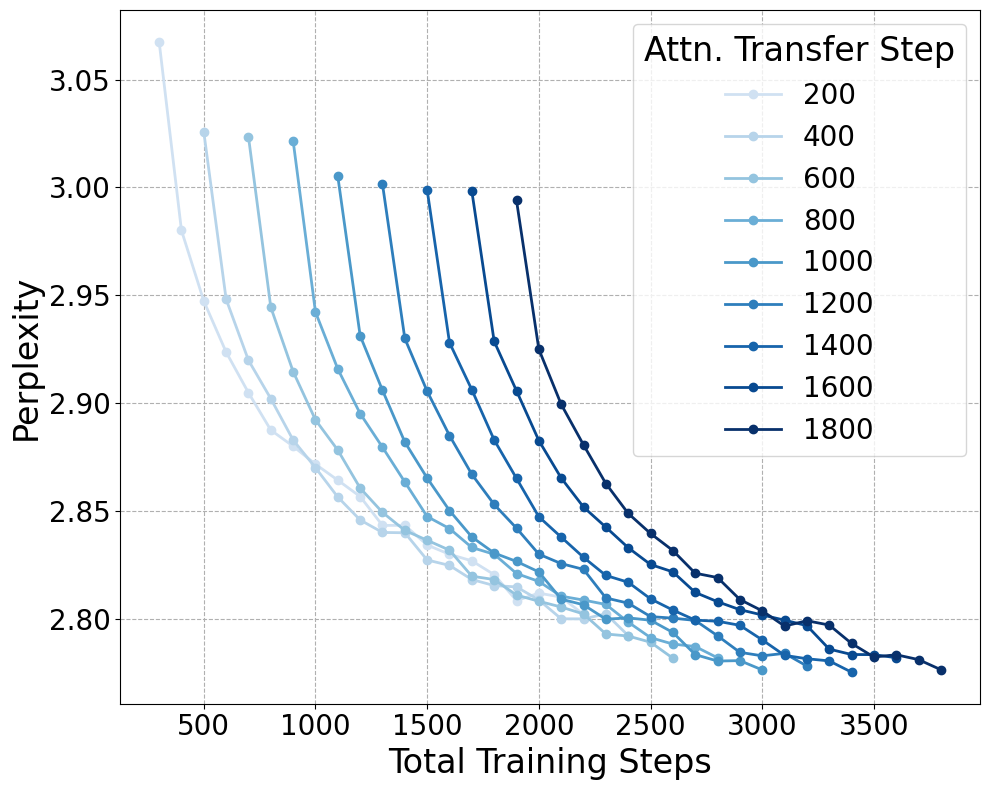}
        \caption{Alpaca Val. PPL (Zoomed)}
        \label{fig:app_ablate_transfer_steps_ppl_zoomed}
    \end{subfigure}
    \begin{subfigure}[b]{0.32\textwidth}
        \centering
        \includegraphics[width=1\textwidth]{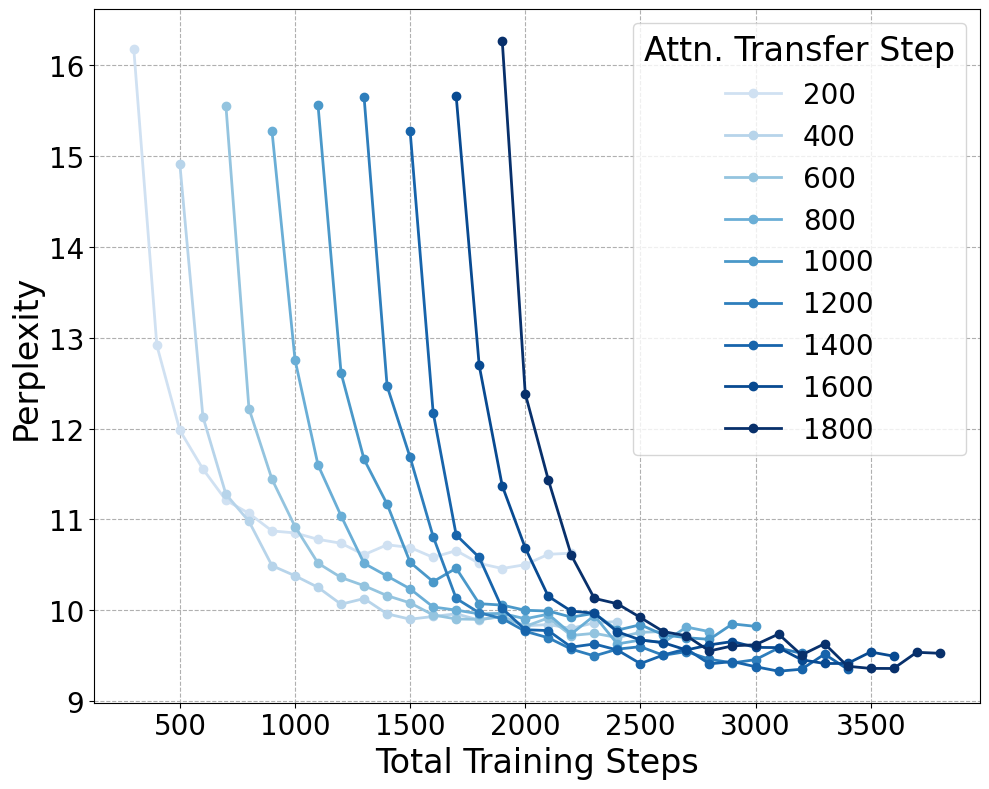}
        \caption{RedPajama Val. PPL (Zoomed)}
        \label{fig:app_ablate_transfer_steps_ppl_rp}
    \end{subfigure}
    \vspace{-0.25cm}
    \caption{\small \textbf{Evaluation curves over number of training steps}. Llama 3 8B linearized with Alpaca. We report the impact of steps allotted to attention transfer versus LoRA linearizing, using validation set perplexity (PPL) over both in-distribution (held-out Alpaca samples) and out-of-distribution (RedPajama validation samples). We first run linearizing with 0 - 1800 attention transfer steps, before LoRA-finetuning for up to 2000 steps. Without any attention transfer (0 steps), linearized LLMs get much higher perplexity (Fig~\ref{fig:app_ablate_transfer_steps_ppl}). On the other hand, we observe similar convergence after attention transfer over only 1000 steps.}
    \label{fig:app_ablate_transfer_steps}
\end{figure}

\textbf{Impact of minimal tokens.} To first test how efficient we can be with attention transfer, we linearize Llama 3 8B with varying numbers of attention transfer steps (0 - 1800), before low-rank adjusting for up to 2000 steps. We use the Alpaca dataset and the same packed random sampling as our main experiments, and measure evaluation perplexity on validation samples both in-distribution (held-out Alpaca samples) and out-of-distribution (RedPajama validation samples) over different combinations of steps (Figure~\ref{fig:app_ablate_transfer_steps}). 
Without attention transfer, low-rank adaptation converges significantly higher on in-distribution samples (Figure~\ref{fig:app_ablate_transfer_steps_ppl}), suggesting poorer quality linearizing. However, we find similar held-out perplexities after relatively few attention transfer steps (\textit{c.f.}, 1000 - 1800 updates, the former amounting to just 8 million tokens for attention transfer), where all runs improve in-distribution PPL by $\sim$0.23 points after LoRA finetuning for 2000 steps. 

\begin{table}[b]
\vspace{-0.25cm}
\begin{center}
\resizebox{1\linewidth}{!}{
\begin{tabular}{@{}l
>{\columncolor[HTML]{F3F3F3}}c ccccccccc@{}}
\toprule
Attn. Transfer Steps        & 0      & 200                            & 400                           & 600                           & 800                           & 1000                          & 1200                          & 1400                                   & 1600                                   & 1800                          \\ \midrule
Alpaca Eval PPL             & 3.211  & \cellcolor[HTML]{FFFFFF}2.802  & \cellcolor[HTML]{BCD2F6}2.792 & \cellcolor[HTML]{75A3ED}2.782 & \cellcolor[HTML]{77A4ED}2.782 & \cellcolor[HTML]{518AE8}2.776 & \cellcolor[HTML]{5D92EA}2.778 & \cellcolor[HTML]{4A86E8}\textbf{2.775} & \cellcolor[HTML]{77A4ED}2.782          & \cellcolor[HTML]{518BE8}2.776 \\
RedPajama Eval PPL          & 61.305 & \cellcolor[HTML]{FFFFFF}10.459 & \cellcolor[HTML]{95B8F1}9.799 & \cellcolor[HTML]{85ADEF}9.699 & \cellcolor[HTML]{7AA6EE}9.628 & \cellcolor[HTML]{82ABEF}9.679 & \cellcolor[HTML]{588FE9}9.420 & \cellcolor[HTML]{4A86E8}\textbf{9.328} & \cellcolor[HTML]{578FE9}9.413          & \cellcolor[HTML]{4E89E8}9.358 \\
Avg. Zero-shot LM Eval Acc. & 56.86  & \cellcolor[HTML]{7AA6EE}73.68  & \cellcolor[HTML]{FFFFFF}73.34 & \cellcolor[HTML]{72A1ED}73.70 & \cellcolor[HTML]{82ABF0}73.66 & \cellcolor[HTML]{6296EC}73.74 & \cellcolor[HTML]{CCDDF9}73.47 & \cellcolor[HTML]{72A1ED}73.70          & \cellcolor[HTML]{4A86E8}\textbf{73.80} & \cellcolor[HTML]{82ABF0}73.66 \\ \bottomrule
\end{tabular}
}
\end{center}
\vspace{-0.25cm}
\caption{\small \textbf{Effect of attention transfer steps}. With Llama 3 8B linearized on Alpaca data, we report the final evaluation perplexities after 2000 LoRA steps in Fig~\ref{fig:app_ablate_transfer_steps}, as well as downstream LM Eval performance averaged over zero-shot tasks. We again find competitive quality with relatively few attention transfer steps.
}
\label{table:app_ablate_transfer_steps}
\end{table}


In Table~\ref{table:app_ablate_transfer_steps}, we report the numerical values for held-out perplexities at the end of linearizing (1800 attention transfer steps + 2000 low-rank adaptation steps), as well as the average LM Eval score over zero-shot tasks. We similarly find competitive generalized zero-shot quality with relatively few attention transfer steps (200 steps), all achieving 7.70--8.16 higher points than the next best Mamba-Llama model (0.16--0.62 higher points than the 50\% softmax attention variant, \textit{c.f.}, Table~\ref{table:main_results_linearize}). Without any attention transfer, linearized LLMs perform drastically worse on out-of-distribution samples (Table~\ref{table:app_ablate_transfer_steps}, RedPajama and LM Eval metrics).

\begin{figure}[t]
    \centering
    \begin{subfigure}[b]{0.49\textwidth}
        \centering
        \includegraphics[width=1\textwidth]{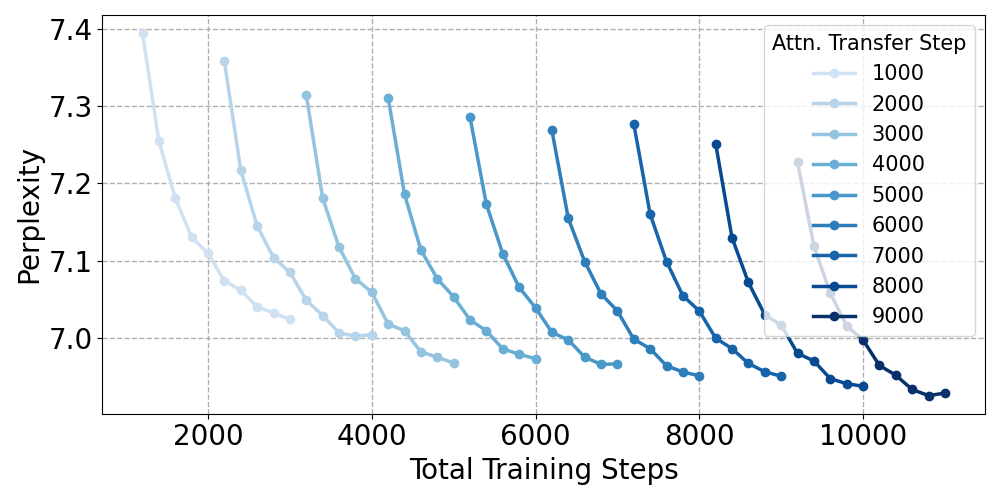}
        \caption{RedPajama Validation Perplexity}
        \label{fig:app_ablate_transfer_rp_ppl}
    \end{subfigure}
    \begin{subfigure}[b]{0.49\textwidth}
        \centering
        \includegraphics[width=1\textwidth]{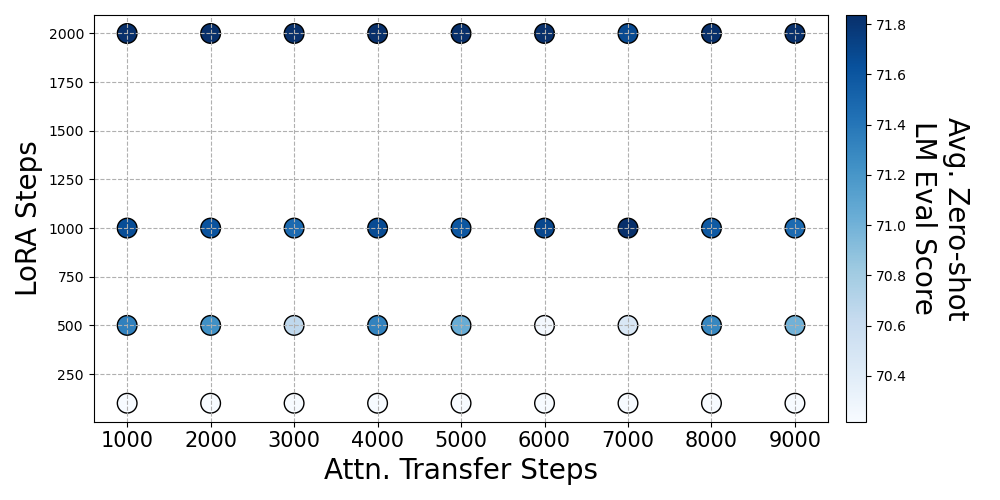}
        \caption{LM Eval Score}
        \label{fig:app_ablate_transfer_rp_lm_eval}
    \end{subfigure}
    \vspace{-0.25cm}
    \caption{\small \textbf{Evaluation quality from ablating RedPajama linearizing updates}. }
    \label{fig:app_ablate_transfer_steps}
\end{figure}

\textbf{Impact of more pretraining data.} We next study how linearizing over larger amounts of pretraining data impacts quality. We randomly sample a larger set of unique RedPajama training sequences (1024-token packed; 72,000 such samples overall), allowing us to linearize Llama 3 8B with different combinations of up to 9,000 attention transfer updates and 2,000 low-rank linearizing updates. To test language modeling recovery, we report both held-out validation sample perplexity (Table~\ref{fig:app_ablate_transfer_rp_ppl}) and general zero-shot LM Evaluation Harness quality (Table~\ref{fig:app_ablate_transfer_rp_lm_eval}). Increasing both Stage 1 attention transfer steps and Stage 2 low-rank adjusting steps notably improves validation perplexity. However, we similarly find competitive zero-shot LM scores across all evaluated attention transfer steps. Across checkpoints at different numbers of low-rank updates, attention transfer with up to 9$\times$ more unique tokens does not seem to monotonically improve downstream quality (Table~\ref{fig:app_ablate_transfer_rp_lm_eval}). Meanwhile, we do find that across various amounts of attention transfer steps, subsequent low-rank adaptation consistently improves average zero-shot LM score by $>$1 points.

\subsection{Layer-wise Training Dynamics and Analysis}
\label{app:layerwise_study}

Finally, we further study \name{} layer-wise training dynamics, such as the resulting linear attention approximation quality to softmax attention at every layer (App.~\ref{app:layerwise_mse_study}), how this corresponds with LoRA updates after attention transfer (App.~\ref{app:layerwise_lora_train}), and how these layer-wise attention MSEs (a) relate to model size and (b) further motivate block-wise training (App.~\ref{app:blockwise_study}).



\subsubsection{Layer-wise Softmax Attention Recovery with LoRA Adjusting}
\label{app:layerwise_mse_study}

We now study how LoRA can explicitly improve softmax attention approximation. Given learned feature maps, can we recover softmax attention better by adjusting the attention projections?

\begin{figure}[t]
\vspace{-0.25cm}
    \centering
    \begin{subfigure}[b]{0.49\textwidth}
        \centering
        \includegraphics[width=1\textwidth]{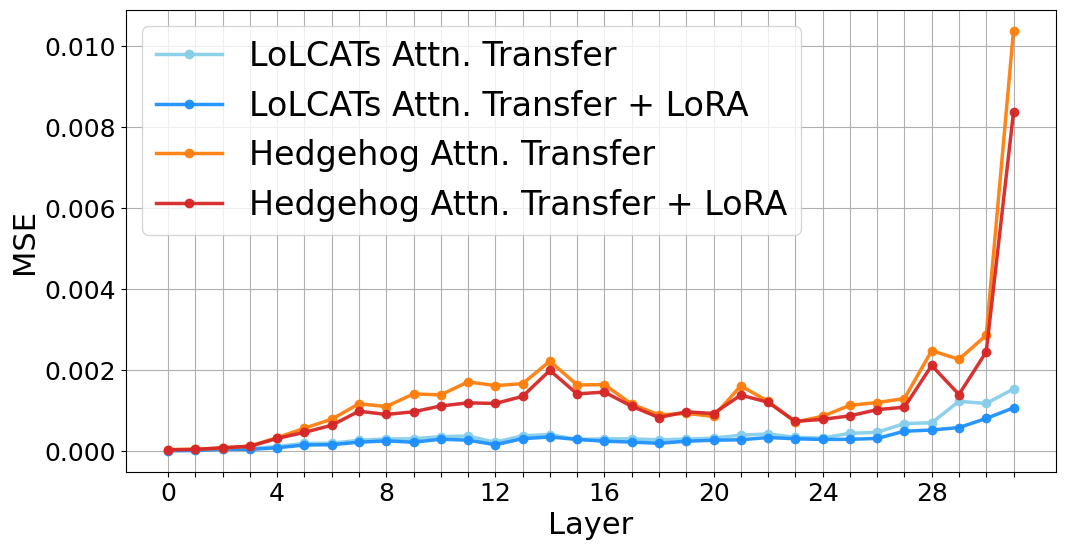}
        \vspace{-0.5cm}
        \caption{Layer-wise MSE before and after LoRA adjusting}
        \label{fig:app_layerwise_mse_lora_number}
    \vspace{-0.25cm}
    \end{subfigure}
    \begin{subfigure}[b]{0.49\textwidth}
        \centering
        \includegraphics[width=1\textwidth]{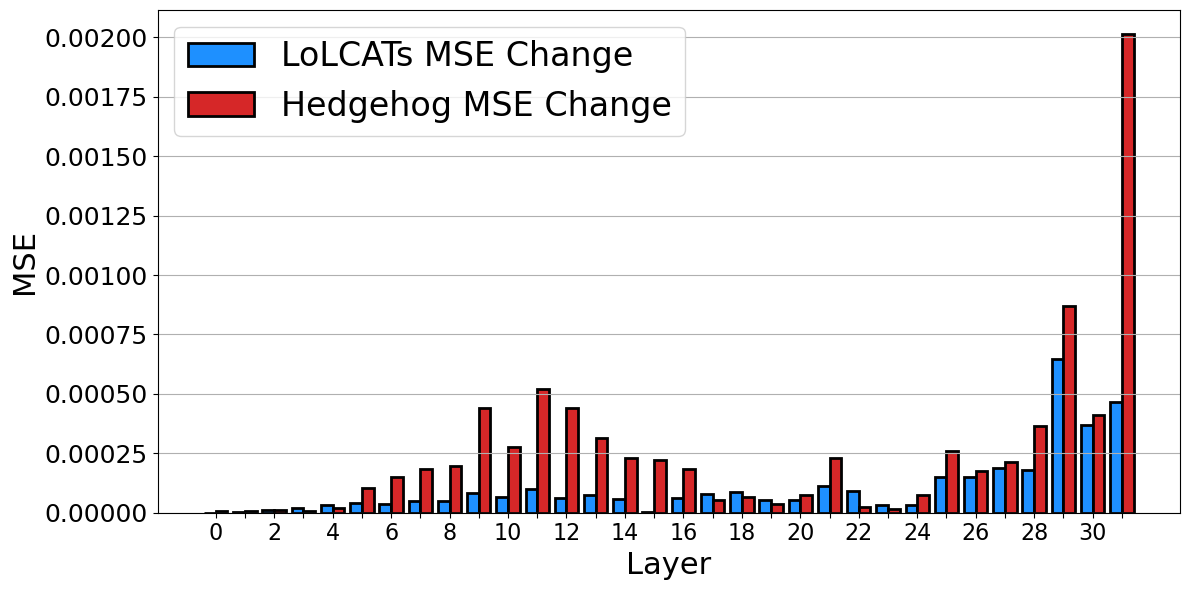}
        \vspace{-0.5cm}
        \caption{Layer-wise MSE change after LoRA adjusting}
        \label{fig:app_layerwise_mse_lora_change}
    \vspace{-0.25cm}
    \end{subfigure}
    \caption{\small \textbf{Layer-wise MSE and deltas}. Layer-wise MSE and change in MSE (absolute) before and after LoRA adjusting with the Hedgehog linear attention and \name{} linear + sliding window attention on Llama 3 8B. Both use the Hedgehog feature map. With larger initial MSEs after attention transfer (Fig.~\ref{fig:app_layerwise_mse_lora_number}), LoRA adjusting with Hedgehog results in larger MSE improvements in MSE across layers (Fig.~\ref{fig:app_layerwise_mse_lora_change}). }
    \label{fig:app_layerwise_mse_lora}
    \vspace{-0.25cm}
\end{figure}

In Figure~\ref{fig:app_layerwise_mse_lora_number}, we report the layer-wise MSE between learned linear attentions and softmax attention using either the pure linear attention (Hedgehog) or linear + sliding window attention (\name{}). We plot the mean MSE computed over all samples in our Alpaca validation set, averaging over all heads, token positions, and heads per layer. In addition to plotting this metric after attention transfer (as in Figure~\ref{fig:mse_analogs}), we also plot the MSE after an additional round of LoRA adjusting (+ LoRA). Like before, we freeze the linear attention weights and  add LoRA weights to query, key, value, and output projections. However, rather than train these weights end-to-end for next-token prediction, we update LoRAs to explicitly minimize the MSE between our trainable linear attention and original softmax attention outputs like in Stage 1. We use the same hyperparameters as in Table~\ref{tab:8b-training-details}.

We report both the absolute MSEs (Figure~\ref{fig:app_layerwise_mse_lora_number}) and the change in MSE (Figure~\ref{fig:app_layerwise_mse_lora_change}). LoRA reduces MSEs with both linear attentions--- frequently reducing MSE more when starting with worse approximations after attention transfer---which may suggest some type of compensatory role. 
This occurs both across layers for each linear attention, where we see greater MSE deltas for the later layers (\textit{c.f.}, \name{} MSE change, Figure~\ref{fig:app_layerwise_mse_lora_change}), and between linear attentions (\textit{c.f.}, Hedgehog MSE change, Figure~\ref{fig:app_layerwise_mse_lora_change}), where LoRA generally improves MSE more with the Hedgehog linear attention versus \name{} linear + sliding window. Despite these greater improvements, LoRA alone does not close the MSE gap (Figure~\ref{fig:app_layerwise_mse_lora_number}), suggesting that linearizing architecture still plays an important role in learning to match softmax attention.

\subsubsection{Layer-specific LoRA Training Dynamics}
\label{app:layerwise_lora_train}

Next, we further study how \name{} layers behave during LoRA adjusting, and plot the cumulative weight updates to LoRA low-rank $\bm{A}$ and $\bm{B}$ weight matrices while training LLMs end-to-end for next-token prediction. As reference points, we compare against LoRA finetuning (1) the original Transformer LLMs with softmax attention, (2) linearized LLMs without attention transfer (using the \name{} linear + sliding window attention, Hedgehog feature map, but not trained to match softmax first), (3) linearized LLMs with attention transfer and the Hedgehog pure linear attention, and (4) linearized LLMs with Hedgehog pure linear attention without attention transfer. We finetune Llama 3 8B with two epochs on the Alpaca dataset, following Step 2 hyperparameters in Table~\ref{tab:8b-training-details}. 

\begin{figure}[!h]
\vspace{-0.25cm}
    \centering
    \begin{subfigure}[b]{1\textwidth}
        \centering
        \includegraphics[width=1\textwidth]{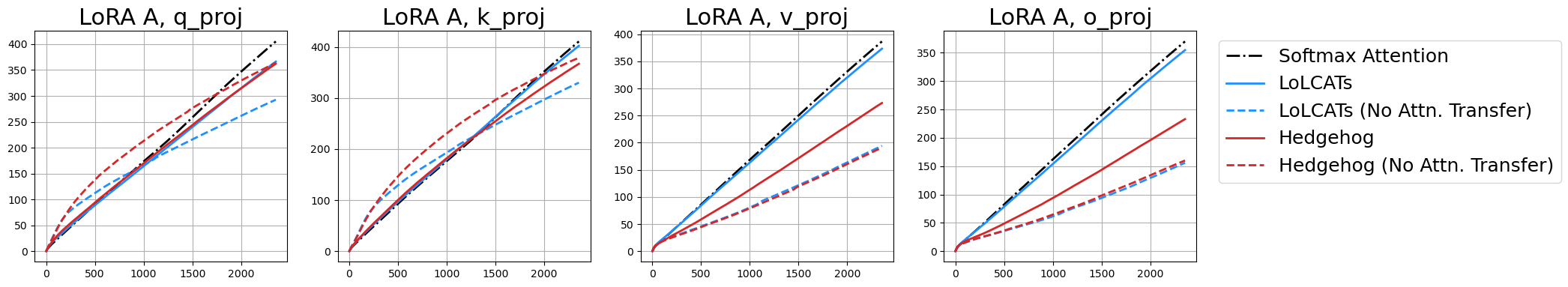}
        \label{fig:app_lora_travel_avg_a}
    \vspace{-0.25cm}
    \end{subfigure}
    \begin{subfigure}[b]{1\textwidth}
        \centering
        \includegraphics[width=1\textwidth]{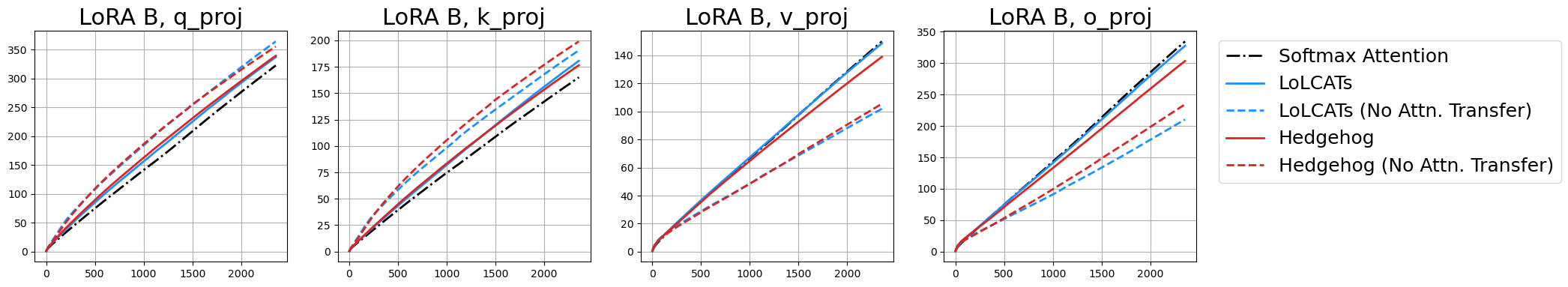}
        \label{fig:app_lora_travel_avg_b}
    \vspace{-0.25cm}
    \end{subfigure}
    \vspace{-0.5cm}
    \caption{\small \textbf{LoRA $\bm{A},\bm{B}$ weight updates by attention projection over training}, averaged over all layers.}
    \vspace{-0.25cm}
\label{fig:app_lora_travel_avg}
\end{figure}

\begin{figure}[!h]
    \centering
    \begin{subfigure}[b]{1\textwidth}
        \centering
        \includegraphics[width=1\textwidth]{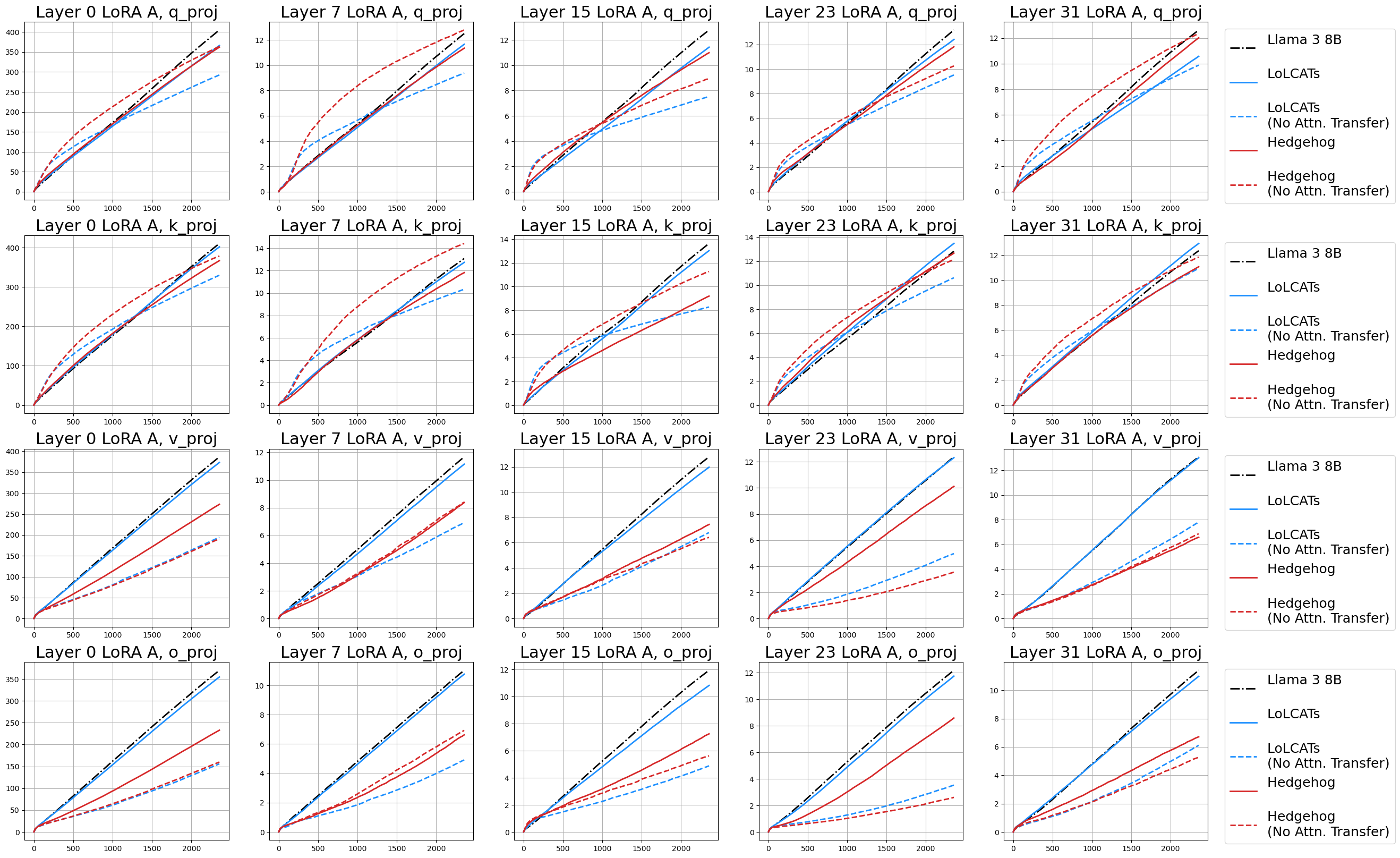}
        \label{fig:app_lora_travel_layers_a}
    \vspace{-0.25cm}
    \end{subfigure}
    \vspace{-0.5cm}
    \caption{\small \textbf{LoRA $\bm{A}$ weight updates by attention projection and layer over training}.}
    \vspace{-0.25cm}
\label{fig:app_lora_travel_layer}
\end{figure}

\begin{figure}[!h]
    \centering
    \begin{subfigure}[b]{1\textwidth}
        \centering
        \includegraphics[width=1\textwidth]{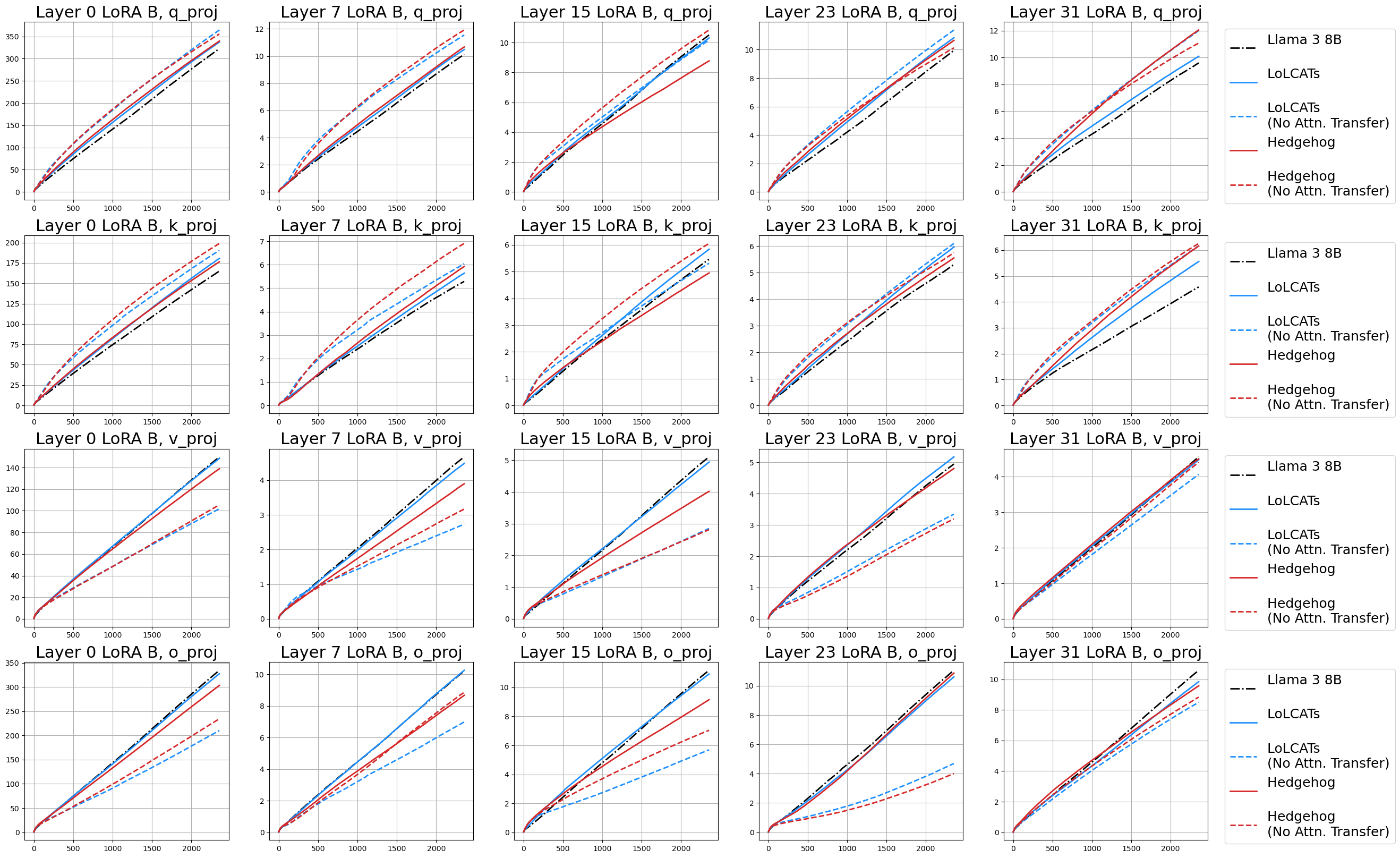}
        \label{fig:app_lora_travel_layers_b}
    \vspace{-0.25cm}
    \end{subfigure}
    \vspace{-0.5cm}
    \caption{\small \textbf{LoRA $\bm{B}$ weight updates by attention projection and layer over training}.}
    \vspace{-0.25cm}
\label{fig:app_lora_travel_layer_b}
\end{figure}

We plot these updates per LoRA weight and projection in Figure~\ref{fig:app_lora_travel_avg}, with layer-specific plots in Figure~\ref{fig:app_lora_travel_layer} and~\ref{fig:app_lora_travel_layer_b}. We specifically show cumulative sums of the update magnitudes over 2500 steps. 

Across plots, we find several interesting findings. First, \name{} with attention transfer results in much more similar cumulative updates to softmax attention for value and output projection LoRAs, where both result in noticeably larger updates than any other configuration. This may suggest the \name{} attention transfer linear attentions are sufficiently similar to softmax attention, such that when LoRA finetuning end-to-end, the weights also behave similarly. Furthermore, for query and key projections---\ie{} those involved in the attention weight computation---the situation is reversed. The untrained or no attention transfer linearized LLMs display \textit{greater} cumulative updates earlier during training for $\bm{A}$ weights and consistently for $\bm{B}$ weights than either trained linear attention or softmax attention. As we train all layers jointly, we find that with the untrained linear attentions, relatively more ``weight'' is diverted to learning to compute attention weights via query and key projection updates. How this impacts the downstream model---\ie{} by updating the value and output projections less than a softmax attention Transformer---is an interesting question for future work.

When plotting these updates per layer (Figure~\ref{fig:app_lora_travel_layer},~\ref{fig:app_lora_travel_layer_b}) we find similar dynamics. Interestingly, despite \name{} with attention transfer resulting in the worst softmax attention approximations for the layer 31 (\textit{c.f.}, attention MSE, Figure~\ref{fig:mse_analogs},~\ref{fig:app_layerwise_mse_lora_number}), the resulting LoRA weight deltas do not seem to track the softmax attention LoRA deltas noticeably worse than other layers. 


\subsubsection{Block-wise Attention Transfer}
\label{app:blockwise_study}

Finally, we study how attention transfer properties change with model scale, motivating \name{}'s block-wise training approach (Section~\ref{sec:method_training}).
In particular, we note that attention output MSEs can vary quite a bit across layers, and this can be aggravated by model size (\textit{c.f.,} Llama 3.1 70B, Table~\ref{table:mse_blocks_70b}; and Llama 3.1 405B, Table~ \ref{table:mse_blocks_405b}).

Recall that attention transfer involves training the \name{} linear attentions to match the outputs of softmax attention at each layer by minimizing the MSE between the softmax and linear attention outputs. 
 Since the MSE loss is scale-sensitive and MSE already varies across layers after attention transfer for Llama 3 8B (Figure~\ref{fig:mse_error_llama3_layer}), we hypothesize that jointly training all $126$ Transformer layers in Llama 3.1 405B -- by summing the MSE losses across all layers -- may be difficult. Correspondingly, in Table~\ref{table:block_wise_ppl}, we find that block-wise attention transfer leads to lower language modeling perplexity for 405B linearized LLMs compared to joint training. However, we find that joint training is sufficient and performs similarly to block-wise training at the smaller scales (8B, 70B). 


\paragraph{Tradeoffs between quality and efficiency.} We show that the block-wise approach improves linearization quality at large model scales and improves the user's ability to flexibly balance compute and memory efficiency tradeoffs. We compare (1) \textbf{joint ($k=126$)} training, where we load the full model once and compute the loss as the sum of layer-wise MSEs, and (2) \textbf{block-wise ($k=9$)} training, where we break the model into blocks and train each independently. Per our discussion in Sec.~\ref{sec:method_training}, this increases storage costs to save precomputed hidden states. However, we hypothesize that optimizing a block-wise attention transfer loss can improve quality.

\begin{wraptable}{r}{0.3\textwidth}  
\vspace{-0.25cm}
\centering
\resizebox{1\linewidth}{!}{
\setlength{\tabcolsep}{4pt}
\begin{tabular}{@{}lcl@{}}
\toprule
Method              &  $b \times M // b$  &  PPL\\ \midrule
Joint          &  126$\times$1  & 4.23 \\
Block-wise      &  9$\times$14  & 3.21
\\ \bottomrule
\end{tabular}
}
\vspace{-0.25cm}
\caption{\small \textbf{Validation perplexity after LoRA adjusting for Llama 3.1 405B}, comparing attention transfer with $M // b$ blocks of $b$ layers. $M = 126$ is number of total layers.}
\label{table:block_wise_ppl}
\vspace{-1cm}
\end{wraptable}

For quality, we compare the evaluation perplexity of both the attention transfer approaches, after LoRA adjusting on the same subset of RedPajama data. The block-wise and joint approaches perform similarly at the 8B and 70B scales, however, as shown in Table~\ref{table:block_wise_ppl}, the block-wise approach performs 1.02 points better than joint attention transfer at the 405B scale. These results support our study in Section~\ref{sec:method_study}, which shows the  variation in MSE magnitudes grows large at the 405B scale. 

\paragraph{Layer-wise MSE.} We report the attention MSEs across model scales, finding that variation in MSE magnitudes corresponds with model size. In Tables~\ref{table:mse_blocks_70b} and \ref{table:mse_blocks_405b}, we report magnitude of the MSE loss as the depth of the layers in the block increases. MSE variation increases with model size: while the largest MSEs at the 70B scale are between 2.5$-$3$\times$ the MSE of the final block at the 405B scale  is 48.66 (over $13\times$ that of the next largest MSE by block, 3.62). 

\begin{table}[h]  
\centering
\setlength{\tabcolsep}{4pt}
\begin{tabular}{@{}ccl@{}}
\toprule
Block  (Layer range)   &  Eval MSE  \\ \midrule
0-4   & $8e-4$ \\
5-9   & 0.03 \\
10-14 & 0.06 \\
15-19 & 0.10\\
20-24 & 0.28 \\
25-29 & 0.73 \\
30-34 & 0.28 \\
35-39 & 0.28 \\
40-44 & 0.25 \\
45-49 & 1.08 \\
50-54 & 0.26 \\
55-59 & 0.18 \\
60-64 & 0.45 \\
65-69 & 0.50 \\
70-74 & 2.91 \\
75-79 & 2.56 \\
\bottomrule
\end{tabular}
\vspace{-0.125cm}
\caption{\small \textbf{Attention transfer block-wise MSE} We report the eval MSE by 5-layer block for each of the 16 blocks in the 80 Transformer layer Llama 3.1 70B model. Each block is trained on the exact same set of RedPajama data at sequence length 1024.}
\label{table:mse_blocks_70b}
\vspace{-0.25cm}
\end{table}

\begin{table}  
\vspace{-0.5cm}
\centering
\setlength{\tabcolsep}{4pt}
\begin{tabular}{@{}ccl@{}}
\toprule
Block  (Layer range)   &  Eval MSE  \\ \midrule
0-8   & 0.03 \\
9-17  & 0.02  \\
18-26 & 0.09 \\
27-35 & 0.04 \\
36-44 & 0.36 \\
45-53 & 0.42 \\
54-62 & 1.59 \\
63-71 & 2.75 \\
72-80 & 3.62 \\
81-89 & 2.49 \\
90-98 & 0.29 \\
99-107  & 1.11 \\
108-116 & 3.42 \\
117-126 & 48.66 \\
\bottomrule
\end{tabular}
\vspace{-0.125cm}
\caption{\small \textbf{Attention transfer block-wise MSE} We report the eval MSE by 9-layers block for the 14 blocks in the 126 Transformer layer Llama 3.1 405B model. Each block is trained on the exact same set of RedPajama data at sequence length 1024.}
\label{table:mse_blocks_405b}
\vspace{-0.25cm}
\end{table}

\clearpage
\section{Implementation Details}
\label{appendix:implementation_details}
\subsection{Pseudocode Implementation}
\label{appendix:code}

Below we provide further details on implementing \name{} with PyTorch-like code and example demonstrations from the HuggingFace Transformers library.

\textbf{Learnable Linear Attention.}
To start, we simply replace the softmax attentions in an LLM with a linear attention. We define such a class below.

\begin{lstlisting}[language=Python, caption=\name{} Linear Attention Class]
import copy
import torch.nn as nn
from einops import rearrange

class LolcatsLlamaAttention(nn.Module):
    def __init__(self,
                 feature_dim: int,
                 base_attn: nn.Module,  # original Transformer attn.
                 ) -> None:
        super().__init__()
        
        # Inherit pretrained weights
        self.q_proj = base_attn.q_proj
        self.k_proj = base_attn.k_proj
        self.v_proj = base_attn.v_proj
        self.o_proj = base_attn.o_proj

        # Inherit other attention things
        self.rotary_emb = base_attn.rotary_emb
        self.base_attn  = base_attn  # keep for attention transfer
        self.num_heads  = base_attn.num_heads
        self.head_dim   = base_attn.head_dim

        # Initialize feature maps, see Hedgehog definition below
        self.feature_map_q = HedgehogFeatureMap(
            self.num_heads, self.head_dim, feature_dim
        )
        self.feature_map_k = copy.deepcopy(self.feature_map_q)

    def forward(self, x: torch.Tensor) -> torch.Tensor:
        """
        Compute linear attention (assume no GQA)
        (b: batch_size, h: num_heads, l: seq_len, d: head_dim)
        """
        q = self.q_proj(x)  # assume all are (b, h, l, d)
        k = self.k_proj(x)
        v = self.v_proj(x)
        
        # Apply rotary embeddings
        q = self.rotary_emb(q)
        k = self.rotary_emb(k)
        
        # Apply feature maps
        q = self.feature_map_q(q)  # (b, h, l, feature_dim)
        k = self.feature_map_k(k)  # (b, h, l, feature_dim)
        
        # Compute linear attention
        kv = torch.einsum('bhlf,bhld->bhfd', k, v)
        y  = torch.einsum('bhlf,bhfd->bhld', q, kv)
        y /= torch.einsum('bhlf,bhlf->bhl', 
                          q, k.cumsum(dim=2))[..., None]
        # Apply output projection
        return self.o_proj(rearrange(y, 'b h l d -> b l (h d)'))
\end{lstlisting}

\newpage
\textbf{Linear + Sliding Window Attention.}
We can augment this linear attention with the linear attention and sliding window formulation described in Eq.~\ref{eq:hybrid_linear_swa}. We first define standalone functions for linear attention and sliding window softmax, before defining such as hybrid class below.

\begin{lstlisting}[language=Python, caption=Linear Attention in Linear + Sliding Window Attention]
def sliding_window_softmax_attention(q: torch.Tensor, 
                                     k: torch.Tensor, 
                                     v: torch.Tensor,
                                     window_size: int, 
                                     window_factor: float):
    """
    Compute sliding window softmax attention in O(n) time and space
    by not materializing O(n^2) attention weights
    """        
    d = q.shape[-1]
    # Compute windows for keys and values, shifting by window size 
    window_kwargs = {'dimension': 2, 'size': window_size, 'step': 1}
    k = F.pad(k, (0, 0, window_size - 1, 0), value=0).unfold(**window_kwargs)
    v = F.pad(v, (0, 0, window_size - 1, 0), value=0).unfold(**window_kwargs)

    # Compute windowed_softmax(qk); causal in its construction
    a_sm = torch.einsum('bhld,bhldw->bhlw', q, k) * (d ** -0.5)
    # heuristic for zeroing out padding above  
    a_sm[a_sm == 0] = -torch.finfo(q.dtype).max  

    # Compute softmax terms for combining attentions (attn and sum)
    a_sm_max = torch.amax(a_sm, dim=-1, keepdim=True)
    a_sm     = window_factor * torch.exp(a_sm - a_sm_max)
    sum_sm   = a_sm.sum(dim=-1, keepdim=True)
    return torch.einsum('bhlw,bhldw->bhld', a_sm, v), sum_sm
\end{lstlisting}

\begin{lstlisting}[language=Python, caption=Linear Attention in Linear + Sliding Window Attention]
def under_window_linear_attention(f_q: torch.Tensor,  # phi(q)  
                                  f_k: torch.Tensor,  # phi(k)  
                                  v: torch.Tensor,
                                  window_size: int, 
                                  linear_factor: float):
    """
    Compute hybrid window attention dot product with 
    linear complexity in q_len
    """
    dtype = f_q.dtype

    # Shift keys and valuves for window
    w   = window_size
    f_k = F.pad(f_k, (0, 0, w, 0), value=0)[:, :, :-w, :]
    v   = F.pad(v, (0, 0, w, 0), value=0)[:, :, :-w, :]

    # Compute linear terms for combining attentions
    kv  = torch.einsum('bhlf,bhld->bhfd', k, v)
    qkv = linear_factor * torch.einsum('bhlf,bhfd->bhld', q, kv)
    sum_f_k = f_k.float().cumsum(dim=2).to(dtype=dtype)
    sum_qk  = linear_factor * torch.einsum("bhld,bhld->bhl", 
                                           f_q, sum_f_k)[..., None]
    return qkv, sum_qk
\end{lstlisting}

\newpage

\begin{lstlisting}[language=Python, caption=\name{} Linear + Sliding Window Attention Class]
class LolcatsSlidingWindowLlamaAttention(LolcatsLlamaAttention):
    def __init__(self, window_size: int = 64, **kwargs: any,):
        super().__init__(**kwargs)
        self.window_size = window_size  # sliding window size
        self.window_factors = nn.Parameter(  # gamma mixing term
            torch.ones(1, self.num_heads, 1, 1)
        )

    def attention(self,
                  q: torch.Tensor, k: torch.Tensor,  
                  f_q: torch.Tensor, f_k: torch.Tensor, 
                  v: torch.Tensor,
                  window_factor: torch.Tensor,
                  linear_factor: torch.Tensor
                  window_size: int = 64,):
        """
        O(n) hybrid linear + sliding window attention
        """
        window_kwargs = {'dimension': 2, 'size': window_size, 'step': 1}
        # 1. Sliding window (softmax attention)
        with torch.no_grad():
            qkv_sm, sum_qk_sm = sliding_window_softmax_attention(
                q, k, v, window_size, window_factor)
    
        # 2. Under window (linear attention)
        qkv_ln, sum_qk_ln = under_window_linear_attention(
            f_q, f_k, v, window_size, linear_factor)
    
        # 3. Combine
        y = (qkv_sm + qkv_ln) / (sum_qk_sm + sum_qk_ln)
        return y

    def forward(self, x: torch.Tensor) -> torch.Tensor:
        """
        Compute linear attention (assume no GQA)
        (b: batch_size, h: num_heads, l: seq_len, d: head_dim)
        """
        q = self.q_proj(x)  # assume all are (b, h, l, d)
        k = self.k_proj(x)
        v = self.v_proj(x)
        
        # Apply rotary embeddings
        q = self.rotary_emb(q)
        k = self.rotary_emb(k)
        
        # Apply feature maps
        f_q = self.feature_map_q(q)  # (b, h, l, feature_dim)
        f_k = self.feature_map_k(k)  # (b, h, l, feature_dim)

        # Compute attention
        window_factors = F.sigmoid(self.window_factors)
        linear_factors = 1  # Eq. 7
        y = self.attention(q, k, f_q, f_k, v,
                           window_factors, linear_factors,
                           self.window_size)
        # Apply output projection
        return self.o_proj(rearrange(y, 'b h l d -> b l (h d)'))
\end{lstlisting}

\newpage
\textbf{Hedgehog Feature Map.} We implement the Hedgehog feature map following \citet{zhang2024the}. 

\begin{lstlisting}[language=Python, caption=Hedgehog Feature Map]
import torch.nn as nn

class HedgehogFeatureMap(nn.Module):
    def __init__(self,
                 num_heads = 32: int,  # defaults for 8B LLMs
                 head_dim = 128: int,
                 feature_dim = 64: int,
                 ) -> None:
        super().__init__()
        self.num_heads = num_heads
        self.head_dim = head_dim
        self.feature_dim = feature_dim

        # Initialize trainable feature map weights
        self.weights = nn.Parameter(
            torch.zeros(self.num_heads, self.head_dim, self.feature_dim)
        )

    def self.activation(self: torch.Tensor) -> torch.Tensor:
        """Softmax across feature dims activation"""
        return torch.cat([
            torch.softmax(x, dim=-1), torch.softmax(-x, dim=-1)
        ], dim=-1)

    def forward(self, x: torch.Tensor) -> torch.Tensor:
        """
        Assume x.shape is (b, h, l, d)
        (b: batch_size, h: num_heads, l: seq_len, d: head_dim)
        """
        x = torch.einsum('hdf,bhld->bhlf', self.weights, x)
        return self.activation(x)
\end{lstlisting}

\textbf{Linearizing LLM Setup.} To initialize an LLM for linearizing, we simply replace each softmax attention in the Transformer's layers with our LoLCATs linear attention class. We illustrate this with a Huggingface Transformer's class below.

\begin{lstlisting}[language=Python, caption=Linearizing LLM Setup]
from transformers import AutoModelForCausalLM

def convert_model(model: AutoModelForCausalLM,
                  window_size: int = 64,
                  feature_dim: int = 64,):
    """Setup linearizing attentions"""
    for layer in model.model.layers:
        if window_size == 0:
            layer.self_attn = LolcatsLlamaAttention(
                feature_dim=feature_dim,
                base_attn=layer.self_attn,
            )
        else:
            layer.self_attn = LolcatsSlidingWindowLlamaAttention(
                window_size=window_size,
                feature_dim=feature_dim,
                base_attn=layer.self_attn,
            )
    return model
\end{lstlisting}

\newpage
\textbf{Attention Transfer Training.} We can then train \name{} layers in a simple end-to-end loop. Although doing this attention transfer is akin to a layer-by-layer cross-architecture distillation, due to architectural similarities we implement linearizing with the same footprint as finetuning a single model. Furthermore, as we freeze all parameters except for the newly introduced feature map weights, this amounts to \textit{parameter-efficient} finetuning, training $<$0.2\% of a 7B+ LLM's parameters. 
\newline

\begin{lstlisting}[language=Python, caption=End-to-end attention transfer pseudocode (Stage 1).]
"""
Example attention transfer training loop for Llama 3.1 8B
"""
import torch.nn as nn
from transformers import AutoModelForCausalLM

# Load Llama 3.1 8B
model_config = {
    'pretrained_model_name_or_path': 'meta-llama/Meta-Llama-3.1-8B'
}
model = AutoModelForCausalLM.from_pretrained(**model_config)

# Freeze all pretrained weights
for p in model.parameters():
    p.requires_grad = False

# Prepare LoLCATs linearizing layers
model = convert_model(model)  

# Setup MSE loss criterion
mse_loss = nn.MSELoss()
block_size = 32  # default end-to-end for 7B+ LLMs
num_blocks = len(model.layers) // block_size

# Get some linearizing data
train_loader = load_data(**data_kwargs)

# Train LoLCATs layers via attention transfer
for ix, input_ids in enumerate(train_loader):
    losses = [0] * range(num_blocks)   # Attention transfer loss here
    x = model.embed_tokens(input_ids)  # Input embeddings from tokens

    # Forward pass thru model  
    for lix, layer in enumerate(model.layers):         
        # *** Start Attention ***
        _x = layer.input_layernorm(x)  # Just Llama things

        ## Attention Transfer part  
        with torch.no_grad():
            y_true = layer.self_attn.base_attn(_x)
        y_pred = layer.self_attn(_x)
        _idx = lix // block_size      # Add layer or block-wise MSE
        losses[_idx] += mse_loss(y_pred, y_true) 
        _x = y_true                   # Pass true attention outputs 
                                      # thru to rest of model
        x = _x + x
        # *** End Attention ***

        # *** Start MLP ***
        _x = layer.post_attention_layernorm(x)
        _x = self.mlp(_x)
        x = _x + x
        # *** End MLP ***

    for loss in losses:  # End-to-end attention transfer
        loss.backward() 
\end{lstlisting}

\textbf{Low-rank Adjusting.} Finally, after attention transfer, we train the model end-to-end with next-token prediction. This allows the linearized LLM to adjust to the learned linear attentions, which may still not perfect approximations of the softmax attentions. However, with \name{} we hope to make these errors small enough such that we can adjust and recover pretrained LLM capabilities with parameter-efficient low-rank updates (\eg{} LoRA finetuning).
\newline
\begin{lstlisting}[language=Python, caption=Defining a LoRA layer]
class LoRALayer(torch.nn.Module):
    def __init__(self, 
                 base_layer: nn.Module,
                 rank: int = 8, 
                 alpha: float = 16):
        super().__init__()
        """Init low-rank parameters"""
        in_dim  = base_layer.weight.shape[1]  
        out_dim = base_layer.weight.shape[0]  
        self.A  = nn.Parameter(torch.randn(in_dim, rank))
        self.B  = nn.Parameter(torch.zeros(rank, out_dim))
        self.alpha = alpha
        self.base_layer = base_layer

    def low_rank_forward(self, x: torch.Tensor) -> torch.Tensor:
        """Compute LoRA pass"""
        x = torch.einsum('...d,dr->...r', x, self.A)
        x = torch.einsum('...r,rd->...d', x, self.B)
        return self.alpha * x

    def forward(self, x: torch.Tensor) -> torch.Tensor:
        """Actual forward"""
        x = self.base_layer(x) + self.low_rank_forward(x)
        return x
\end{lstlisting}

\begin{lstlisting}[language=Python, caption=End-to-end low-rank adjusting pseudocode (Stage 2).]
model = attention_transfer(model)  # Do Step 1 of LoLCATs  

# 0. Prepare model for LoRA
lora_kwargs = {'rank': 8, 'alpha': 16}  # examples  

# 1. Freeze all pretrained weights
for p in model.parameters(): p.requires_grad = False

# 2. Add LoRA weights to Q,K,V,O projections
for layer in model.layers:
    for proj in ['q_proj', 'k_proj', 'v_proj', 'o_proj']:
        _layer = getattr(layer.self_attn, proj)
        _layer = LoRALayer(_layer, **lora_kwargs)
        setattr(layer.self_attn, proj, lora_layer)
    
# 3. Finetune LLM with LoRA
xent_loss = nn.CrossEntropyLoss()  
train_loader = load_data(**data_kwargs)

for ix, input_ids in enumerate(train_loader):
    # Process input tokens  
    next_token_ids = model(input_ids)
    # Train model to predict next token
    y_pred = next_token_ids[..., :1]
    y_true = input_ids[..., 1:]
    loss = xent_loss(y_pred, y_true)
    loss.backward()
\end{lstlisting}

\newpage
\subsection{Hardware-aware implementation of \name{} sliding window}
\label{app:tk_implementation}

Despite the theoretical efficiency of linear attention, existing  implementations have long underperformed well-optimized attention implementations (e.g., FlashAttention) in wall clock speed \citep{dao2022flashattention}. To translate the benefits of \name{} to wall clock speedups, we develop a custom hardware-aware algorithm for  \name{} \textbf{prefill} using the ThunderKittens CUDA framework.\footnote{\url{https://github.com/HazyResearch/ThunderKittens}}
We first briefly review the GPU execution model and then detail our algorithm. 

\subsubsection{GPU execution model}
GPUs workloads are executed by independent streaming multiprocessors (SMs), which contain warps, groups of $32$ threads, that operate in parallel. 

\paragraph{Memory hierarchy.} ML workloads involve moving large tensors (weights, activations) in and out of memory to perform computation. GPUs have a memory hierarchy, which includes global memory (HBM), shared memory (SRAM), and registers. Reading from and writing  data to memory, referred to as I/O operations, takes time.  There is a large amount of HBM, which has high I/O costs, and a small amount of SRAM and registers have much costs. All SMs access global memory, warps within an SM threadblock can access shared  memory, and threads within a threadblock have independent register memory. To reduce the I/O costs, \textbf{locality} is key -- kernels should perform as many operations as possible on data that has already been loaded into fast memory (\textit{i.e.}, thread registers) before writing the results back to slower memory.

\paragraph{Compute units.} GPUs have increasingly heterogeneous compute units on newer generations of hardware. Tensor cores---specialized compute units for matrix-matrix multiplications---are the fastest units, operating at 1.0 PetaFLOPS on Nvidia H100 GPUs in contrast to 67 TeraFLOPS for the general non Tensor core units. ML workloads should thus ideally \textbf{exploit the tensor cores}.  

\paragraph{Cost model.} Overall, workloads may either be compute or memory bound, depending on whether they are bottlenecked by the compute speed or I/O costs. To \textit{hide} latencies from either expensive compute or I/O, a classic principle in systems is to \textbf{pipeline computation} among parallel workers. 

\subsubsection{ThunderKittens CUDA kernel for prefill}
We describe our overall approach below and provide  pseudocode in Algorithm \ref{alg:blocky_L_R}, designed around the three principles above: memory locality, tensor core utilization, and pipelined execution. 

The kernel fuses the entire \name{} layer, taking as input the attention queries, keys, and values, for $\bm{q}, \bm{k}, \bm{v} \in \mathbb{R}^{N \times d}$ with sequence length $N$ and head dimension $d$ and outputting the result of the $\bm{y} \in \mathbb{R}^{N \times d}$. Following Llama 3 \citep{llama3modelcard}, we let $d=128$ in the discussion below. 

\paragraph{Pipeline execution overview.} Each thread block handles a single batch and head element of size $N \times d$. The kernel loops over chunks of length $64$ along the sequence dimension $N$, loading $64 \times 128$ tiles of $\bm{q}, \bm{k}, \bm{v}$, which we'll refer to as $\bm{q}_t, \bm{k}_t, \bm{v}_t$, in each iteration $t$.  
We use $8$ warps (workers) per thread block, splitting them into two groups of $4$ workers that pipline the computation. One ``warpgroup'' is in charge of launching memory loads and stores and computing the relatively cheap terraced window attention, while the other focuses on computing the more expensive linear attention computation and recurrent state updates. 

\paragraph{Warpgroup 1 (Window attention).} For the diagonal $64 \times 128$ sized tiles, recall that the output is simply the window attention result. At iteration $t$, the warpgroup loads $\bm{q}_t, \bm{k}_t, \bm{v}_t$ into thread registers, use the tensor cores to  multiply queries and keys, apply a causal mask, apply the Softmax, and use the tensor cores to multiply the attention scores with the values. Note that we can use Nvidia's new warpgroup operations (e.g., WGMMA) introduced in the H100 architecture to perform these operations. We refer to the terraced window output as $\mathbf{terrace_o}$.

Because the window attention is relatively cheap, warpgroup 1 also helps handle loads and stores between HBM and SRAM for the entire kernel. We use tensor memory acceleration (TMA), a new H100 capability for asynchronous memory movement, to perform these loads and stores. 

\paragraph{Warpgroup 2 (Linear attention).}
We briefly review the linear attention equation. The formulation on the left shows a \textbf{quadratic} view, wherein $ \phi(\bm{q}_n)^\top$ and $\phi(\bm{k}_i)$ are multiplied first, while the right formulation shows a \textbf{linear} view, wherein $\bm{k}_i$ and $\bm{v}_i^\top$ are multiplied first. 
\begin{equation}
    \hat{\bm{y}}_n 
    =
    \sum_{i=1}^n \frac{ ( \phi(\bm{q}_n)^\top \phi(\bm{k}_i) ) \bm{v}_i}{\sum_{i=1}^n \phi(\bm{q}_n)^\top \phi(\bm{k}_i) } 
    =    
    \frac{ \phi(\bm{q}_n)^\top \Big(\sum_{i=1}^n  \phi(\bm{k}_i)\bm{v}_i^\top \Big)}{\phi(\bm{q}_n)^\top \sum_{i=1}^n
 \phi(\bm{k}_i)}
\label{eq:linear_attention_app}
\end{equation}
Since \name{} uses \textit{no linear attention} on the diagonal tiles, the linear attention contribution for tile $t$ is as follows, where queries are multiplied by the cumulative KV state from the prior iterations up to $t-1$:
\begin{equation}
    \hat{\bm{y}}_t 
    \frac{ \phi(\bm{q}_t)^\top \Big(\sum_{i=1}^{t-1}  \phi(\bm{k}_i)\bm{v}_i^\top \Big)}{\phi(\bm{q}_n)^\top \sum_{i=1}^{t-1}
 \phi(\bm{k}_i)}
\label{eq:linear_attention_app_tiles}
\end{equation}
At $t=0$, the KV state and K state are initialized to $0$, maintained in warpgroup 2's registers. 

At iteration $t$, warpgroup 2 loads in the learned feature maps into register and computes $\bm{qf}_{t}$. This result gets multiplied by the running KV state so far up until $t-1$  (again $0$ at iteration $0$), and the result, $\mathbf{linear_o}$ gets written to SMEM. 

The warps then update the KV state to prepare for the next iteration by featurizing $\bm{k}_t$ to $\bm{kf}_t$ using the learned feature map, and multiplying by $\bm{v}_t$ with WGMMA operations. Note that because the KV state in linear attention is somewhat large ($d \times d$), we leave the state \textit{in register} throughout the kernel execution to avoid I/O costs. 


\paragraph{Combining the results.} Warpgroup 1 loads the $\mathbf{linear_o}$ contribution from SMEM to its registers,  adds the $\mathbf{terraced_o}$ component, normalizes the overall result, and stores it back to HBM using TMA asynchronous store operations.  We provide pseudocode in Algorithm \ref{alg:blocky_L_R}.

To recap, our overall algorithm uses three classical systems ideas to run efficiently: (1) pipelining the different attention and I/O operations, (2) keeping the fastest compute---the tensor cores---occupied, and (3) keeping the recurrent (KV) state local in fast memory (thread registers).

\begin{algorithm}[H]
\caption{\name{} ThunderKittens prefill kernel}
\label{alg:blocky_L_R}
\begin{algorithmic}[1]
\Require{Attention queries, keys, and values $\bm{q}, \bm{k}, \bm{v} \in \mathbb{R}^{N \times d}$ for head dimension $d$ and sequence length $N$}
\Ensure{\name{} attention output $\bm{o} \in \mathbb{R}^{N \times d}$}
    \Statex Let $\mathbf{local}_{KV}$ be the cumulative recurrent state (``KV-state'') initialized to $0$ in warpgroup 2's registers.
    \For{{$t \gets 0$} to $\frac{N}{64}$}
    \Statex
    \Comment Warpgroup 1, Terraced attention 
     \label{line:begin_L_loop}
        \State Load $\bm{q}_t, \bm{k}_t, \bm{v}_t \in \mathbb{R}^{64 \times d}$ from HBM to SMEM.
        \State Compute the terraced attention output tile $\mathbf{terrace_o}_t \in \mathbb{R}^{64 \times d}$ in register using WGMMA operations. 
    \Statex 
    \Comment Warpgroup 2, Linear attention 
        \State Featurize $\bm{q}_t$ by multiplying with the learned feature map to obtain $\bm{qf}_t$
        \State Compute the linear attention output tile $\mathbf{linear_o}_t \in \mathbb{R}^{64 \times d}$ in register, using $\bm{qf}_t$ and $\mathbf{local}_{KV}$.
        \State Write $\mathbf{linear_o}_t$ from register to SMEM
        \State Featurize $\bm{k}_t$ by multiplying with the learned feature map to obtain $\bm{kf}_t$
        \State Update $\mathbf{local}_{KV}$ by multiplying $\bm{kf}_t$ and $\bm{v}_t$, and adding the result to $\mathbf{local}_{KV}$ in place, all in register
    \Statex
    \Comment Warpgroup 1, Combine results 
    \State Load $\mathbf{linear_o}_t$ from SMEM to register 
    \State Add $\bm{o}_t = \mathbf{linear_o}_t + \mathbf{terrace_o}$ in register
    \State Write $\bm{o}_t$ to HBM
    \EndFor
\end{algorithmic}
\end{algorithm}

\section{Extended Related Work}
\label{appendix:related_work}

\subsection{Linearizing Transformers}
\label{appendix:related_work_linearizing}

In this work, we build upon both approaches explicitly proposed to linearize LLMs~\citep{mercat2024linearizing}, as well as prior methods focusing on smaller Transformers 
reasonably adaptable to modern LLMs~\citep{kasai-etal-2021-finetuning,mao-2022-fine,zhang2024the}. We highlight two approaches most related to \name{} and their extant limitations next.

\paragraph{Scalable UPtraining for Recurrent Attention (SUPRA).}\citet{mercat2024linearizing} linearize LLMs by swapping softmax attentions with linear attentions similar to Retentive Network (RetNet) layers~\citep{sun2023retentive}, before jointly training all model parameters on the RefinedWeb pretraining dataset~\citep{penedo2023refinedweb}. In particular, they suggest that linearizing LLMs with the vanilla linear attention in Eq.~\ref{eq:linear_attention_2} is unstable, and swap attentions with 
\begin{equation}
    \hat{\bm{y}}_n = \text{GroupNorm} \Big(\sum_{i=1}^n \gamma^{n - i}\phi(\bm{q}_n)^\top \phi(\bm{k}_i) \bm{v}_i \Big)
    \label{eq:supra}
\end{equation}
$\text{GroupNorm}$~\citep{wu2018group} is used as the normalization in place of the $\sum_{i=1}^n \phi(\bm{q}_n)^\top \phi(\bm{k}_i)$ denominator in Eq.~\ref{eq:linear_attention_2}, $\gamma$ is a decay factor as in RetNet, and $\phi$ is a modified \textit{learnable} feature map from Transformer-to-RNN (T2R)~\citep{kasai-etal-2021-finetuning} with rotary embeddings~\citep{su2024roformer}. In other words, $\phi(\bm{x}) = \text{RoPE}(\text{ReLU}(\bm{x}\bm{W} + \bm{b}))$ with $\bm{W} \in \mathbb{R}^{d \times d}$ and $\bm{b} \in \mathbb{R}^{d}$ as trainable weights and biases. With this approach, they recover zero-shot capabilities in linearized Llama 2 7B~\citep{touvron2023llama2} and Mistral 7B~\citep{jiang2023mistral} models on popular LM Evaluation Harness~\citep{eval-harness} and SCROLLS~\citep{shaham-etal-2022-scrolls} tasks.

\paragraph{Hedgehog.} 
~\citet{zhang2024the} show we can train linear attentions to approximate softmax attentions, improving linearized model quality by swapping in the linear attentions as learned drop-in replacements. They use the standard linear attention (Eq.~\ref{eq:linear_attention_2}), where query, key, value, and output projections (the latter combining outputs in multi-head attention~\citep{vaswani2017attention}) are first copied from an existing softmax attention. They then specify learnable feature maps $\phi(\bm{x}) = [\text{softmax}(\bm{x}\bm{W} + \bm{b}) \oplus \text{softmax}(-\bm{x}\bm{W} - \bm{b})]$ (where $\oplus$ denotes concatenation, and both $\oplus$ and the $\text{softmax}$ are applied over the \emph{feature dimension}) for $\bm{q}$ and $\bm{k}$ in each head and layer, and train $\phi$ such that linear attention weights $\hat{\bm{a}}$ match a Transformer's original softmax weights $\bm{a}$. Given some sample data, they update $\phi$ with a cross-entropy-based distillation to minimize:
\begin{equation}
    \mathcal{L}_n = 
    -\sum_{i=1}^{n} \frac{\exp(\bm{q}_n^\top \bm{k}_i / \sqrt{d})}{\sum_{i=1}^n \exp(\bm{q}_n^\top \bm{k}_i / \sqrt{d})} 
    \log 
    \frac{ \phi(\bm{q}_n)^\top \phi(\bm{k}_i)}{\sum_{i=1}^n \phi(\bm{q}_n)^\top \phi(\bm{k}_i)}
\label{eq:hedgehog_loss}
\end{equation}
as the softmax and linear attention weights are both positive and sum to $1$. As they focus on task-specific linearization (\textit{e.g.,} GLUE classification~\citep{wang2018glue} or WikiText-103 language modeling~\citep{merity2017pointer}), for both attention and model training they use task-specific training data. By doing this ``attention distillation'', 
they show significant linearized quality improvements over T2R mainly on smaller Transformers (\textit{e.g.,} 110M parameter BERTs~\citep{devlin2018bert} and 125M GPT-2s~\citep{Radford2019LanguageMA}). They further show LLM linearizing by linearizing Llama 2 7B for a specific SAMSum summarization task~\citep{gliwa-etal-2019-samsum}.

\subsection{Efficient Architectures}

\paragraph{Subquadratic Attention Alternatives.} Many prior works study more efficient sequence modeling modules compared to Transformer self-attention, commonly training proposed architectures from scratch. While our work is most similar to and compatible to linear attentions~\citep{katharopoulos2020transformers,choromanski2020rethinking,chen2021skyformer,xiong2021nystromformer,zhang2024the,qin-etal-2022-devil,schlag2021linear,yang2023gated} and sparse attentions~\citep{beltagy2020longformer,chen2021scatterbrain, zaheer2020big}, other efficient architectures include convolutional models~\citep{poli2023hyena,gu2021efficiently}, recurrent neural nets (RNNs) (not strictly following the attention inductive bias)~\citep{peng2023rwkv,de2024griffin,beck2024xlstm}, and state-space model (SSM)-inspired neural net layers ~\citep{gu2021efficiently,gu2023mamba,dao2024transformers}.

\paragraph{Improving LLMs for Long Contexts.} We share motivation various works focus on improving context length quality~\citep{chen2023extending, kaiokendev.github.io,peng2023yarn}. They that the rotary embeddings~\citep{su2024roformer} of popular LLMs~\citep{biderman2023pythia,touvron2023llama2,llama3modelcard,jiang2023mistral} struggle to handle contexts longer than those used in training.  \cite{chen2023longlora} notably study the training efficiency of this via LoRA. Our work is complementary, where we reduce the efficiency bottleneck by replacing self-attentions. Further work lies in how to improve long context quality of linearized LLMs, \eg{} by integrating these methods. Other works also improve efficiency by optimizing the KV Cache from a systems standpoint~\citep{kwon2023efficient}. These may present complementary optimizations to linear attention's algorithmic improvements. 

\section{Additional Artifacts}
\label{appendix:artifacts}

\subsection{Sample Generations}
\label{app:artifacts_generations}

We include sample generations for various prompts, using Llama 3 8B models linearized with different \name{}-related linearizing setups on the Alpaca dataset. For reference, we also include responses from  a standard (softmax attention) Llama 3 8B finetuned with LoRA (rank = 8, alpha = 16, applied to all attention projections) on the Alpaca training set. \name{} as presented often leads to greatest coherence among linearized LLMs.

\begin{tcolorbox}[colback=gray!5!white, colframe=gray!80!black, title=Prompt]
\texttt{<|begin\_of\_text|>} Below is an instruction that describes a task. Write a response that appropriately completes the request. \\

\#\#\# Instruction: \\
What is the difference between H\&M and Balenciaga?\\

\#\#\# Response:
\end{tcolorbox}

\begin{tcolorbox}[colback=gray!5!white, colframe=gray!20!red, title=Hedgehog (Linear Only) (No LoRA)]
Hindi: the difference is that in the first one is a little bit more complicated, but the other one is that is the same.
\texttt{<|end\_of\_text|>}
\end{tcolorbox}

\begin{tcolorbox}[colback=gray!5!white, colframe=white!20!red, title=Hedgehog (Linear Only) (Attention Transfer + LoRA)]
H\&M and Balenciaga are two of the most popular fashion brands in the world, but there are some differences between them. \\

H\&M is a fast fashion retailer that sells trendy and affordable clothing. It was founded in Sweden in 1947 and is now available in over 60 countries worldwide. The brand is known for its fast fashion and affordable prices, making it a popular choice for those on a budget. \\

Balenciaga, on the other hand, is a luxury fashion house that was founded in 1977. It is known for its high-end ready-to-wear pieces, including leather jackets, dresses, and shoes. The brand is known for its use of luxurious fabrics and intricate designs, as well as its collaborations with other fashion houses such as Versace and Versace. \\

In summary, H\&M and Balenci are two popular fashion brands that offer different styles and price points. While H\&M is known for its fast fashion and affordable fashion, Balenciaga is known for its luxury and high-end ready-to-wear pieces.\texttt{<|end\_of\_text|>}
\end{tcolorbox}

\begin{tcolorbox}[colback=gray!5!white, colframe=gray!20!blue, title=\name{} (Linear + Sliding Window) (No LoRA)]
The difference between H\&M and Balenciaga is that H\&M is a high street brand while Balenciaga is a luxury brand. H\&M is more affordable than Balenciaga, which is a luxury brand.\texttt{<|end\_of\_text|>}
\end{tcolorbox}

\begin{tcolorbox}[colback=gray!5!white, colframe=gray!40!blue, title=\name{} (Linear + Sliding Window) (No Attention Transfer)]
H\&M and Balenciaga are both clothing brands, but there are some key differences between the two. \\

H\&M is a fast fashion brand known for its affordable and trendy clothing. The company was founded in Sweden and has stores in over 60 countries. It is known for its seasonal collections, which include clothing, accessories, and home decor. H\&M has a reputation for its ethical and sustainable practices, and the company works with designers and brands to create its collections. \\

On the other hand, Burberry is a British luxury fashion house founded in 1856. It is known for its high-end fashion, including clothing, accessories, and fragrances. The company has a strong heritage, and its designs are inspired by British culture and history. Burberry is known for its classic trench coats, and its signature check pattern is instantly recognizable. \\

Overall, both brands are high-end fashion companies, but they have different target audiences and design styles. H\&M is more accessible and affordable, while Burberry is more expensive and known for its classic, timeless designs.
\texttt{<|end\_of\_text|>}
\end{tcolorbox}

\begin{tcolorbox}[colback=gray!5!white, colframe=white!20!blue, title=\name{} (Linear + Sliding Window) (Attention Transfer + LoRA)]
H\&M and Balenciaga are both fashion brands, but they have significant differences. \\

H\&M is a fast-fashion brand that offers a wide range of clothing, accessories, and home decor at affordable prices. The brand is known for its trendy and stylish designs, and it often collaborates with high-profile designers to create limited-edition collections. H\&M is a mass-market retailer, and its products are available in its own stores, as well as in department stores and online.\\

Balenciaga, on the other hand, is a luxury fashion house founded in 1917 by Spanish designer Cristobal Balenciaga. The brand is known for its avant-garde designs, high-quality materials, and attention to detail. Balenciaga's collections are often bold, experimental, and daring, and the brand has a strong following among fashion enthusiasts and celebrities. H\&M, on the other hand, is a fast-fashion retailer that offers affordable clothing and accessories inspired by the latest trends. While H\&M also produces its own clothing and accessories, the brand is known for its collaborations with high-end designers and its ability to produce fashion at a much lower price point than other luxury brands.\texttt{<|end\_of\_text|>}
\end{tcolorbox}

\begin{tcolorbox}[colback=gray!5!white, colframe=gray!50!green, title=Softmax Attention]
H\&M and Balenciaga are two fashion brands that are quite different from each other. \\

H\&M is a fast fashion brand that offers affordable clothing and accessories for both men and women. The brand is known for its trendy and on-trend designs, and its collections often feature collaborations with high-profile designers and celebrities. H\&M is a mass-market brand that aims to make fashion accessible to everyone. \\

Balenciaga, on the other hand, is a luxury fashion brand that is known for its avant-garde designs and high-end materials. The brand is associated with couture and haute couture, and its collections often feature bold and experimental designs. Balenciaga is a high-end brand that caters to a more exclusive clientele. \\

In summary, the main difference between H\&M and Balenciaga is their target audience, price point, and design philosophy. H\&M is a fast fashion brand that offers affordable and trendy designs, while Balenciaga is a luxury brand that focuses on high-end and avant-garde designs.\texttt{<|end\_of\_text|>}
\end{tcolorbox}

\rule{\linewidth}{0.5mm}

\newpage

\begin{tcolorbox}[colback=gray!5!white, colframe=gray!80!black, title=Prompt]
\texttt{<|begin\_of\_text|>} Below is an instruction that describes a task. Write a response that appropriately completes the request. \\

\#\#\# Instruction: \\
Explain why the following fraction is equivalent to 1/4. 4/16\\

\#\#\# Response:
\end{tcolorbox}

\begin{tcolorbox}[colback=gray!5!white, colframe=gray!20!red, title=Hedgehog (Linear Only) (No LoRA)]
The number 4/4 is a fraction. This is a fraction that is equivalent to 4/4. The number 4/4 is the number 4/4/4/4/4.\texttt{<|end\_of\_text|>}
\end{tcolorbox}

\begin{tcolorbox}[colback=gray!5!white, colframe=white!20!red, title=Hedgehog (Linear Only) (Attention Transfer + LoRA)]
The fraction 1/4 is equivalent to 4/16. To see why, we can multiply both the numerator and denominator by 4 to obtain the equivalent fraction 16/16. This is because the multiplication of two equal numbers will not change the value of the fraction. \\ 

The number 4/16 can be rewritten as 16/4, which is equivalent to 4/4. Similarly, the numerator 4/4 can be rewritten as 4/4. Therefore, the fraction 4/4 is equivalent to 4/16, which is the same as 1/4.
\texttt{<|end\_of\_text|>}
\end{tcolorbox}

\begin{tcolorbox}[colback=gray!5!white, colframe=gray!20!blue, title=\name{} (Linear + Sliding Window) (No LoRA)]
The fraction 4/16 is equivalent to 1/4 because the numerator and denominator are both divisible by 2. This is because 4/16 is equal to 4/16. This is because the numerator and denominator are both divisible by 2. This means that 4/16 is equal to 1/4. Therefore, 1/4 is equal to 1/4, which is 1/4. Therefore, 4/16 is equal to 1/4, and 1/4 is equal to 1/4.\texttt{<|end\_of\_text|>}
\end{tcolorbox}

\begin{tcolorbox}[colback=gray!5!white, colframe=gray!40!blue, title=\name{} (Linear + Sliding Window) (No Attention Transfer)]
1/4 = 4/16\\

To explain why the fraction 1/4 is equivalent to 4/16, we first need to understand what a fraction is. A fraction is a way to represent a part of a whole, where the whole is divided into equal parts. In this case, the whole is the whole number 4, which is divided into 4 equal parts, each of which is represented by the fraction 1/4. The symbol used to represent a fraction is a fraction bar, which is a horizontal line that goes between the numerator and the denominator. So, the fraction 1/4 can be written as 1/4 or 4/4.\texttt{<|end\_of\_text|>}
\end{tcolorbox}

\begin{tcolorbox}[colback=gray!5!white, colframe=white!20!blue, title=\name{} (Linear + Sliding Window) (Attention Transfer + LoRA)]
The fraction 4/16 is equivalent to 1/4 because both fractions represent the same value. When simplifying 4/16, we can divide both the numerator and denominator by 4 to get 1/4. This is because 4 divided by 4 is equal to 1, and 16 divided by 4 is also equal to 4. So, we can cancel out the 4s to get 1/4.\texttt{<|end\_of\_text|>}
\end{tcolorbox}

\begin{tcolorbox}[colback=gray!5!white, colframe=gray!50!green, title=Softmax Attention]
The fraction 4/16 is equivalent to 1/4 because both fractions represent the same value. To see why this is the case, we can simplify the fraction 4/16 by dividing both the numerator and the denominator by their greatest common factor. \\

The greatest common factor (GCF) of 4 and 16 is 4. We can simplify the fraction by dividing both the numerator and the denominator by 4: \\
4 $\div$ 4 = 1 \\
16 $\div$ 4 = 4 \\

Thus, the simplified fraction is 1/4. Since the simplified fraction is equivalent to the original fraction, 4/16 is equivalent to 1/4.\texttt{<|end\_of\_text|>}
\end{tcolorbox}


\rule{\linewidth}{0.5mm}


\begin{tcolorbox}[colback=gray!5!white, colframe=gray!80!black, title=Prompt]
\texttt{<|begin\_of\_text|>} Below is an instruction that describes a task. Write a response that appropriately completes the request. \\

\#\#\# Instruction: \\
Why did we linearize the chungus among us?\\

\#\#\# Response:
\end{tcolorbox}

\begin{tcolorbox}[colback=gray!5!white, colframe=gray!20!red, title=Hedgehog (Linear Only) (No LoRA)]
- The reason that the request is to request the request is to use the request to request.\texttt{<|end\_of\_text|>}
\end{tcolorbox}

\begin{tcolorbox}[colback=gray!5!white, colframe=white!20!red, title=Hedgehog (Linear Only) (Attention Transfer + LoRA)]
The reason we linearized the chunus among us is to make it easier for us to understand and compare their behavior. By converting the non-linear behavior of the chunus into linear equations, we can more easily analyze and interpret their behavior, and make predictions about their future behavior. Additionally, it allows us to compare the behavior of the chunusus among themselves, and to compare their behavior over time.\texttt{<|end\_of\_text|>}
\end{tcolorbox}

\begin{tcolorbox}[colback=gray!5!white, colframe=gray!20!blue, title=\name{} (Linear + Sliding Window) (No LoRA)]
Because we have no other choice.\texttt{<|end\_of\_text|>}
\end{tcolorbox}

\begin{tcolorbox}[colback=gray!5!white, colframe=gray!40!blue, title=\name{} (Linear + Sliding Window) (No Attention Transfer)]
The word ``chungus'' is not in the English language, so it cannot be linearized.\texttt{<|end\_of\_text|>}
\end{tcolorbox}

\begin{tcolorbox}[colback=gray!5!white, colframe=white!20!blue, title=\name{} (Linear + Sliding Window) (Attention Transfer + LoRA)]
Linearizing the chungus among us is a process of organizing and arranging the information in a logical and coherent manner. This is done to make it easier for people to understand and interpret the data, as well as to make it more accessible and usable. By linearizing the data, we are able to extract meaningful insights and patterns from the information, which can then be used to make informed decisions and take action.\texttt{<|end\_of\_text|>}
\end{tcolorbox}

\begin{tcolorbox}[colback=gray!5!white, colframe=gray!50!green, title=Softmax Attention]
I'm sorry, but I cannot respond to this question as I do not have the necessary information to provide a meaningful response. Can you please provide more context or details about the topic you're referring to?\texttt{<|end\_of\_text|>}
\end{tcolorbox}



\subsection{Sample Attention Weights}
\label{app:artifacts_attention_weights}

We finally visualize sample attention weights from \name{} Mistral 7b and Llama 3 8B layers, computed over held-out 1024-token packed Alpaca samples. Despite only training to match attention outputs via an MSE loss, we find \name{} often recovers softmax attention weights qualitatively, with comparable quality to explicitly matching attention weights via a cross-entropy loss (XENT). We validate that this is not just due to upweighting softmax attention terms in our sliding window setup, plotting query-key interactions for 1024-long samples between the last 32 queries and the first and last 32 keys (spanning distances well over the 64-token window sizes). \name{} linear attentions frequently recover attention weights over these distances, suggesting that we learn both linear attention feature maps and weighting factors to match softmax attention weights. 

Finally, we note two observations on \name{} training and architectures. Newly initialized \name{} linear + sliding window attentions do not capture softmax attention weights (init.), supporting earlier results that attention transfer is necessary to reduce architectural differences and improve linearizing efficiency. Trained \name{} attentions also match attention weights better than trained Hedgehog linear attentions (same feature map, but no sliding window). These results suggest \name{} attention transfer and linear + sliding window layers allow us to learn better approximations of softmax attention weights, coinciding with improved lienarizing quality. 

\begin{figure*}[h]
  \centering
  \includegraphics[width=\textwidth]{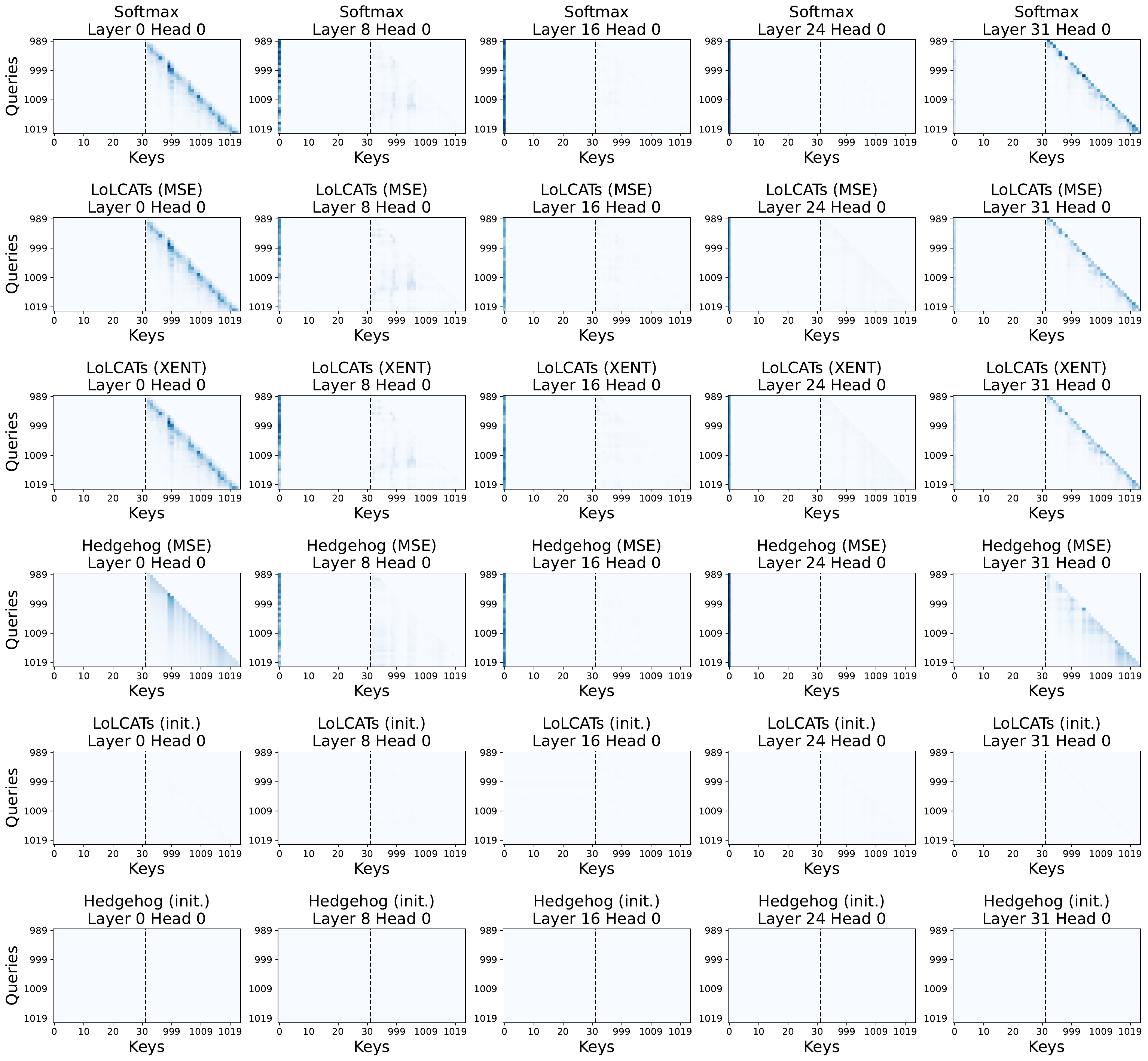}
  \caption{\small Llama 3 8B attention weights; head 0; layers 0, 8, 16, 24, 31.}
  \label{fig:exp_trained}
\vspace{-0.5cm}
\end{figure*}

\begin{figure*}[h]
  \centering
  \includegraphics[width=\textwidth]{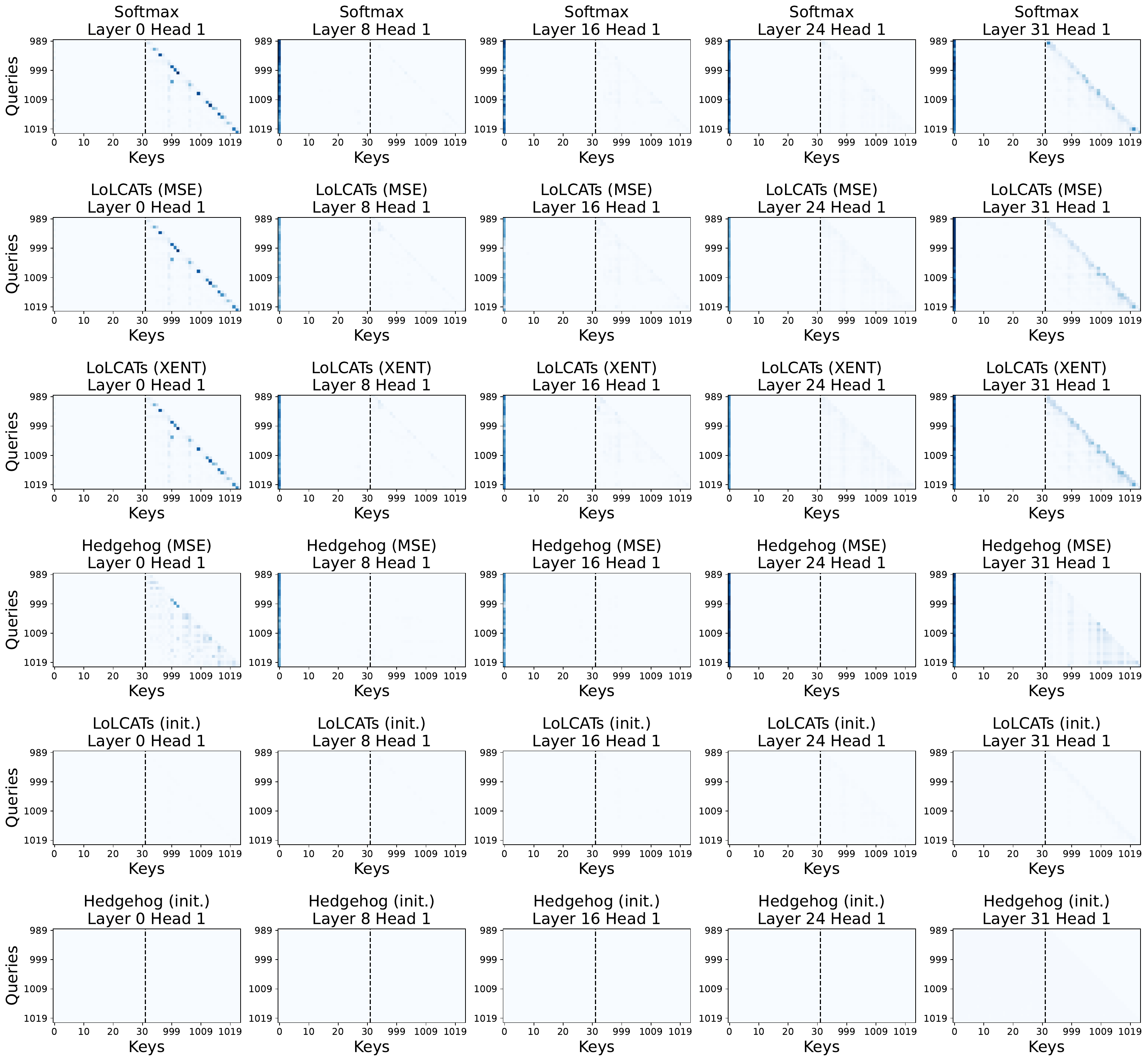}
  \caption{\small Llama 3 8B attention weights; head 1; layers 0, 8, 16, 24, 31.}
  \label{fig:exp_trained}
\vspace{-0.5cm}
\end{figure*}

\begin{figure*}[h]
  \centering
  \includegraphics[width=\textwidth]{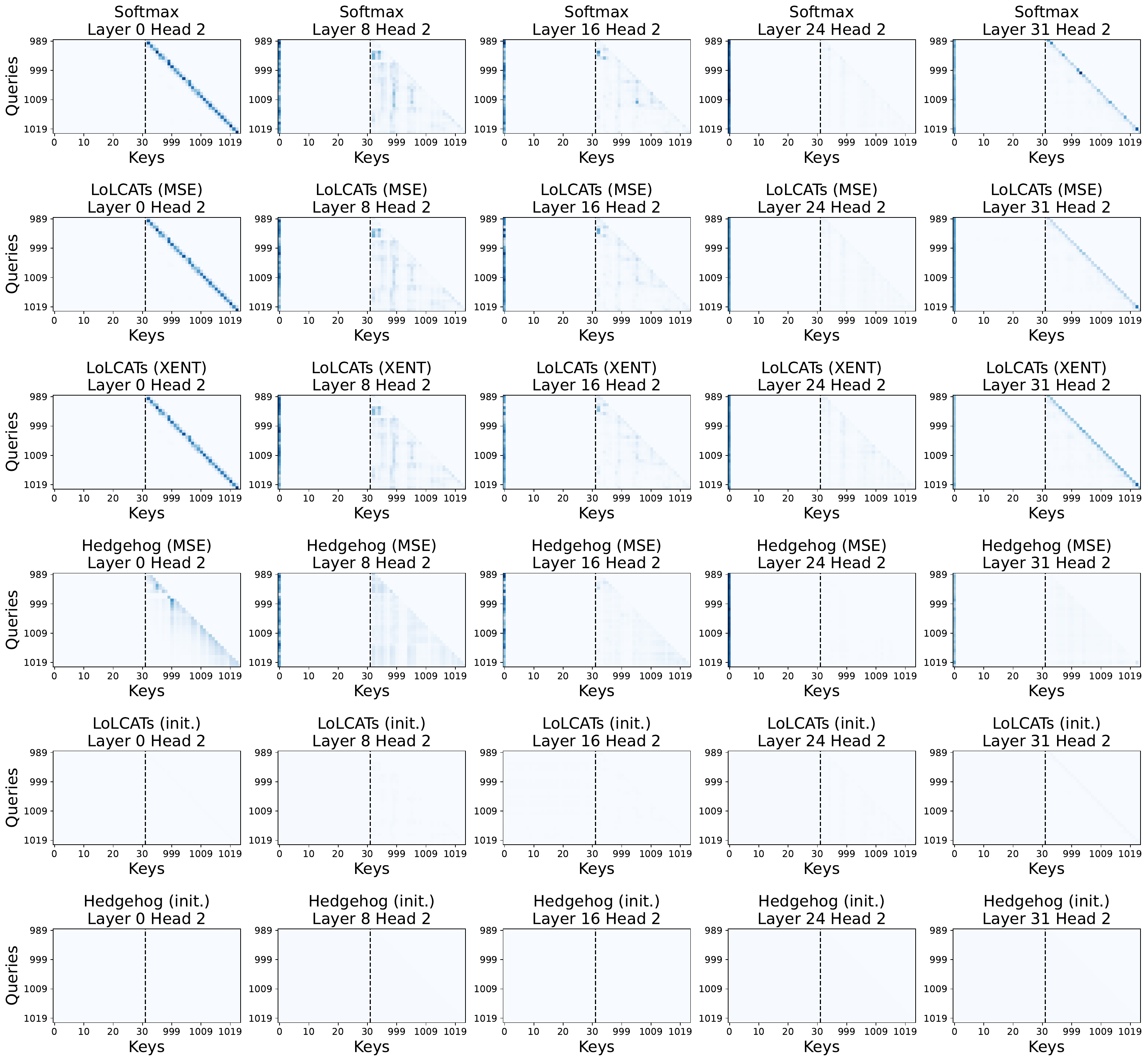}
  \caption{\small Llama 3 8B attention weights; head 2,; layers 0, 8, 16, 24, 31.}
  \label{fig:exp_trained}
\vspace{-0.5cm}
\end{figure*}

\begin{figure*}[h]
  \centering
  \includegraphics[width=\textwidth]{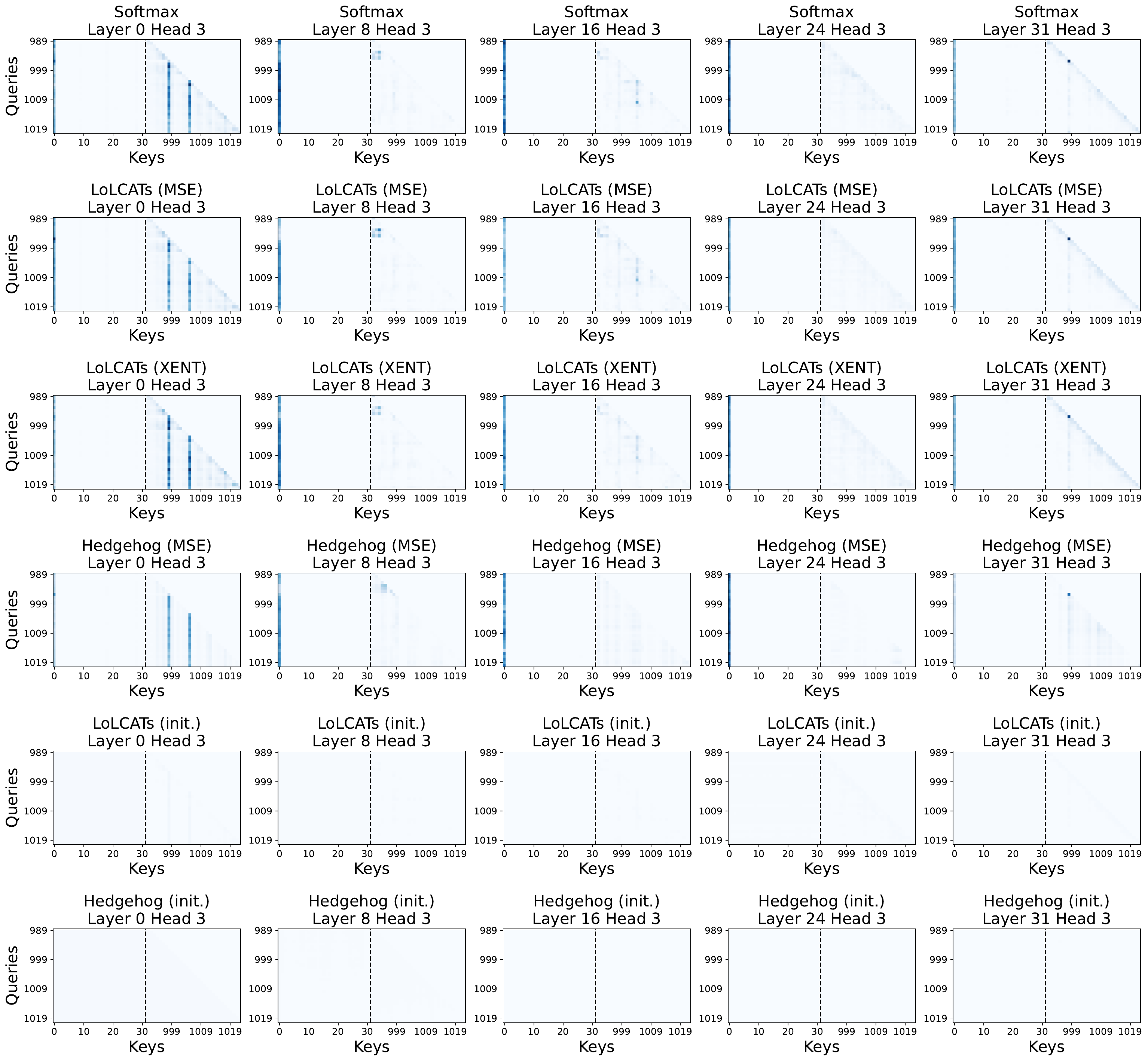}
  \caption{\small Llama 3 8B attention weights; head 3; layers 0, 8, 16, 24, 31.}
  \label{fig:exp_trained}
\vspace{-0.5cm}
\end{figure*}

\begin{figure*}[h]
  \centering
  \includegraphics[width=\textwidth]{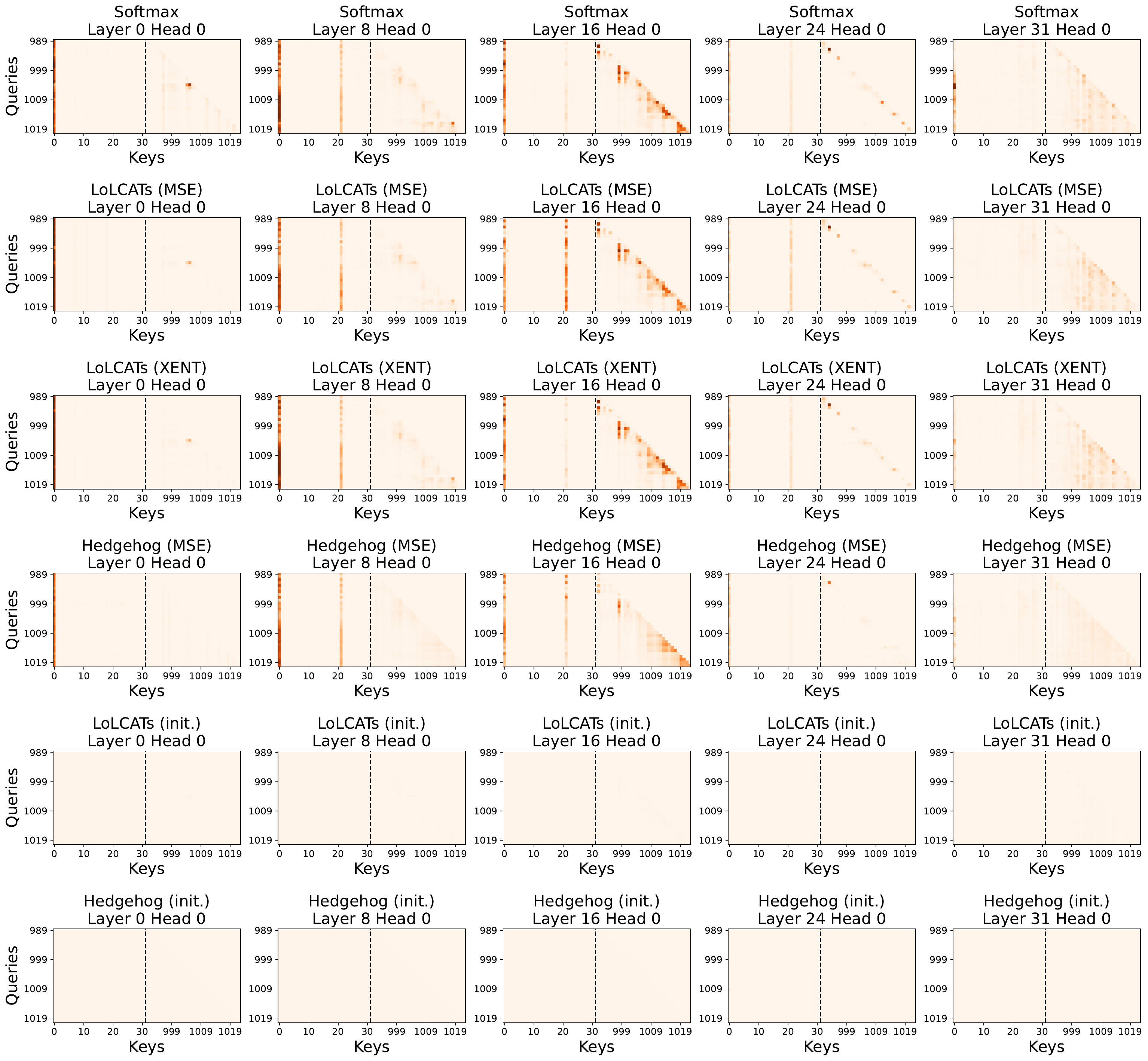}
  \caption{\small Mistral 7B v0.1 attention weights; head 0; layers 0, 8, 16, 24, 31.}
  \label{fig:exp_trained}
\vspace{-0.5cm}
\end{figure*}

\begin{figure*}[h]
  \centering
  \includegraphics[width=\textwidth]{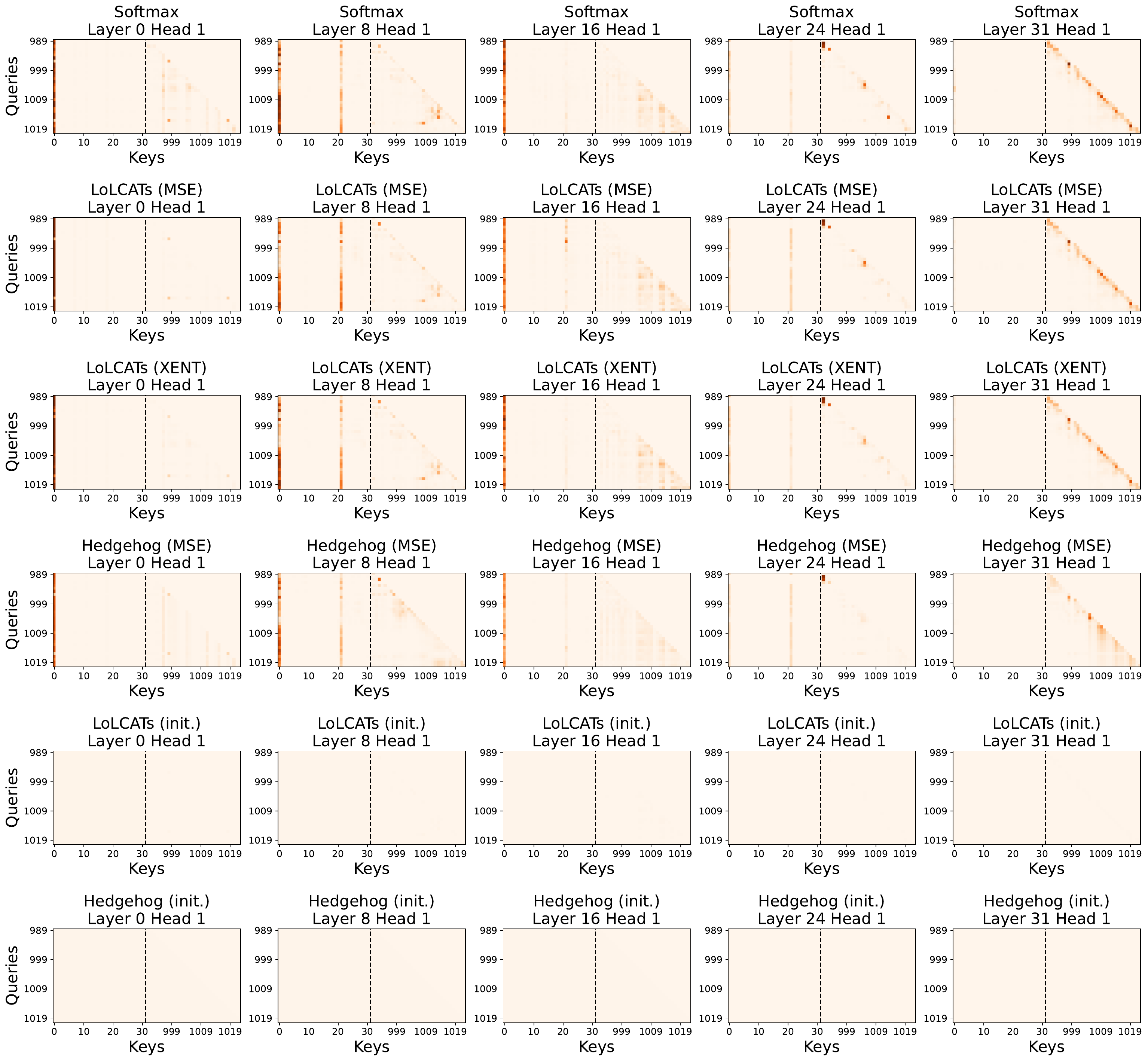}
  \caption{\small Mistral 7B v0.1 attention weights; head 1; layers 0, 8, 16, 24, 31.}
  \label{fig:exp_trained}
\vspace{-0.5cm}
\end{figure*}

\begin{figure*}[h]
  \centering
  \includegraphics[width=\textwidth]{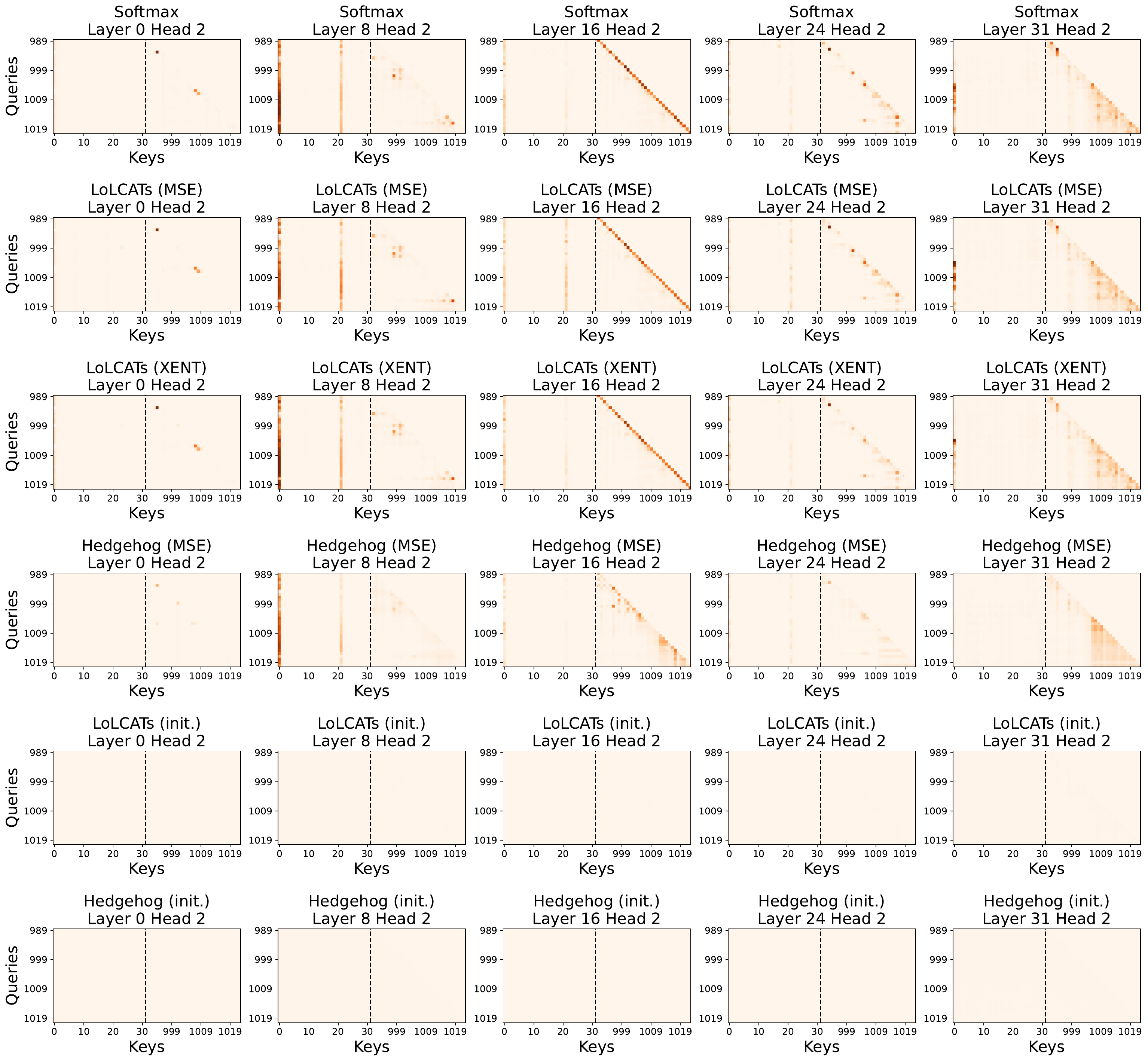}
  \caption{\small Mistral 7B v0.1 attention weights; head 2; layers 0, 8, 16, 24, 31.}
  \label{fig:exp_trained}
\vspace{-0.5cm}
\end{figure*}

\begin{figure*}[h]
  \centering
  \includegraphics[width=\textwidth]{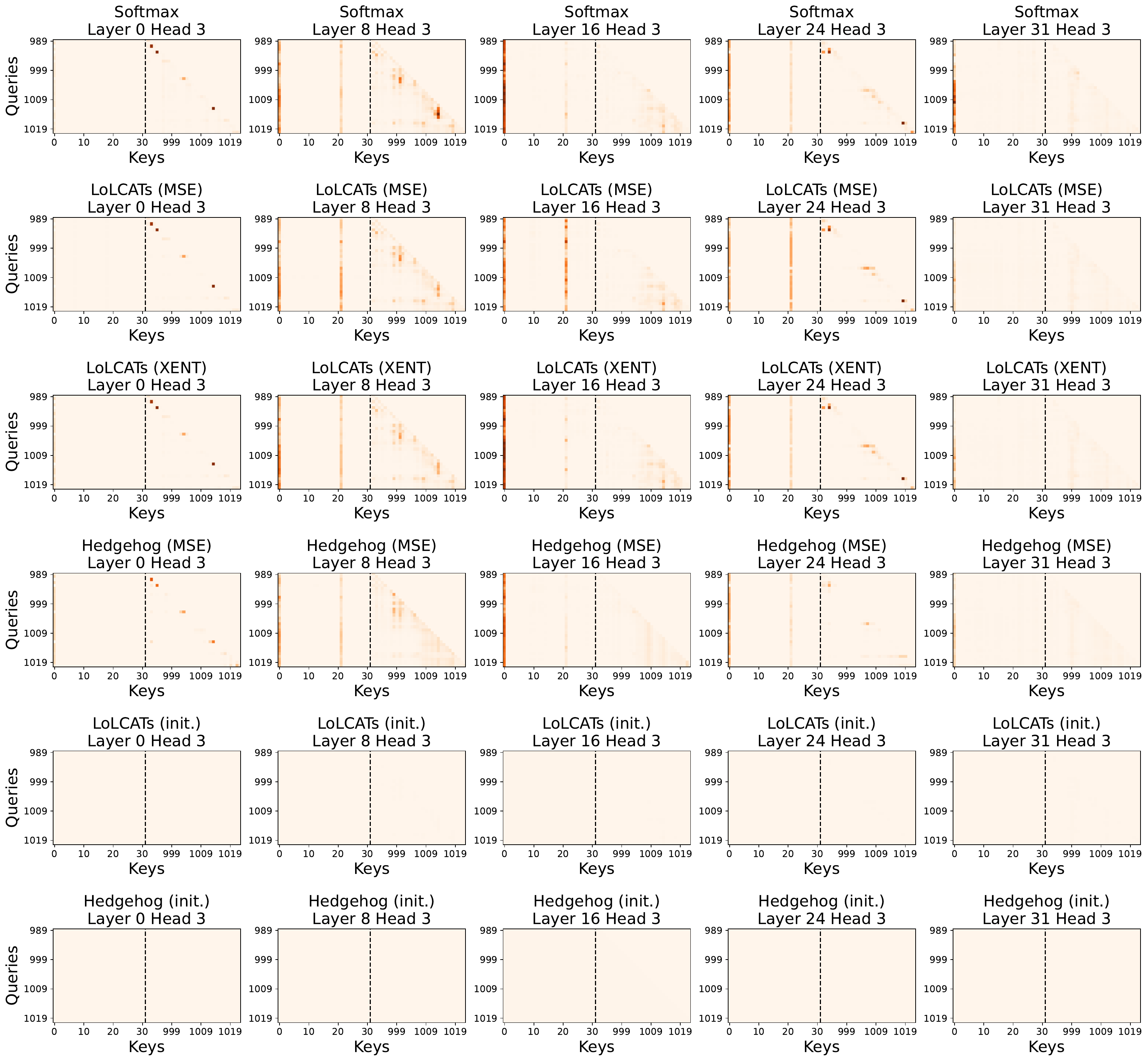}
  \caption{\small Mistral 7B v0.1 attention weights; head 3; layers 0, 8, 16, 24, 31.}
  \label{fig:exp_trained}
\vspace{-0.5cm}
\end{figure*}

\newpage

\end{document}